\documentclass{article}

\usepackage[final]{neurips_2022}
\usepackage[utf8]{inputenc} 
\usepackage[T1]{fontenc}    
\usepackage{hyperref}       
\usepackage{url}            
\usepackage{booktabs}       
\usepackage{amsfonts}       
\usepackage{nicefrac}       
\usepackage{microtype}      
\usepackage{xcolor}         
\usepackage{multirow}
\usepackage{graphicx}
\usepackage{color, colortbl}
\usepackage{pifont} 
\usepackage{graphicx}
\usepackage[pdf]{pstricks}
\usepackage{mathrsfs}
\usepackage[bold]{hhtensor}
\usepackage{wrapfig}
\usepackage{algorithm}
\usepackage{listings}
\usepackage[misc]{ifsym} 
\usepackage{pythonhighlight}
\usepackage{caption}

\usepackage[caption=false,position=bottom]{subfig}
\definecolor{Gray}{gray}{0.9}
\definecolor{White}{gray}{1.0}
\newcommand{\cmark}{\ding{51}}%
\newcommand{\xmark}{\ding{55}}%

\newcommand{\squishlist}{
 \begin{list}{$\bullet$}
  { \setlength{\itemsep}{0pt}
     \setlength{\parsep}{1pt}
     \setlength{\topsep}{1pt}
     \setlength{\partopsep}{0pt}
     \setlength{\leftmargin}{1.5em}
     \setlength{\labelwidth}{1em}
     \setlength{\labelsep}{0.5em} } }
\newcommand{\squishend}{
  \end{list}  }

\title{Decoupling Classifier for Boosting Few-shot  Object Detection and Instance Segmentation}

\author{
	Bin-Bin Gao$^{1}$\thanks{Corresponding author: B.-B. Gao~(csgaobb@gmail.com) and C. Wang~(jasoncjwang@tencent.com).} ~~~
	Xiaochen Chen$^{1}$  ~~~
	Zhongyi Huang$^{1}$ ~~~
	Congchong Nie$^{1}$ ~~~
	Jun Liu$^{1}$ \\
	\textbf{Jinxiang Lai}$^{1}$ ~~~
	\textbf{Guannan Jiang}$^{2}$ ~~~
	\textbf{Xi Wang}$^{2}$ ~~~
	\textbf{Chengjie Wang}$^{1}$  \\
	$^{1}$Tencent YouTu Lab ~~~~~
	$^{2}$CATL \\
}

\begin{document}
\makeatletter
\def\thanks#1{\protected@xdef\@thanks{\@thanks
        \protect\footnotetext{#1}}}
\makeatother

\maketitle

\begin{abstract}
  This paper focus on few-shot object detection~(FSOD) and instance segmentation~(FSIS), which requires a model to quickly adapt to novel classes with a few labeled instances. The existing methods severely suffer from bias classification because of the missing label issue which naturally exists in an instance-level few-shot scenario and is first formally proposed by us. Our analysis suggests that the standard classification head of most FSOD or FSIS models needs to be decoupled to mitigate the bias classification. Therefore, we propose an embarrassingly simple but effective method that decouples the standard classifier into two heads. Then, these two individual heads are capable of independently addressing clear positive samples and noisy negative samples which are caused by the missing label. In this way, the model can effectively learn novel classes while mitigating the effects of noisy negative samples. Without bells and whistles, our model without any additional computation cost and parameters consistently outperforms its baseline and state-of-the-art by a large margin on PASCAL VOC and MS-COCO benchmarks for FSOD and FSIS tasks.\footnote{\url{https://csgaobb.github.io/Projects/DCFS}.}
\end{abstract}

\section{Introduction}
Fully supervised deep convolutional neural network have achieved remarkable progress on various computer 
vision tasks, such as image classification~\cite{he2016deep}, object detection~\cite{girshick2015fast,dai2017deformable,redmon2017yolo9000}, 
semantic segmentation~\cite{long2015fully,chen2017deeplab} and instance segmentation~\cite{he2017mask} 
in recent years. However, the superior performance heavily depends on a large amount of annotated images.
In contrast, humans can quickly learn novel concepts from a few training examples. To this end,
a few-shot learning paradigm~\cite{fink2004object} is presented, and its goal aims to adapt novel classes when only providing 
a few labeled examples~(instances). Unfortunately, existing few-shot models are still far behind humans, especially for 
few-shot object detection (FSOD) and few-shot instance segmentation (FSIS).

Various methods have been proposed to tackle the problem of the FSOD and FSIS. The earlier works~\cite{wang2019metalearn,yan2019metarcnn} 
mainly follow meta-learning paradigm~\cite{finn2017model} to acquire task-level knowledge on base classes and generalize better to novel classes. However, 
these methods usually suffer from a complicated training process (episodic training) and data organization~(support query pair).
The recent transfer-learning~(fine-tuning) methods~\cite{wang2020tfa,wu2020multi,sun2021fsce,cao2021fadi,qiao2021defrcn} significantly 
outperforms the earlier meta-learning ones. Furthermore, it is more simple and efficient. These transfer-learning methods mainly follow a fully 
supervised object detection or instance segmentation framework, e.g., Faster-RCNN~\cite{ren2015faster} or Mask-RCNN~\cite{he2017mask}. 
Therefore, it may be suboptimal for few-shot scenario.

\begin{figure}[t]
  \centering
  \subfloat[PASCAL VOC]
  {
  \includegraphics[width= 0.3\columnwidth]{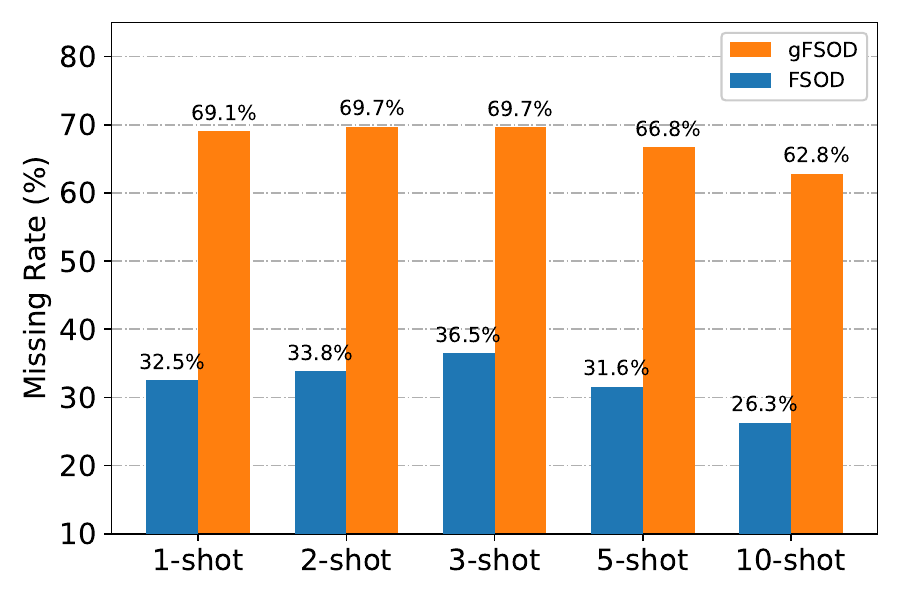}
  }
  \subfloat[MS-COCO]
  {
  \includegraphics[width= 0.3\columnwidth]{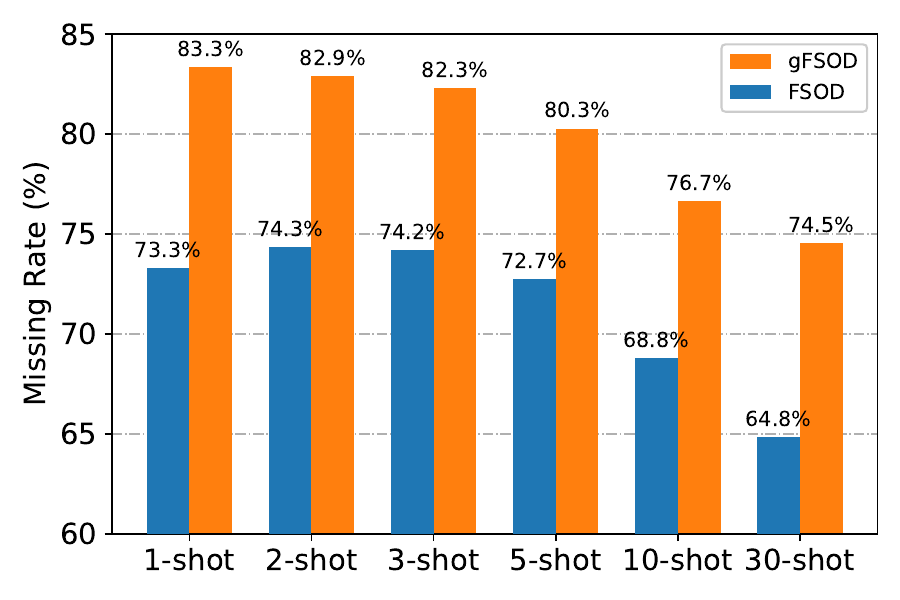}
  }
  \subfloat[{a one-shot labeled image}]
  {
  \includegraphics[width=0.3\columnwidth]{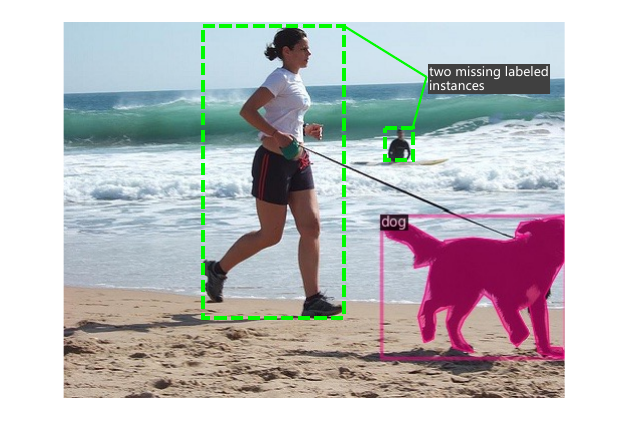}
  }
  \vspace{-8pt}
  \caption{\small{The proportion of missing instances in the training set for FSOD and gFSOD on (a) PASCAL-VOC and (b) MS-COCO datasets. 
  It can be observed that there is a high missing rate in each shot, especially for the gFSOD. In (c), two ``person'' instances present in 
  the one-shot labeled image, but they are mislabeled.}} \label{fig:miss}
  \vspace{-10pt}
 \end{figure}

The PASCAL VOC~\cite{everingham2010pascal} and MS-COCO~\cite{lin2014microsoft} are widely used to evaluate the performance of 
object detection or instance segmentation. Under a fully supervised setting, the model can be well-trained on these two datasets  
because all interest objects are almost completely labeled. Under an instance-level few-shot scenario, however, we find that there is a large 
number of instances that are missing annotations as shown in Fig.~\ref{fig:miss} (detailed discussion in Sec.~\ref{sec:revist}). 
The reason is that the community considers an instance as a shot for controlling the number of labeled instances when building instance-level benchmarks. This is different from image-level few-shot image classification~\cite{vinyals2016matching} because there are generally multiple instances in an image for instance-level few-shot learning. In fact, missing (partial or incomplete) label learning is more difficult and challenging, especially instance-level few-shot scenarios. It requires that learning algorithms deal with training images each associated with multiple instances, among which only partial instances are labeled; while the common supervised learning typically assumes that all interest instances are fully labeled. In some real-world applications, such as open-vocabulary object detection~\cite{zareian2021open}, it is almost impossible to label all instances, and thus there still may exist some instances left to be missing labeled. In addition, it is more friendly and convenient for users to label partial instances than all ones even in few-shot settings.

Most methods~\cite{zhang2020partial,nguyen2020incomplete} have been developed to address missing label (partial label or incomplete label) learning but mainly focus on image-level multi-class or multi-label classification. To address the instance-level missing label issue, some recent works have attempted to regard the missing (unlabeled) instances as hard negative samples and re-weight~\cite{wu2019soft} or re-calibrate~\cite{Zhang2020SolvingMO} their losses. However, these works still only focus on general object detection.
For instance-level few-shot recognition, it may result in biased classification and thus limit the generalization ability of novel classes using the model trained on these mislabeled datasets if we don't take any action.

Recently, one work closely related to ours is the state-of-the-art DeFRCN~\cite{qiao2021defrcn} which decouples Faster-RCNN to alleviate the foreground-background confusion between base pre-training and novel fine-tuning in FSOD. It also can be interpreted from a missing label perspective. Here, we could view fine-tuning few-shot learning paradigm as a domain adaption procedure from base to novel. In this procedure, a few-shot detector may suffer from foreground-background confusion because one background proposal (negative class may be potential novel class)  in the base learning stage will become foreground (positive class) in the novel fine-tuning phase. To mitigate the label conflict between the two domains, DeFRCN decouples RCNN and RPN by stopping gradient backpropagation of RPN in Faster-RCNN. Different from the missing label of cross-domain in DeFRCN, we focus on the missing label issue only in the novel (or balanced base-novel) fine-tuning stage. Another recent work, Pseudo-Labelling~\cite{niitani2019sampling,kaul2022label} mines the missing labeled instances for increasing the number of positive training samples and reducing the biased classification. Unfortunately, this method may lead to a chicken-and-egg problem–we need a good few-shot detector to generate good pseudo labels, but we need good few-shot annotations to train a good few-shot detector.
Unlike this work, our method completely avoids using any pseudo-label information. 

In this paper, we propose a 
simple decoupling method to mitigate the biased classification issue.  Specifically, we firstly decouple the standard 
classifier into two parallel heads, positive and negative ones. Then, these two heads 
independently process clear positive and noisy negative samples with different strategies.  
Our contributions are summarized as follows:

\squishlist 
\item We rethink FSOD and FSIS from the perspective of label completeness and discover that existing transfer-learning few-shot methods severely suffer from bias classification because the missing label issue naturally 
exists in instance-level few-shot scenarios.
To be best of our knowledge, this is the first to propose missing label issue in FSOD and FSIS.
\item To mitigate the bias classification, we propose a simple but effective method that decouples the standard classifier into two parallel heads to independently 
process clear positive samples and noisy negative ones. Without bells and whistles, the proposed 
decoupling classifier can be taken as an alternative to the standard classifier in state-of-the-art FSOD or FSIS models.
\item 
Comprehensive experimental results on PASCAL VOC and MS-COCO show that our approach without 
any additional parameters and computation cost outperforms state-of-the-art both on FSOD and FSIS tasks.  
\squishend

\section{Related Work}

\textbf{FSOD} aims to recognize novel objects and localize them with bounding boxes when only providing 
a few training instances on each novel class. Existing works can be roughly grouped into two families, meta-learning 
and transfer-learning, according to training paradigm. 
The meta-learning methods~\cite{wang2019metalearn,
kang2019frsw,yan2019metarcnn,fan2020few,Xiao2020FSDetView,li2021tip,li2021cme,zhu2021semantic,hu2021dcnet,han2022ct} use episodic training 
to acquire task-level knowledge on base classes and generalize better to novel classes.
The transfer-learning methods~\cite{wang2020tfa,wu2020multi,sun2021fsce,zhu2021semantic,cao2021fadi,qiao2021defrcn} generally 
utilize two-stage training strategy first pre-training on base classes and then fine-tuning on novel classes, 
which have significantly outperformed many earlier meta-learning approaches. 
Recent FSCE~\cite{sun2021fsce} shows that the degradation of detection performance mainly comes from misclassifying 
novel instances as confusable classes, and they propose a contrastive proposal encoding loss to ease the issue.
Similar to the FSCE, FADI~\cite{cao2021fadi} explicitly associates each novel class with a semantically 
similar base class to learn a compact intra-class distribution. DeFRCN~\cite{qiao2021defrcn} decouples 
Faster-RCNN~\cite{ren2015faster} to alleviate the foreground-background confusion between pre-training and fine-tuning stage.

\textbf{FSIS} needs to not only recognize novel objects and their location but also perform pixel-level semantic segmentation 
for each detected instance. Many FSIS approaches typically use the FSOD framework, e.g., Mask-RCNN~\cite{he2017mask}, it 
is built on the Faster-RCNN by adding a mask segmentation head. Therefore, most FSIS methods generally follow the FSOD learning paradigm, i.e., 
meta-learning~\cite{michaelis2018one,yan2019metarcnn,fan2020fgn,nguyen2021fapis} or transfer-learning~\cite{ganea2021incremental}. 
Siamese Mask-RCNN~\cite{michaelis2018one} and Meta-RCNN~\cite{yan2019metarcnn} commonly compute embeddings of support images 
and combine them with those features of query images produced by a backbone network. Their difference is the combination strategy, e.g., subtraction 
in~\cite{michaelis2018one} and concatenation in~\cite{yan2019metarcnn}. These works only focus on the performance 
of novel classes and ignore that of base classes. In real-world applications, we expect that one few-shot model not only recognizes
novel classes but also remembers base classes. The recent iMTFA~\cite{ganea2021incremental} introduces incremental learning 
into FSIS and propose incremental FSIS task.

\section{Methodology}\label{sec:method}
\subsection{Few-shot Object Detection and Instance Segmentation Setting}

Given an image dataset $D=\{X_i, Y_i\}_{i=1}^N$, where $X_i$ denotes the $i$-th image and $Y_i$ 
is its corresponding annotation. For object detection, $Y_i$$=$$\{b_k, c_k\}_{k=1}^M$, where $b_k$ and 
$c_k$ represent bounding box coordinates and category of the $k$-th instance presented image $X_i$, 
respectively. 
For instance segmentation, $Y_i$ includes pixel-level mask $m_k$ annotation beyond category 
and bounding box ones, i.e., $Y_i$$=$$\{b_k, m_k, c_k\}_{k=1}^M$. Under few-shot learning setting, 
these annotations can be grouped into two sets, base and novel classes, denoted as $C_{base}$ and $C_{novel}$. 
Note that the base and novel classes are non-overlapping (i.e., $C_{base}$$\cap$$C_{novel}$$=$$\emptyset$).

FSOD or FSIS aims to detect/segment novel class instances through training a model based on plenty of labeled 
instances on a set of base classes and a few instances~(e.g, 1, 2, 3 and 5) on each novel class.
Note, FSOD or FSIS only focuses on recognizing novel class instances but ignores base class ones. It is 
impractical from the perspective of many real-world applications because people always expect that a few-shot 
model is capable of not only recognizing novel classes but also remembering base classes. To this 
end, generalized FSOD and FSIS~(abbreviated as gFSOD and gFSIS)~\cite{gidaris2018dynamic,perez2020incremental,fan2021generalized,ganea2021incremental} stresses that a good few-shot learning system should adapt to new tasks~(novel classes) rapidly while maintaining the performance on previous knowledge~(base classes) without forgetting.

As mentioned above, transfer-learning FSOD and FSIS methods mainly consist of two stages, i.e., pre-training on 
base classes and fine-tuning on novel classes. The former is trained
on plenty of labeled instances in $C_{base}$ while the latter only uses a few labeled instances in $C_{novel}$. For 
generalized few-shot learning, the main difference is novel stage training which uses few labeled instances in both $C_{base}$ 
and $C_{novel}$.  Note that the labeled instances are abundant at the base stage, but there are only a few labeled instances at novel stage both FSOD/FSIS and gFSOD/gFSIS. Therefore, 
most instances are unlabeled~(missing labeled) as only $K$ instances are provided at the fine-tuning stage. 
This is, in fact, reasonable from the perspective of data privacy~(incremental learning).

\subsection{Revisiting Object Detection and Instance Segmentation}\label{sec:revist}
We focus on transfer-learning FSOD and FSIS methods because 
they are more simple and effective compared to meta-learning ones. As we know, Faster-RCNN and Mask-RCNN are very 
popular and powerful solutions as two-stage stacking architecture for fully supervised object detection and 
instance segmentation. We firstly describe the two-stage detector and segmenter. In general, 
the first stage is designed to generate class-agnostic region proposals which can be formulated as:
\begin{equation}\label{eq:s1}
  \mathcal {F}_{\rm{s1}}(\theta_{s1};\cdot) = \mathcal{F}_{\rm{ROI}} 	\circ  \mathcal{F}_{\rm{RPN}} \circ  \mathcal{F}_{\rm{EF}}(\cdot),
\end{equation}
where the $\theta_{s1}$ is all the parameters of the first stage.
Specifically, an input image $X_i$ is firstly fed into a backbone network to Extract high-level Features ($\mathcal{F}_{\rm{EF}}$). Then, a Region Proposal Network (RPN) is adopted to generate candidate regions based on these extracted features 
($\mathcal{F}_{\rm{RPN}}$). Finally, all these region proposals are pooled into fixed size feature maps using a Region of Interest~(ROI)
pooling module ($\mathcal{F}_{\rm{ROI}}$) for the following stage. The structure of the second stage varies depending on the specific 
task. For object detection, the ROI features of~\emph{sampled} region proposals will be parallelly fed into two heads for performing box classification 
and regression, that is
\begin{equation}\label{eq:od}
  \mathcal {L}_{\rm{s2}}^{\rm{OD}} = \mathcal{L}_{\rm{CLS}}(\theta_{cls};\cdot) + \mathcal{L}_{\rm{REG}}(\theta_{reg};\cdot),
\end{equation}
where $\theta_{cls}$ and $\theta_{reg}$ are the parameters of classification and regression head, respectively.
Instance segmentation method~(e.g., Mask-RCNN) follows the second stage structure of object detection and applies it to three heads, i.e., 
box classification, box regression and mask segmentation, that is 
\begin{equation}\label{eq:is}
  \mathcal {L}_{\rm{s2}}^{\rm{IS}}(\cdot) = \mathcal{L}_{\rm{CLS}}(\theta_{cls};\cdot) +  \mathcal{L}_{\rm{REG}}(\theta_{reg};\cdot) + \mathcal{L}_{\rm{SEG}}(\theta_{seg};\cdot),
\end{equation}
where $\theta_{seg}$ is parameters of the mask segmentation head.
Given an input image $X_i$ and its corresponding annotation $Y_i$$=$$\{b_k, c_k\}_{k=1}^M$ or $Y_i$$=$$\{b_k, m_k, c_k\}_{k=1}^M$,
Eqs.~\ref{eq:od} and~\ref{eq:is} can be jointly optimized end-to-end by minimizing $\mathcal {L}_{\rm{s2}}^{\rm{OD}}(\cdot)$ and 
$\mathcal {L}_{\rm{s2}}^{\rm{IS}}(\cdot)$, which follows a multi-task learning paradigm. For simplicity, we omit the RPN learning in 
Eqs.~\ref{eq:od} and~\ref{eq:is}.

\textbf{Missing Label Issue.}
We can obtain a powerful model by minimizing Eqs.~\ref{eq:od} and~\ref{eq:is} when all interesting instances are 
completely labeled in a large-scale image dataset $D$. The completeness of annotation at label space is, in fact, a general precondition for most fully supervised learning algorithms. 
However, it is not satisfactory for a few-shot learning scenario. The reason is that only a few instances
are manually labeled in given training images, and a lot of potential instances may be presented but unlabeled. From the 
perspective of practical application, it is natural and unavoidable to miss annotations when facing a few-shot setting. As shown in Fig.~\ref{fig:miss} (c), 
we can see that there are at least three instances including two ``person'' and one ``dog'' in this image, but only the ``dog'' instance 
carrying annotation (box, mask, category), and other two ``person'' instances are unlabeled (i.e., missing labels).

In order to quantitatively measure the proportion of missing labeled instances, we compute the average missing rate for each shot on PASCAL VOC and 
MS-COCO benchmark datasets as shown in Fig.~\ref{fig:miss}. 
Firstly, it can be seen that there is a high missing rate on each shot. 
For example, 74.3\% novel class instances potentially present on 3-shot MS-COCO training images, but they are not given any annotations. 
Secondly, the missing rate is further increased under a generalized few-shot setting. 
For example, the missing rate increases nearly two times of few-shot setting on the 
PASCAL VOC. 
Despite some efforts try mining these missing labeled instances in a semi-supervised manner for boosting few-shot 
performance, the fully supervised loss with Eqs.~\ref{eq:od} and~\ref{eq:is} may result in a suboptimal solution when some instances are unlabeled. 
To be best of our knowledge, this is the first to propose the missing label issue in few-shot object detection and instance segmentation.

\textbf{Biased Classification Issue.}
The Eqs.~\ref{eq:od} and~\ref{eq:is} can be optimized using sampled ROI features and their labels as mentioned above. 
The sampling operation performs box matching between ``ROI features'' and ``annotation'', and assigns training labels 
(positive or negative) to the corresponding ROIs. Here, the positive (foreground) ROIs are sampled from object proposals that have 
an IoU overlap with the ground-truth bounding box at a threshold (e.g., 0.5), while negative (background) ROIs are sampled from the 
remaining proposals. The classification head is trained based on these sampled positive and negative ROIs. Different from the 
classification head, the box regression, and mask segmentation head learning are only associated with positive ROIs.  

The annotations of positive ROIs are accurate, e.g., the annotations of the ``dog'' instances in 
Fig.~\ref{fig:miss} (c). However, the sampled negative ROIs may be noisy because of the missing label issue under 
the few-shot setting. For example, those two ``people'' instances  will be assigned to negative labels (i.e., background) 
if they are sampled in Fig.~\ref{fig:miss} (c), according to the above label assignment strategy. This will make the 
standard classification head confused with positive and noisy negative samples.
On one hand, the model is correctly guided to recognize positive objects because all positive samples are accurate. 
On the other hand, the model may be misguided by noisy negative samples and thus incorrectly recognize positive objects as background.
Therefore, the bias classification may happen when meeting the missing label issue, especially in a few-shot scenario. 
Furthermore, it potentially limits the generalization ability to adapt to the novel class quickly and efficiently.

\subsection{Decoupling Classifier to Mitigate the Bias Classification}
\textbf{Standard Classifier.}
We assume that $\vec x\in \mathbb{R} ^{C+1}$ is the predicted logit of a sampled ROI feature obtained from Eq.~\ref{eq:s1} 
and its corresponding class label vector is $\vec y \in \mathbb{R}^{C+1}$, 
where there are $C$ foreground categories and one background class, and
$y_i$ is 1 if the corresponding proposal belongs to the $i$-th category, 0 otherwise. 
Then, we use a softmax function to transform it 
into a probability distribution, that is
\begin{equation}\label{eq:sm}
  \hat p_i = \frac{{\rm{exp}} {(x_i)}}{\sum_t {\rm{exp}}(x_t)}.
\end{equation}
The cross-entropy loss is used as the measurement of the similarity between the ground-truth $\vec y$ and predicted 
distribution $\vec {\hat {p}}$, that is
\begin{equation}\label{eq:ce}
  \mathcal{L}_{\rm{CLS}}  = -\sum_{i=0}^{C} y_i {\rm {log}}(\hat p_i).
\end{equation}
Note that the standard classifier (Eqs.~\ref{eq:sm} and \ref{eq:ce}) may confuse with clear positive and noisy 
negative samples in few-shot scenario.

\textbf{Decoupling Classifier.}
In order to process positive and negative samples differentially, we decouple the standard classifier into two heads, i.e., positive (foreground) head and negative (background) head, which are formulated as 
\begin{equation}\label{eq:de}
  \mathcal{L}_{\rm{CLS}}  = \mathcal{L}_{\rm{CLS}}^{\rm{fg}} + \mathcal{L}_{\rm{CLS}}^{\rm{bg}}.
\end{equation}
Here, the positive and negative heads are responsible for positive and negative samples, respectively.
Considering that the labels of positive samples (foreground) are accurate, we can use cross-entropy loss~(i.e., Eq.~\ref{eq:ce}) for all 
positive instances. The labels of those negative examples may be noisy because of the missing label issue. Therefore, it is not reasonable to employ normal cross-entropy loss for training the negative head. 
Note that these negative examples are generally sampled from those object proposals that have a maximum IoU overlap with the ground truth bounding 
box at an interval $[0, 0.5)$,
and thus we can infer that they may not belong to the ground truth class, although we don't know their true category. 
We expect that the bias classification would be mitigated if the negative head performs learning only between few-shot labeled categories and the background class. 
To this end, we first obtain an image-level multi-label with instance-level few-shot annotation of a training image, and denote it as $\vec m$$=$$[m^0, m^1, \cdots, m^{C-1}, m^{C}]^T$,
where $m^i$ is a binary indicator, and $m^i$ is 1 if the image is labeled with the $i$-th category, 0 otherwise. Note that $m^{C}$$=$$1$ indicates that each image at least contains a background class.
Then, we can obtain a constrained logit $\vec {\bar x}$ conditioned on the $\vec m$, that is
\begin{equation}\label{eq:bglogit}
  \bar {x}_i = m_i x_i.
\end{equation}
Substituting Eq.~\ref{eq:bglogit} into the softmax function Eq.~\ref{eq:sm} yields:
\begin{equation}\label{eq:bgsm}
  {\bar {p}}_i = \frac{{\rm{exp}} {(m_i x_i)}}{\sum_t {\rm{exp}}(m_t x_t)}.
\end{equation}
We compute cross-entropy loss between $\vec {{\bar {p}}}$ and the corresponding ground truth $\vec y^{\rm{bg}}$, that is 
\begin{equation}\label{eq:bgce}
  \mathcal{L}_{\rm{CLS}}^{\rm{bg}} = -\sum_{i=0}^{C} y_i^{\rm{bg}}{\rm {log}}({{\bar p}}_{i}),
\end{equation}
where $y_{C}^{\rm{bg}}$$=$$1$ and $y_{i\neq {C}}^{\rm{bg}}$$=$$0$.

\textbf{Optimization with Decoupling Classifier.}
Considering the joint optimization goal (Eq.~\ref{eq:od} and~\ref{eq:is}) of object detection and instance segmentation, 
the optimal parameters $\Theta$ is determined by minimizing Eq.~\ref{eq:od} and~\ref{eq:is}, where 
$\Theta=\{\theta_{s1}, \theta_{cls}, \theta_{reg}\}$ in object detection, and $\Theta=\{\theta_{s1}, \theta_{cls}, \theta_{reg},  \theta_{seg}\}$ in 
instance segmentation. 
For simplicity, we only consider the optimization for the classification head and omit the 
box regression and mask segmentation head
in the following analysis. $\theta_{cls}$ is updated by a gradient descent step, that is 
\begin{equation}\label{eq:grad}
  \theta_{cls} \leftarrow  \theta_{cls} - \lambda \frac{\partial \mathcal{L}_{\rm{CLS}}} {\partial \theta_{cls}},
\end{equation}
where $\lambda $ is the learning rate.
Note that we have decoupled the standard  classifier into positive and negative learning in Eq.~\ref{eq:de}. 
We firstly analyze the $\theta_{cls}$ optimization for positive head. 
According to the chain rule in Eq.~\ref{eq:sm} and \ref{eq:ce}, we have 
\begin{equation}\label{eq:gradfg}
\frac{\partial \mathcal{L}_{\rm{CLS}}^{\rm{fg}}} {\partial \vec{x}} =\vec{\hat p} - \vec y^{\rm{fg}}.
\end{equation}
Then, the derivative of $L_{\rm{CLS}}^{\rm{fg}}$ with respect to $\theta_{cls}$ is 
\begin{equation}\label{eq:pufg}
  \frac{\partial \mathcal{L}_{\rm{CLS}}^{\rm{fg}}} {\partial {\theta_{cls}}} =(\vec{\hat p} - \vec y^{\rm{fg}}) \frac{\partial \vec x}{\theta_{cls}}.
\end{equation}
Combining the negative head in Eq.~\ref{eq:bgce} and Eqs.~\ref{eq:bgsm} and \ref{eq:bglogit}, we can similarly obtain 
derivative of $L_{\rm{CLS}}^{\rm{bg}}$ with respect to $\theta_{cls}$ is 
\begin{equation}\label{eq:pubg}
  \frac{\partial \mathcal{L}_{\rm{CLS}}^{\rm{bg}}} {\partial {\theta_{cls}}} =\vec m (\vec{\bar p} - \vec y^{\rm{bg}}) \frac{\partial \vec x}{\theta_{cls}},
\end{equation}
where $\vec y^{\rm{bg}}$ is the ground truth label vector of a negative sample. 
Comparing Eqs.~\ref{eq:pufg} with \ref{eq:pubg}, we can 
see that the parameters $\theta_{cls}$ of the classification head will be updated with different ways for positive and negative examples. 
For the positive head, the gradient is updated in each dimension of the class space. 
But for the negative head, the gradient is limited in some special dimensions because of the introduced $\vec m$ and thus the bias classification may be alleviated.

\begin{wrapfigure}{r}{0.5\textwidth}
  \centering
  \vspace{-20pt}
  \subfloat[Positive head]
  {
  \includegraphics[width= 0.24\columnwidth]{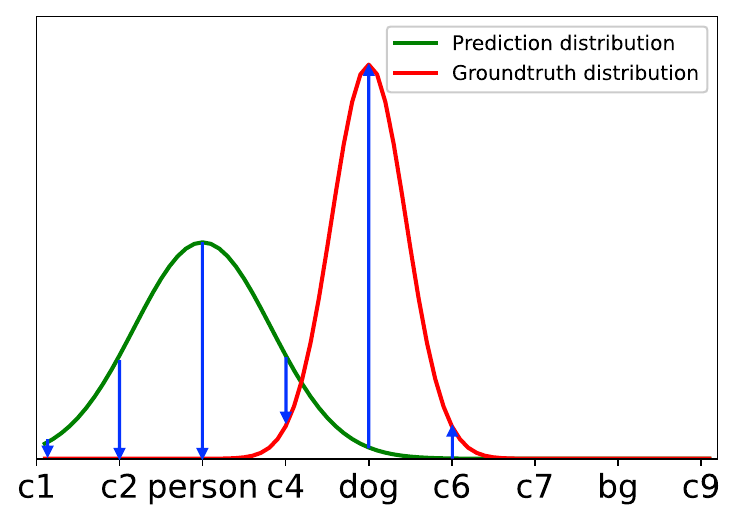}
  }
  \subfloat[Negative head]
  {
  \includegraphics[width= 0.24\columnwidth]{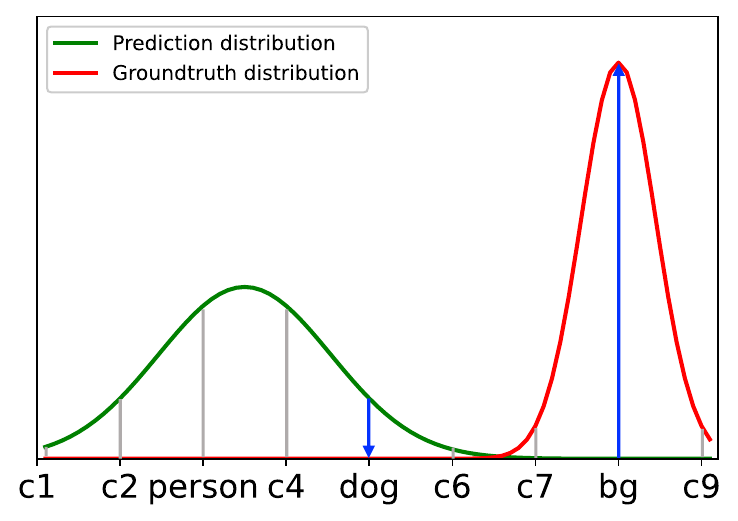}
  }
  \caption{\small{Illustration of the gradient of decoupling classifier, where the blue arrow represents the gradient direction. 
  (a) illustrates the gradient propagation on the positive head, and (b) reveals that the gradient propagation is constrained between few-shot labeled class~(e.g., dog) and the background and thus the bias classification is mitigated. Best viewed in color and zoom in.}} \label{fig:grad}
  \end{wrapfigure}
In order to further understand Eqs.~\ref{eq:pufg} and~\ref{eq:pubg}, we give a visualization example for decoupling classifier as 
illustrated in Fig.~\ref{fig:grad}. Note that we use Gaussian normal distribution for prediction and ground truth distribution for intuition.
Here, we take the one-shot labeled image in Fig.~\ref{fig:miss} (c) for example, where only the ``dog'' instance is labeled. 
We assume that one ``person'' instance is sampled as negative sample, and it will be mistaken as background class (ground truth).
Due to the proposed decoupling classifier, the optimization of the ``person'' instance is constrained between the ``dog'' and 
``background'' as shown in Fig.~\ref{fig:grad} (b) and doesn't affect the predictions of other categories such as the ``person'' class.

\vspace{-10pt}
\section{Experiments}\label{sec:exp}
\vspace{-5pt}
In this section, we empirically evaluate the proposed method for FSOD/gFSOD and FSIS/gFSIS tasks and demonstrate its effectiveness by comparison with state-of-the-art methods.

\subsection{Experimental Setup}\label{sec:setup}
\textbf{Datasets.} We follow the previous works and evaluate our method on PASCAL VOC~\cite{everingham2010pascal} and 
MS-COCO~\cite{lin2014microsoft} datasets. For a fair comparison, we use the same data splits given in~\cite{wang2020tfa,qiao2021defrcn}.

\textbf{PASCAL VOC} covers 20 categories, which are randomly split into 15 base classes and 5 novel classes. There are three such splits in total.
In each class, there are $K$ (1, 2, 3, 5, 10) objects for few-shot training. And the PASCAL VOC 2007 testing set is used for evaluation. 
The dataset is only used to evaluate FSOD task. We report the Average Precision (IoU=0.5) 
for novel classes (AP50).

\textbf{MS-COCO} contains 80 classes. The 20 categories presented in the PASCAL VOC are used as novel classes and the remaining 60 
categories are used as base classes. We train a few-shot model based on $K$ (1, 2, 3, 5, 10, 30) instances for each class and evaluate on the 
MS-COCO validation set. This dataset has been widely used to evaluate the performance of FSOD and FSIS. We report Average Precision
(IoU=0.5:0.95), Average Precision (IoU=0.5) on novel classes for FSOD and FSIS settings. 
In addition, we also report AP and AP50 for overall classes, base classes, and novel classes under gFSOD and gFSIS settings, respectively.

\textbf{Experimental Details.} The experiments are conducted with Detectron2~\cite{wu2019detectron2} on NVIDIA GPU V100 on CUDA 11.0. We use standard Faster-RCNN with 
ResNet-101 backbone extracted features from the final convolutional layer of the 4-th stage for few-shot object detection, which is the 
same as DeFRCN~\cite{qiao2021defrcn}. For instance segmentation, we add a mask prediction head at the ROI of the Faster-RCNN. 
For model training, we employ a two-stage transfer-learning approach: first training the network on the base classes with pre-trained
by ImageNet and then fine-tuning on $K$-shots for every class. The SGD is used to optimize our network end-to-end with a mini-batch 
size of 16, momentum 0.9, and weight decay $5e^{-5}$ on 8 GPUs. The learning rate is set to 0.02 during base training and 
0.01 during few-shot fine-tuning. Following the previous work~\cite{wang2020tfa}, all experimental results are averaged over 10 seeds. 
For a fair comparison with DeFRCN~\cite{qiao2021defrcn}, we also report the average results over 10 times repeated runs on \texttt{seed0}.

\textbf{Strong Baseline.} For FSOD and gFSOD, we take the state-of-the-art DeFRCN~\cite{qiao2021defrcn} as a strong baseline of our method. 
For FSIS and gFSIS, we extend DeFRCN similarly to how Mask-RCNN extends Faster-RCNN, i.e., adding a mask prediction head at 
the ROI of the DeFRCN and keeping others as same as DeFRCN, and thus we call it Mask-DeFRCN. 
Our method only replaces the standard classifier in DeFRCN and Mask-DeFRCN with our decoupling classifier. 
Therefore, DeFRCN and Mask-DeFRCN can be taken as our strong baseline for FSOD/gFSOD and FSIS/gFSIS tasks.

\subsection{Comparison with the State-of-the-Art}
\textbf{Few-shot Instance Segmentation on the MS-COCO.}
\emph{Our method (simple decoupling classifier) outperforms the state-of-the-art on the MS-COCO in both FSIS and gFSIS settings.} 
The main results on MS-COCO are reported in Table~\ref{tab:fsis-coco} and~\ref{tab:gfsis-coco} for FSIS and gFSIS, respectively.
Based on the experiment results, we have the following observations: 
\textbf{1)} The strong baseline (i.e., Mask-DeFRCN) outperforms all the state-of-the-art methods for FSIS and gFSIS;
\textbf{2)} Our method consistently outperforms the baseline Mask-DeFRCN for FSIS; Compared to FSIS, our method has significant improvements for gFSIS. 
This is not surprised because the missing rate of gFSIS always is higher than that of FSIS as shown in Fig.~\ref{fig:miss} (b). This indicates 
our method is capable of addressing the missing label issue under few-shot setting;
\textbf{3)} Our method has a better advantage, especially in low-shot~(1-, 2-, 3-shot), and thus it is very suitable for a few-shot scenario. 
\begin{table}[t]
  \caption{\small{FSIS performance for Novel classes on MS-COCO. 
  The superscript~$^\dagger$ indicates that the results are our re-implementation.
  The results are averaged over all 10 seeds and the best ones are in bold, the same below.}}\label{tab:fsis-coco}
  \vspace{-2pt}
  \centering
  \resizebox{1.0\textwidth}{!}{
  \begin{tabular}{lr|c|rr|rr|rr|rr|rr|rr}
         \toprule
   \multirow{2}{*}{Methods} &\multirow{2}{*}{} &\multirow{2}{*}{Tasks}      
   &\multicolumn{2}{c|}{{1}} &\multicolumn{2}{c|}{{2}} &\multicolumn{2}{c|}{{3}} 
   &\multicolumn{2}{c|}{{5}} &\multicolumn{2}{c|}{{10}} &\multicolumn{2}{c}{{30}} \\
    \cmidrule(r){4-15}
     &     &
     &\multicolumn{1}{c}{\textbf{AP}} &\multicolumn{1}{c|}{\textbf{AP50}} 
     &\multicolumn{1}{c}{\textbf{AP}} &\multicolumn{1}{c|}{\textbf{AP50}} 
     &\multicolumn{1}{c}{\textbf{AP}} &\multicolumn{1}{c|}{\textbf{AP50}} 
     &\multicolumn{1}{c}{\textbf{AP}} &\multicolumn{1}{c|}{\textbf{AP50}} 
     &\multicolumn{1}{c}{\textbf{AP}} &\multicolumn{1}{c|}{\textbf{AP50}} 
     &\multicolumn{1}{c}{\textbf{AP}} &\multicolumn{1}{c}{\textbf{AP50}} \\
     \toprule
      Meta R-CNN~\cite{yan2019metarcnn}     &\emph{ICCV 19}     & \multirow{5}{*}{Det}   &- &-   &- &- &- &- &3.5&9.9 &5.6&14.2    &- &-\\                  
    MTFA~\cite{ganea2021incremental}      &\emph{CVPR 21}  &      &2.47   &4.85       &- &- &- &- &6.61   &12.32  &8.52   &15.53  &- &- \\
    iMTFA~\cite{ganea2021incremental}     &\emph{CVPR 21}  &      &3.28   &6.01      &- &- &- &- &6.22   &11.28    &7.14   &12.91 &- &-  \\
     Mask-DeFRCN$^\dagger$ \cite{qiao2021defrcn}    &\emph{ICCV 21}     & \multirow{3}{*}{Det}  
     &7.54    &14.46
     &11.01  &20.20
     &13.07  &23.28 
     &15.39  &27.29
     &18.72  &32.80  
     &22.63  &38.95  \\                         
     \rowcolor[HTML]{EFEFEF}
     \textbf{Ours}                          &      &
     &\textbf{8.09}   &\textbf{15.85} 
     &\textbf{11.90}    &\textbf{22.39}  
     &\textbf{14.04}    &\textbf{25.74}
     &\textbf{16.39}   &\textbf{29.96} 
     &\textbf{19.33}    &\textbf{34.78}  
     &\textbf{22.73}    &\textbf{40.24}    \\
    \toprule
     Meta R-CNN~\cite{yan2019metarcnn}       &\emph{ICCV 19}    & \multirow{5}{*}{Seg}    &- &-    &- &- &- &- &2.8&6.9  &4.4&10.6 &- &- \\
    MTFA~\cite{ganea2021incremental}       &\emph{CVPR 21} &                        &2.66 &4.56           &- &- &- &-&6.62   &11.58  &8.39   &14.64 &- &- \\
    iMTFA~\cite{ganea2021incremental}      &\emph{CVPR 21}  &                       &2.83 &4.75     &- &- &- &-&5.24   &8.73    &5.94   &9.96  &- &-  \\    
    Mask-DeFRCN$^\dagger$ \cite{qiao2021defrcn}     &\emph{ICCV 21}                       & \multirow{3}{*}{Seg}                         
    &6.69   &13.24  
    &9.51 &18.58  
    &11.01 &21.27  
    &12.66 &24.58
    &15.39 &29.71 
    &18.28 &35.20 \\
     \rowcolor[HTML]{EFEFEF}
     \textbf{Ours}                            & &
     &\textbf{7.18}       &\textbf{14.33} 
     &\textbf{10.31}      &\textbf{20.43}  
     &\textbf{11.85}     &\textbf{23.24}  
     &\textbf{13.48}    &\textbf{26.67} 
     &\textbf{15.85}      &\textbf{31.33}
     &\textbf{18.34}      &\textbf{35.99}   \\
      \bottomrule
  \end{tabular}}
  \vspace{-12pt}
\end{table}

\begin{table}[t]
  \caption{\small{gFSIS performance for Overall, Base and Novel classes on MS-COCO.}}
  \label{tab:gfsis-coco}
  \vspace{-2pt}
  \centering
  \resizebox{1.0\textwidth}{!}{
  \begin{tabular}{l|c|rr|rr|rr|rr|rr|rr}
    \toprule
       \multirow{4}{*}{Shots}        &\multirow{4}{*}{Methods}        
       &\multicolumn{6}{c|}{\textbf{Object Detection}} &\multicolumn{6}{c}{\textbf{Instance Segmentation}} \\
    \cmidrule(r){3-14}
         &     
        &\multicolumn{2}{c|}{\textbf{Overall}} &\multicolumn{2}{c|}{\textbf{Base}} &\multicolumn{2}{c|}{\textbf{Novel}} 
        &\multicolumn{2}{c|}{\textbf{Overall}} &\multicolumn{2}{c|}{\textbf{Base}} &\multicolumn{2}{c}{\textbf{Novel}}\\
        \cmidrule(r){3-14}
        &     
        &\multicolumn{1}{c}{\textbf{AP}} &\multicolumn{1}{c|}{\textbf{AP50}} 
        &\multicolumn{1}{c}{\textbf{AP}} &\multicolumn{1}{c|}{\textbf{AP50}} 
        &\multicolumn{1}{c}{\textbf{AP}} &\multicolumn{1}{c|}{\textbf{AP50}} 
        &\multicolumn{1}{c}{\textbf{AP}} &\multicolumn{1}{c|}{\textbf{AP50}} 
        &\multicolumn{1}{c}{\textbf{AP}} &\multicolumn{1}{c|}{\textbf{AP50}} 
        &\multicolumn{1}{c}{\textbf{AP}} &\multicolumn{1}{c}{\textbf{AP50}} \\

    \midrule
    &Base-Only                && &39.86 &59.25 &&    &&  &32.58 &55.12  && \\
    \midrule
   \multirow{3}{*}{1} &iMTFA~\cite{ganea2021incremental}  &21.67 &31.55 &27.81 &40.11 &3.23 &5.89 &20.13 &30.64 &25.90 &39.28 &2.81 &4.72\\
    &Mask-DeFRCN$^\dagger$ \cite{qiao2021defrcn}
         &23.82      & 35.70      & 30.11      & 44.42      & 4.95      & 9.55       & 19.58      & 33.38      & 24.63      & 41.57      & 4.45      & 8.81\\
    \rowcolor[HTML]{EFEFEF}
    \cellcolor{White}{} &\textbf{Ours}       
         &\textbf{27.35}    &\textbf{42.55}     &\textbf{34.35}      &\textbf{52.46}      &\textbf{6.34}      &\textbf{12.79}      &\textbf{22.45}      &\textbf{39.33}      &\textbf{28.03}      &\textbf{48.60}      &\textbf{5.72}      &\textbf{11.53}\\
    \hline
    \multirow{2}{*}{2} 
    &Mask-DeFRCN$^\dagger$ \cite{qiao2021defrcn}
         &25.42     &38.31      & 31.06      & 45.82      & 8.52       & 15.79      & 21.09      & 35.92      & 25.61      & 43.03      & 7.54       & 14.59      \\
    &\cellcolor{Gray}{\textbf{Ours}} 
         &\cellcolor{Gray}{\textbf{28.63}}      &\cellcolor{Gray}{\textbf{44.74}}      &\cellcolor{Gray}{\textbf{34.67}}     &\cellcolor{Gray}{\textbf{52.82}}      
         &\cellcolor{Gray}{\textbf{10.52}}      &\cellcolor{Gray}{\textbf{20.49}}      &\cellcolor{Gray}{\textbf{23.73}}     &\cellcolor{Gray}{\textbf{41.49}}      
         &\cellcolor{Gray}{\textbf{28.52}}      &\cellcolor{Gray}{\textbf{49.12}}      &\cellcolor{Gray}{\textbf{9.38}}      &\cellcolor{Gray}{\textbf{18.62}}  \\
         \hline
         \multirow{2}{*}{3} 
    &Mask-DeFRCN$^\dagger$ \cite{qiao2021defrcn}
         &26.54      & 40.01      & 31.77      & 46.83      & 10.87      & 19.55      & 22.04      & 37.48      & 26.22      & 43.95      & 9.48       & 18.06      \\
         &\cellcolor{Gray}{\textbf{Ours}} 
         &\cellcolor{Gray}{\textbf{29.59}}      &\cellcolor{Gray}{\textbf{46.21}}     &\cellcolor{Gray}{\textbf{35.07}}      &\cellcolor{Gray}{\textbf{53.30}}      
         &\cellcolor{Gray}{\textbf{13.15}}      &\cellcolor{Gray}{\textbf{24.95}}     &\cellcolor{Gray}{\textbf{24.55}}      &\cellcolor{Gray}{\textbf{42.81}}      
         &\cellcolor{Gray}{\textbf{28.91}}      &\cellcolor{Gray}{\textbf{49.61}}     &\cellcolor{Gray}{\textbf{11.46}}      &\cellcolor{Gray}{\textbf{22.43}}      \\
         \hline
    \multirow{3}{*}{5} &iMTFA~\cite{ganea2021incremental}   
         &19.62 &28.06 &24.13 &33.69 &6.07 &11.15 &18.22 &27.10 &22.56 &33.25 &5.19 &8.65\\
    &Mask-DeFRCN$^\dagger$ \cite{qiao2021defrcn}
        &27.82      & 42.12      & 32.54      & 48.03      & 13.69      & 24.41      & 23.03      & 39.37      & 26.84      & 45.04      & 11.60      & 22.36  \\
    \rowcolor[HTML]{EFEFEF}
    \cellcolor{White}{}  &\textbf{Ours}       
        &\textbf{30.48}     &\textbf{47.75}      &\textbf{35.30}      &\textbf{53.65}      &\textbf{16.02}      &\textbf{30.05}      &\textbf{25.20}      &\textbf{44.12}      &\textbf{29.10}      &\textbf{49.87}      &\textbf{13.50}      &\textbf{26.86}\\
    \hline
    \multirow{3}{*}{10} &iMTFA~\cite{ganea2021incremental}   
         &19.26 &27.49 &23.36 &32.41 &6.97 &12.72 &17.87 &26.46 &21.87 &32.01 &5.88 &9.81 \\
    &Mask-DeFRCN$^\dagger$ \cite{qiao2021defrcn}
         &29.88    &45.25    &34.17    &50.48    &17.02    &29.58    &24.75    &42.32    &28.23    &47.33    &14.32    &27.29    \\
         &\cellcolor{Gray}{\textbf{Ours}}       
         &\cellcolor{Gray}{\textbf{31.77}}      &\cellcolor{Gray}{\textbf{49.77}}     &\cellcolor{Gray}{\textbf{36.14}}      &\cellcolor{Gray}{\textbf{54.85}}      
         &\cellcolor{Gray}{\textbf{18.67}}      &\cellcolor{Gray}{\textbf{34.55}}     &\cellcolor{Gray}{\textbf{26.36}}      &\cellcolor{Gray}{\textbf{46.13}}      
         &\cellcolor{Gray}{\textbf{29.91}}      &\cellcolor{Gray}{\textbf{51.11}}     &\cellcolor{Gray}{\textbf{15.71}}      &\cellcolor{Gray}{\textbf{31.19}}  \\
         \hline

         \multirow{2}{*}{30}
    &Mask-DeFRCN$^\dagger$ \cite{qiao2021defrcn}
         &31.66      & 48.11      & 35.10      & 52.01      & 21.33      & 36.44      & 26.23      & 44.97      & 29.12      & 48.82      & 17.57      & 33.42      \\
    &\cellcolor{Gray}{\textbf{Ours}}
         &\cellcolor{Gray}{\textbf{32.92}}     &\cellcolor{Gray}{\textbf{51.37}}      &\cellcolor{Gray}{\textbf{36.45}}      &\cellcolor{Gray}{\textbf{55.05}}      
         &\cellcolor{Gray}{\textbf{22.30}}     &\cellcolor{Gray}{\textbf{40.31}}      &\cellcolor{Gray}{\textbf{27.31}}      &\cellcolor{Gray}{\textbf{47.61}}    
         &\cellcolor{Gray}{\textbf{30.32}}     &\cellcolor{Gray}{\textbf{51.41}}      &\cellcolor{Gray}{\textbf{18.29}}      &\cellcolor{Gray}{\textbf{36.22}}     \\  
    \bottomrule
  \end{tabular}}
  \vspace{-15pt}
\end{table}
\begin{table}
  \centering
  \caption{FSOD and gFSOD performance ($\rm{AP}_{50}$) for Novel classes on PASCAL VOC. The term \textit{w/g} indicates whether we use the {gFSOD} 
  setting~\cite{wang2020tfa}. The superscript~$*$ indicates that the results are averaged over 10 times repeated runs on \texttt{seed0}, the same below.
  }\label{tab:fsod-voc}
  \vspace{-2pt}
	\resizebox{1.0\textwidth}{!}{
	\begin{tabular}{lr|c|rrrrr|rrrrr|rrrrr}
			\toprule[1.1pt]
			& & \multicolumn{1}{c|}{  }  & \multicolumn{5}{c|}{Novel Set 1} & \multicolumn{5}{c|}{Novel Set 2} & \multicolumn{5}{c}{Novel Set 3}    \\
      {\multirow{-2}{*}{Methods / Shots}} & & \multirow{-2}{*}{\textit{w/g}} 
      &\multicolumn{1}{c}{1}&\multicolumn{1}{c}{2}&\multicolumn{1}{c}{3}&\multicolumn{1}{c}{5}&\multicolumn{1}{c|}{10}    
      &\multicolumn{1}{c}{1}&\multicolumn{1}{c}{2}&\multicolumn{1}{c}{3}&\multicolumn{1}{c}{5}&\multicolumn{1}{c|}{10}      
      &\multicolumn{1}{c}{1}&\multicolumn{1}{c}{2}&\multicolumn{1}{c}{3}&\multicolumn{1}{c}{5}&\multicolumn{1}{c}{10}  \\ 
      \midrule[0.9pt]
			{FRCN-ft \cite{yan2019metarcnn}}  &\emph{ICCV 19}     & \xmark         & 13.8          & 19.6          & 32.8          & 41.5          & 45.6          & 7.9           & 15.3          & 26.2          & 31.6          & 39.1          & 9.8           & 11.3          & 19.1          & 35.0          & 45.1          \\
			{FSRW \cite{kang2019frsw}}  &\emph{ICCV 19}         & \xmark         & 14.8          & 15.5          & 26.7          & 33.9          & 47.2          & 15.7          & 15.2          & 22.7          & 30.1          & 40.5          & 21.3          & 25.6          & 28.4          & 42.8          & 45.9          \\
			{MetaDet \cite{wang2019metalearn}}  &\emph{ICCV 19}     & \xmark         & 18.9          & 20.6          & 30.2          & 36.8          & 49.6          & {21.8}        & 23.1          & 27.8          & 31.7          & 43.0          & 20.6          & 23.9          & 29.4          & 43.9          & 44.1          \\
			{MetaRCNN \cite{yan2019metarcnn}}  &\emph{ICCV 19}   & \xmark         & 19.9          & 25.5          & 35.0          & 45.7          & 51.5          & 10.4          & 19.4          & 29.6          & 34.8          & 45.4          & 14.3          & 18.2          & 27.5          & 41.2          & 48.1          \\
			{TFA  \cite{wang2020tfa}} &\emph{ICML 20}& \xmark         & 39.8          & 36.1          & 44.7          & {55.7}        & 56.0          & 23.5          & {26.9}        & 34.1          & 35.1          & 39.1          & 30.8          & {34.8}        & {42.8}        & {49.5}        & {49.8}        \\
			{MPSR  \cite{wu2020multi}}   &\emph{ECCV 20}       & \xmark         & 41.7          & -             & {51.4}        & 55.2          &61.8           & {24.4}        &-              & {39.2}        & {39.9}        & {47.8}        & {35.6}        & -             & 42.3          & 48.0          & 49.7          \\
      {TIP} \cite{li2021tip}                        &\emph{CVPR 21}       & \xmark         &27.7 &36.5 &43.3 &50.2 &59.6 &22.7 &30.1 &33.8 &40.9 &46.9 &21.7 &30.6 &38.1 &44.5 &50.9 \\
      {DCNet} \cite{hu2021dcnet}                   &\emph{CVPR 21}       & \xmark         &33.9 &37.4 &43.7 &51.1 &59.6 &23.2 &24.8 &30.6 &36.7 &46.6 &32.3 &34.9 &39.7 &42.6 &50.7 \\
      {CME}   \cite{li2021cme}                   &\emph{CVPR 21}       & \xmark         &41.5 &47.5 &50.4 &58.2 &60.9 &27.2 &30.2 &41.4 &42.5 &46.8 &34.3 &39.6 &45.1 &48.3 &51.5 \\
      {FSCE}     \cite{sun2021fsce}                  &\emph{CVPR 21}       & \xmark         &44.2           &43.8           &51.4 &61.9 &63.4 &27.3 &29.5 &43.5 &44.2 &50.2 &37.2 &41.9 &47.5 &54.6 &58.5 \\
      SRR-FSD \cite{zhu2021semantic}  &\emph{CVPR 21}  & \xmark &47.8 &50.5 &51.3 &55.2 &56.8 &32.5 &35.3 &39.1 &40.8 &43.8 &40.1 &41.5 &44.3 &46.9 &46.4 \\
      FADI \cite{cao2021fadi} &\emph{NeurIPS 21}  & \xmark &\textbf{50.3} &54.8 &54.2 &59.3 &63.2 &30.6 &35.0 &40.3 &42.8 &48.0 &45.7 &49.7 &49.1 &55.0 &59.6 \\
      {FCT}  \cite{han2022ct}              &\emph{CVPR 22}               & \xmark         &38.5 &49.6 &53.5 &59.8 &64.3 &25.9 &34.2 &40.1 &44.9 &47.4 &34.7 &43.9 &49.3 &53.1 &56.3 \\
			{DeFRCN$^\dagger$ \cite{qiao2021defrcn}} &\emph{ICCV 21} &\xmark   &46.2 &56.4 &59.3 &62.4 &63.7 &32.6 &39.9 &\textbf{44.5} &\textbf{48.3} &\textbf{51.8} &39.8 &49.9 &52.6 &56.1 &59.7  \\
			\rowcolor[HTML]{EFEFEF}
			{\textbf{Ours}} &  &\xmark &{46.2} &\textbf{57.4} &\textbf{59.9} &\textbf{62.9} &\textbf{64.5} &\textbf{32.6} &\textbf{39.9} &43.4 &47.9 &51.3 &\textbf{40.3} &\textbf{50.5} &\textbf{53.8} &\textbf{56.9} &\textbf{60.7}\\  
      \cmidrule(r){1-18}
      {DeFRCN {*} \cite{qiao2021defrcn}} &\emph{ICCV 21} & \xmark         &{53.6}         &57.5           & {61.5}        & {64.1}        &{60.8}         &\textbf{30.1}  & {38.1}        &\textbf{47.0}  &\textbf{53.3}  & {47.9}        &\textbf{48.4}  & {50.9}        & {52.3}        & {54.9}        & 57.4          \\ 
			\rowcolor[HTML]{EFEFEF}
			{\textbf{Ours} *}                            &       &\xmark   &\textbf{56.6} &\textbf{59.6} &\textbf{62.9} &\textbf{65.6} &\textbf{62.5} 	&29.7 &\textbf{38.7} &{46.2} &48.9 &\textbf{48.1} &47.9 &\textbf{51.9} &\textbf{53.3} &\textbf{56.1} &\textbf{59.4}\\
      
			\midrule[0.9pt]
      \midrule[0.9pt]
			{FRCN-ft \cite{yan2019metarcnn}} &\emph{ICCV 19}    & \cmark     & 9.9         & 15.6         & 21.6          & 28.0          & 52.0          & 9.4           & 13.8          & 17.4          & 21.9          & 39.7          & 8.1           & 13.9          & 19.0          & 23.9          & 44.6          \\
			{FSRW \cite{kang2019frsw}}    & \emph{ICCV 19}      & \cmark     & 14.2        & 23.6         & 29.8          & 36.5          & 35.6          & 12.3          & 19.6          & 25.1          & 31.4          & 29.8          & 12.5          & 21.3          & 26.8          & 33.8          & 31.0          \\
			
			{TFA \cite{wang2020tfa}} &\emph{ICML 20} & \cmark         &{25.3}         &{36.4}         & 42.1          & 47.9          & 52.8          & 18.3          & {27.5}          & 30.9          & 34.1          & 39.5          & 17.9          & 27.2          & 34.3          & 40.8          & 45.6          \\
			{FSDetView \cite{Xiao2020FSDetView}}&\emph{ECCV 20} & \cmark      & 24.2          & 35.3          &{42.2}          & {49.1}          & {57.4}          & {21.6}          & 24.6          & {31.9}          &{37.0}          & {45.7}         & {21.2}          & {30.0}          & {37.2}          & {43.8}          & {49.6}  \\
			{DeFRCN \cite{qiao2021defrcn}} & \emph{ICCV 21}  & \cmark         &40.2           &{53.6}         & {58.2} & {63.6} & {66.5} & {29.5} & {39.7} & {43.4} & {48.1} & {52.8} & {35.0} & {38.3} & {52.9} & {57.7} & {60.8} \\ 
			\rowcolor[HTML]{EFEFEF}
			{\textbf{Ours}}   &  &\cmark  &\textbf{45.8} &\textbf{59.1} &\textbf{62.1} &\textbf{66.8} &\textbf{68.0}  &\textbf{31.8} &\textbf{41.7} &\textbf{46.6} &\textbf{50.3} &\textbf{53.7} &\textbf{39.6} &\textbf{52.1} &\textbf{56.3} &\textbf{60.3} &\textbf{63.3} \\
			\bottomrule
\end{tabular}}
  \vspace{-15pt}
\end{table}

\begin{table}[t] \centering 
  \caption{\small{gFSOD performance ($\rm{AP}$) for \textbf{O}verall, \textbf{B}ase and \textbf{N}ovel classes on MS-COCO. 
  }}\label{tab:gfsod-coco}
  \vspace{-2pt}
  \resizebox{1.0\textwidth}{!}{
  \begin{tabular}{l|rrr|rrr|rrr|rrr|rrr|rrr}
      \toprule[1.1pt]
      \multirow{2}{*}{Method / Shots}  &\multicolumn{3}{c|}{1}       &\multicolumn{3}{c|}{2}             &\multicolumn{3}{c|}{3}            &\multicolumn{3}{c|}{5}            &\multicolumn{3}{c|}{10}    &\multicolumn{3}{c}{30}          \\ 
      \cmidrule(r){2-19}
      &\multicolumn{1}{c}{\textbf{O}} &\multicolumn{1}{c}{\textbf{B}}&\multicolumn{1}{c|}{\textbf{N}} &\multicolumn{1}{c}{\textbf{O}} &\multicolumn{1}{c}{\textbf{B}} &\multicolumn{1}{c|}{\textbf{N}} &\multicolumn{1}{c}{\textbf{O}} &\multicolumn{1}{c}{\textbf{B}} &\multicolumn{1}{c|}{\textbf{N} }&\multicolumn{1}{c}{\textbf{O}} &\multicolumn{1}{c}{\textbf{B}} &\multicolumn{1}{c|}{\textbf{N}} &\multicolumn{1}{c}{\textbf{O}} &\multicolumn{1}{c}{\textbf{B}} &\multicolumn{1}{c|}{\textbf{N}} &\multicolumn{1}{c}{\textbf{O}} &\multicolumn{1}{c}{\textbf{B}} &\multicolumn{1}{c}{\textbf{N}}\\
      \midrule[0.9pt]
      FRCN-ft \cite{yan2019metarcnn}           &16.2&21.0& 1.7         &15.8&20.0&3.1  &15.0&18.8& 3.7         &14.4&17.6& 4.6          &13.4&16.1& 5.5         &13.5&15.6& 7.4                  \\
      TFA \cite{wang2020tfa}     &24.4&31.9& 1.9         &24.9&31.9&3.9  &25.3&32.0& 5.1         &25.9&41.2& 7.0         &26.6&32.4& 9.1          &28.7&34.2& 12.1                 \\
      FSDetView \cite{Xiao2020FSDetView}   &&&3.2                  &&&4.9          &&&6.7                  && &8.1                  &&&10.7                 &&&15.9          \\
      DeFRCN \cite{qiao2021defrcn}         &24.4&30.4&4.8          &25.7&31.4&8.5  &26.6&32.1&10.7         &27.8&32.6&13.6          &29.7&34.0&16.8         &31.4&34.8&21.2  \\ 
      \rowcolor[HTML]{EFEFEF}
      \textbf{Ours}                                &\textbf{27.4}&\textbf{34.4}&\textbf{6.2} &\textbf{28.6}&\textbf{34.7}&\textbf{10.4} &\textbf{29.4}&\textbf{34.9}&\textbf{12.9} &\textbf{30.2}&\textbf{35.0}&\textbf{15.7} &\textbf{31.4}&\textbf{35.7}&\textbf{18.3} &\textbf{32.3}&\textbf{35.8}&\textbf{21.9}  \\ 
      \bottomrule[1.1pt]
  \end{tabular}
  }
  \vspace{-12pt}
\end{table}

On one hand, the missing rate of low-shot is generally higher than that of high-shot so that it leaves more improvement space for our method. 
On the other hand, common few-shot models may be weak against noisy negative samples when the number of positive training samples is very small under few-shot conditions.  
In contrast, our method is designed to deal with noisy negative samples issue, and thus it is more effective. In short, the proposed decoupling classifier is a promising 
approach to cope with the missing labels issue for FSIS/gFSIS.

\begin{wraptable}{r}{0.5\textwidth}
    \vspace{-2pt}
    \caption{\small{FSOD performance ($\rm{AP}$) for Novel classes on MS-COCO. 
    The superscripts $\underline{\rm{x}}$ indicate that the results are reported in DeFRCN~\cite{qiao2021defrcn}.}}\label{tab:fsod-coco}
    \resizebox{0.5\textwidth}{!}
    {\begin{tabular}{lr|crrrrrr}
        \toprule[1.1pt]
        {Methods / Shots}  && 1        & 2              & 3             & 5             & 10    &30                \\ \midrule[0.9pt]
        {FRCN-ft \cite{yan2019metarcnn}}         &\emph{ICCV 19}  &\underline{1.0}     &\underline{1.8}     &\underline{2.8}          &\underline{4.0}          & 6.5          &11.1            \\
        {FSRW \cite{kang2019frsw}}            &\emph{ICCV 19}       & -    & -      & -          & -          & 5.6          & 9.1                  \\
        {MetaDet \cite{wang2019metalearn}}      &\emph{ICCV 19}         & -     & -     & -          & -          & 7.1          & 11.3                  \\
        {MetaRCNN \cite{yan2019metarcnn}}     &\emph{ICCV 19}        & -     & -     & -          & -          &  8.7         & 12.4                  \\
        {TFA  \cite{wang2020tfa}}          &\emph{ICML 20}    &\underline{4.4}  &\underline{5.4}       &\underline{6.0}          &\underline{7.7}          &10.0         & 13.7                 \\
        {MPSR  \cite{wu2020multi}}          &\emph{ECCV 20}    &\underline{5.1}  &\underline{6.7}       &\underline{7.4}          &\underline{8.7}          &9.8         & 14.1                 \\
        {FSDetView \cite{Xiao2020FSDetView}}  &\emph{ICCV 20}           &4.5  &6.6       &7.2          &10.7         &12.5     &14.7           \\
        {TIP}   \cite{li2021tip}              &\emph{CVPR 21}   &- &- &- &- &16.3 &18.3 \\
        {DCNet}   \cite{hu2021dcnet}          &\emph{CVPR 21}     &- &- &- &- &12.8 &18.6 \\
        {CME}    \cite{li2021cme}           &\emph{CVPR 21}  &- &- &- &- &15.1 &16.9 \\
        {FSCE}     \cite{sun2021fsce}       &\emph{CVPR 21}  &- &- &- &- &11.1 &15.3 \\
        SRR-FSD \cite{zhu2021semantic}  &\emph{CVPR 21}   &- &- &- &- &11.3 &14.7 \\
        {FADI}    \cite{cao2021fadi}       &\emph{NeurIPS 21}       &5.7 &7.0 &8.6 &10.1 &12.2 &16.1 \\
        {FCT}     \cite{han2022ct}         &\emph{CVPR 22}       &5.1 &7.2 &9.8 &12.0 &15.3 &20.2 \\
        
        {DeFRCN$^\dagger$ \cite{qiao2021defrcn}} &\emph{ICCV 21}   & 7.7 &11.4 &13.3 &15.5 &18.5 &22.5 \\ 
        \rowcolor[HTML]{EFEFEF}
        {\textbf{Ours}}  &  &\textbf{8.1}  &\textbf{12.1} &\textbf{14.4} &\textbf{16.6} &\textbf{19.5} &\textbf{22.7} \\ 
        \cmidrule(r){1-8}
        {DeFRCN * \cite{qiao2021defrcn}}  &\emph{ICCV 21}   & 9.3 &12.9 &\textbf{14.8} &\textbf{16.1} &\textbf{18.5} &\textbf{22.6} \\ 
        \rowcolor[HTML]{EFEFEF}
        {\textbf{Ours} *}  &  &\textbf{10.0} &\textbf{13.6} &14.7 & 15.7 &18.0 &22.2 \\ 
        \bottomrule[1.1pt]
        \vspace{-35pt}
    \end{tabular}}
  \end{wraptable}

  \textbf{Few-shot Object Detection on the PASCAL VOC and MS-COCO.}
  \emph{Our method significantly outperforms the state-of-the-art few-shot object detection methods by a large margin 
  both on the PASCAL VOC and MS-COCO datasets under gFSOD setting again. For the FSOD  setting, our method is also better 
  than the state-of-the-art under most cases.} The results on the PASCAL VOC and MS-COCO are reported 
  Tables~\ref{tab:fsod-voc}, \ref{tab:fsod-coco} and \ref{tab:gfsod-coco}, respectively.
 Some interesting observations are summarized as follows:
  \textbf{1)} Our method significantly and consistently exceeds the current state-of-the-art DeFRCN under 
  the gFSOD setting both on the PASCAL VOC and MS-COCO, which is similar to that of the gFSIS;
  \textbf{2)} Our method is better than the strong DeFRCN in all shots on the MS-COCO and in most cases 
  on the PASCAL VOC under the FSOD setting (averaging the results on 10 seeds). It is worth noting that 
  our method's performance on the MS-COCO is close~(maybe slightly worse, about 0.5\%) to that of DeFRCN 
  if we compare the results based on 10 times repeated runs on the \texttt{seed0}. We recheck the missing 
  rate of the \texttt{seed0}, and find that the corresponding missing rate is significantly 
  lowered (even zero) than that of the other 9 seeds. 
  This also further indicates that our method is robust when the missing rate is small even zero.

\subsection{Ablation Study and Analysis}
We conduct the ablation study to analyze the component of our method. Models in this section are based on the gFSIS setting~(1-, 5-, 10-shot) 
using MS-COCO. Note that the DeFRCN uses a Prototypical Calibration Block~(PCB) to refine the classification score which is effective 
for improving the FSOD performance, but this brings additional computation cost. Therefore, we consider these two factors including decoupling classifier (DC) head 
and PCB in the following analysis.

\begin{figure}[h]
  \vspace{-15pt}
  \centering
  \subfloat[FSIS]
  {
  \includegraphics[width= 0.40\columnwidth]{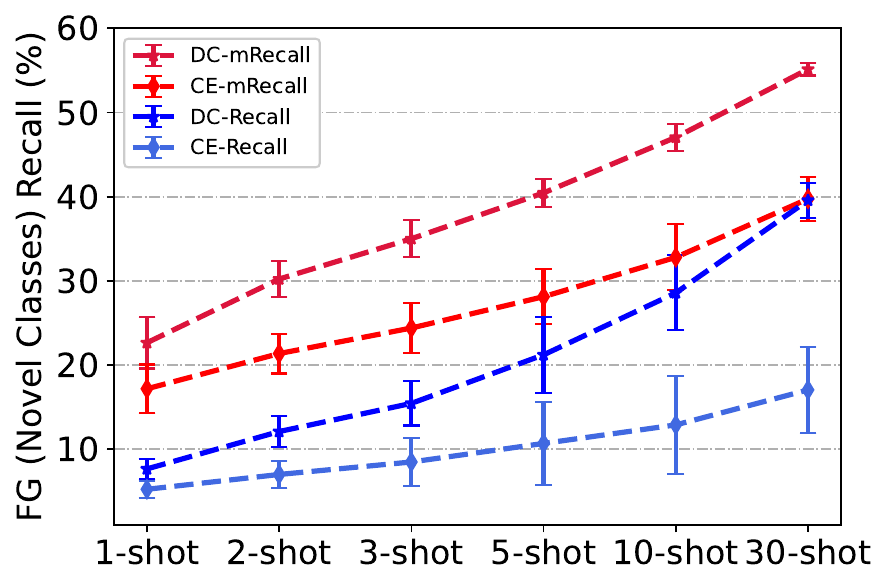}
  }
  \subfloat[gFSIS]
  {
  \includegraphics[width= 0.40\columnwidth]{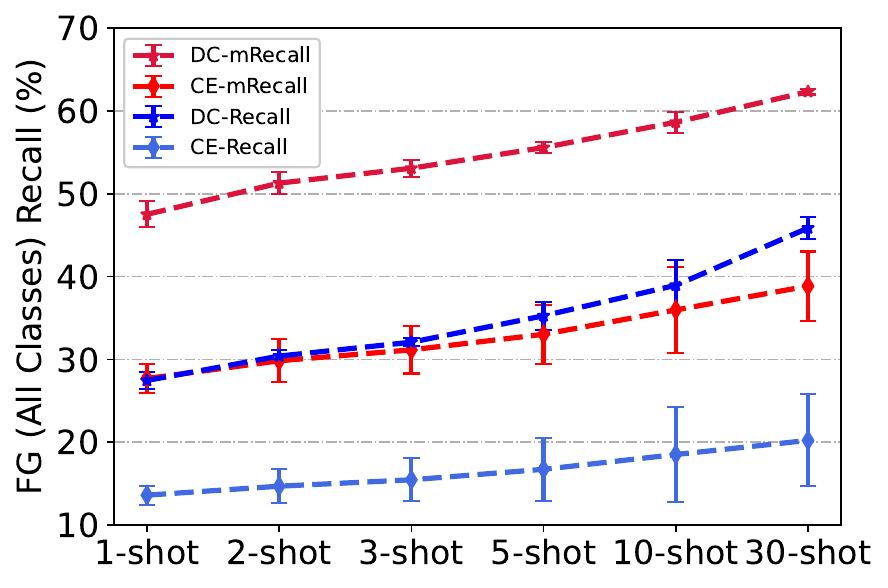}
  }
  \vspace{-8pt}
  \caption{\small{Comparison on $\rm{mRecall}$ and $\rm{Recall}$ of the proposed decoupling classifier~(DC) and standard classification head (CE) under FSIS and gFSIS settings. 
  The mean and standard deviation results are computed on all 10 seeds for each shot. Best viewed in color and zoom in.}} \label{fig:recall}
  \vspace{-8pt}
\end{figure}

\begin{table}
  \caption{\small{The effects of DC and PCB for gFSIS performance on MS-COCO. GFLOPs are averaged over all
5000 MS-COCO validation images.}}\label{fig:recall}
  \vspace{-2pt}
  \centering
  \resizebox{1.0\textwidth}{!}{
  \begin{tabular}{c|c|cc|cc|rr|rr|rr|rr}
    \toprule
      \multirow{4}{*}{Shots}  & \multirow{4}{*}{M-Rate}    &\multirow{4}{*}{DC}  &\multirow{4}{*}{PCB} &\multicolumn{2}{c|}{\textbf{Complexity}}  &\multicolumn{4}{c|}{\textbf{Detection}} &\multicolumn{4}{c}{\textbf{Segmentation}} \\
       \cmidrule(r){5-14}
         &  &  & &$\#$Params. &GFLOPs &\multicolumn{2}{c|}{\textbf{Base}} &\multicolumn{2}{c|}{\textbf{Novel}} &\multicolumn{2}{c|}{\textbf{Base}} &\multicolumn{2}{c}{\textbf{Novel}}\\
        \cmidrule(r){7-14}
        &  & & && &\textbf{AP} &\textbf{AP50} &\textbf{AP} &\textbf{AP50} &\textbf{AP} &\textbf{AP50} &\textbf{AP} &\textbf{AP50}\\
       \hline
     \multirow{4}{*}{1}  
       & \multirow{4}{*}{83.3\%}  &\xmark   & \xmark  &54.9M &334.54  &30.09        &44.45     &3.89     &7.43       &24.62      &41.58      &3.52     &6.88 \\
       &  &\cmark   & \xmark  &54.9M &334.54  &34.35        &52.46     &5.04     &10.03      &28.03      &48.60      &4.59     &9.12\\
       &  &\xmark   & \cmark  &99.4M &377.88   & 30.11       &44.42     &4.95     &9.55       &24.63      &41.57      &4.45     &8.81 \\
       & & \cmark   & \cmark  &99.4M &377.88  &\textbf{34.35}      &\textbf{52.46}      &\textbf{6.34}      &\textbf{12.79}           &\textbf{28.03}      &\textbf{48.60}      &\textbf{5.72}      &\textbf{11.53}\\
        \hline
        \multirow{4}{*}{5} 
      & \multirow{4}{*}{80.3\%}  &\xmark       & \xmark    &54.9M &334.54 &32.54    &48.03      &11.94     &21.16     &26.84      &45.04     &10.10      &19.37 \\
      & &\cmark       & \xmark   &54.9M &334.54  &35.30   &53.65    &14.01      &26.17        &29.10      &49.87      &11.80      &23.38 \\
      & &\xmark       & \cmark   &99.4M &377.88  & 32.54      & 48.03      & 13.69      & 24.41   & 26.84      & 45.04      & 11.60      & 22.36  \\
      & &\cmark       & \cmark    &99.4M&377.88 &\textbf{35.30}      &\textbf{53.65}      &\textbf{16.02}      &\textbf{30.05}       &\textbf{29.10}      &\textbf{49.87}      &\textbf{13.50}      &\textbf{26.86} \\
        \hline
        \multirow{4}{*}{10}
      &\multirow{4}{*}{76.7\%}  &\xmark       & \xmark    &54.9M&334.54 &34.05     &50.21    &14.96     &25.70      &28.12     &47.10    &12.60     &23.81  \\
      & &\cmark       & \xmark   &54.9M&334.54  &36.13    &54.81    &16.66    &30.79       &29.90    &51.07    &13.98    &27.72 \\
      & &\xmark       & \cmark   &99.4M&377.88   &34.17    &50.48    &17.02    &29.58       &28.23    &47.33    &14.32    &27.29  \\
      & &\cmark       & \cmark    &99.4M&377.88   &\textbf{36.14}      &\textbf{54.85}      &\textbf{18.67}      &\textbf{34.55}          &\textbf{29.91}      &\textbf{51.11}      &\textbf{15.71}      &\textbf{31.19}  \\

 \bottomrule
  \end{tabular}}
  \vspace{-15pt}
\end{table} 

\textbf{Effectiveness.}
We only simply replace the standard classifier with the proposed decoupling classifier in DeFRCN, which results in significant improvements, 
especially for the higher missing rate~(e.g., low shot on MS-COCO). What's more, our decoupling classifier is effective not only on novel 
classes but also on base classes, while the PCB seems only effective on novel classes. In addition, our decoupling classifier without the PCB 
significantly outperforms the counterpart with the PCB on base classes and is also comparable on novel classes.

\textbf{Efficiency.} 
Firstly, our decoupling classifier does not introduce any additional parameters or computation cost.
Secondly, our method obtains better detection and segmentation performance when using the same complexity as Mask-DeFRCN whether the PCB 
is used or not.
Last but not least, we only need almost half of the parameters and fewer GFLOPs when removing the PCB block, and still achieve significant 
improvements on base classes and comparable performance on novel classes compared to Mask-DeFRCN using the PCB.

\textbf{Why DC works?}
We have given some analysis from the perspective of gradient optimization in Sec.~\ref{sec:method}. Here, we try to discuss 
from the generalization ability of decoupling classifier and compare with the baseline.
We want to explore whether the decoupling classifier mitigates the bias classification. 
To this end, we employ $\rm{Recall}$ metric to evaluate the classification head for all ground-truth foreground objects. 
Note that the classification head outputs a multi-class probability distribution $\vec {\hat p} \in R^{C+1}$. 
The predicted class is determined by ${\rm{argmax}}_i {\hat{p}}_i$.
We define that an object is recalled if its prediction is not background, i.e., any foreground category. Considering that
the number of each foreground category varies considerably, we also compare  $\rm{mRecall}$ (mean ${\rm{Recall}}$s of all classes).
The comparison results are shown in Fig.~\ref{fig:recall}. We can see that the $\rm{mRecall}$ and $\rm{Recall}$ of 
the decoupling classifier significantly outperforms the standard one on each shot both FSIS and gFSIS. 
This indicates that our decoupling classifier is helpful to mitigate the bias classification thus 
boosting the performance of FSIS and gFSIS .

\begin{figure}[t]
 \captionsetup[subfigure]{labelformat=empty}
  \centering
  \subfloat[]   {\rotatebox{90}{\quad\quad\textbf{\scriptsize{Ours}}   \quad\quad\quad\textbf{\scriptsize{Mask-DeFRCN}}} \hspace{-3pt}}
  \subfloat[]
  {
  \includegraphics[height=0.2925\columnwidth]{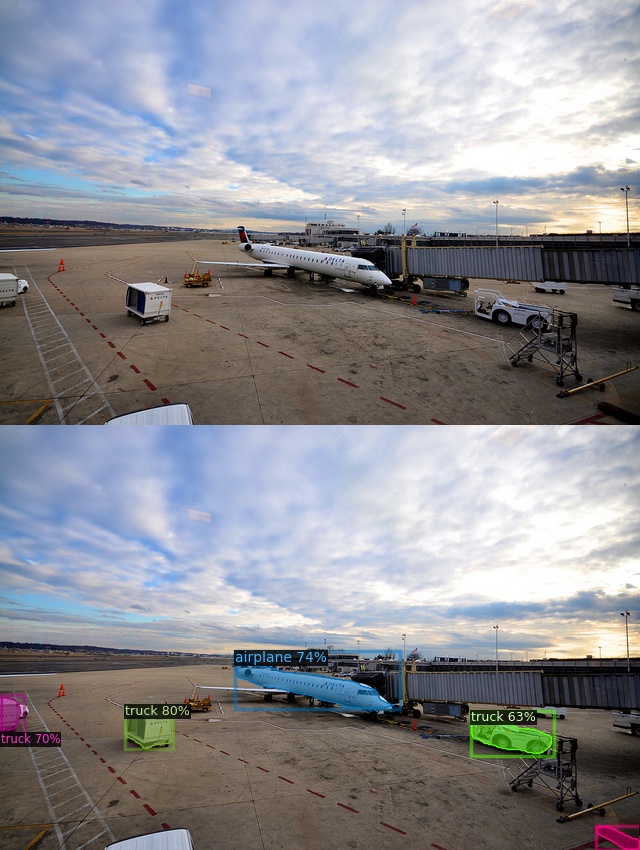}
  \hspace{-6pt}
  }
  \subfloat[]
  {
  \includegraphics[height=0.2925\columnwidth]{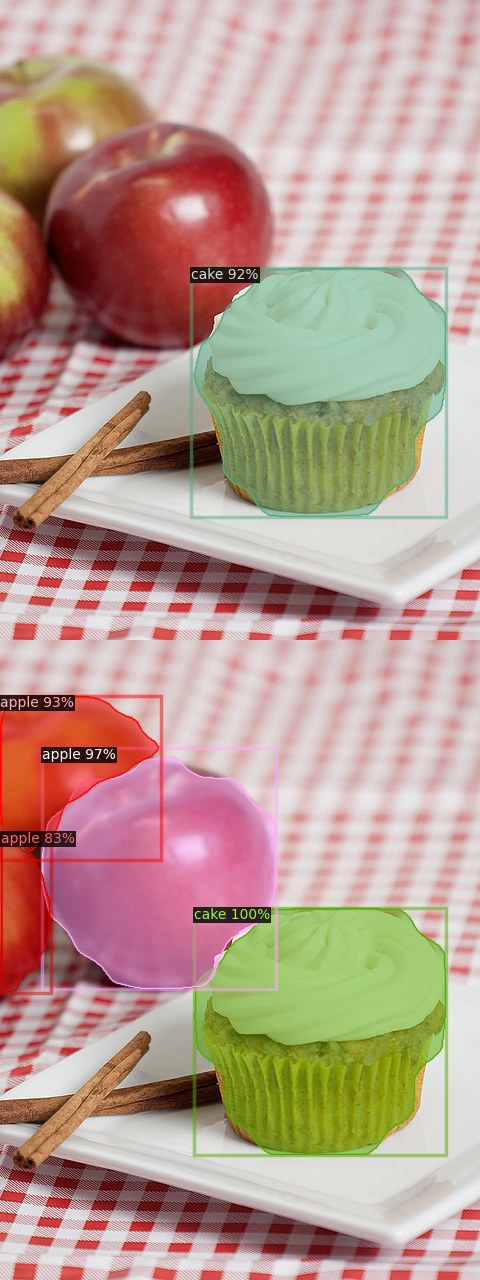}
  \hspace{-6pt}
  }  
  \subfloat[]
  {
  \includegraphics[height=0.2925\columnwidth]{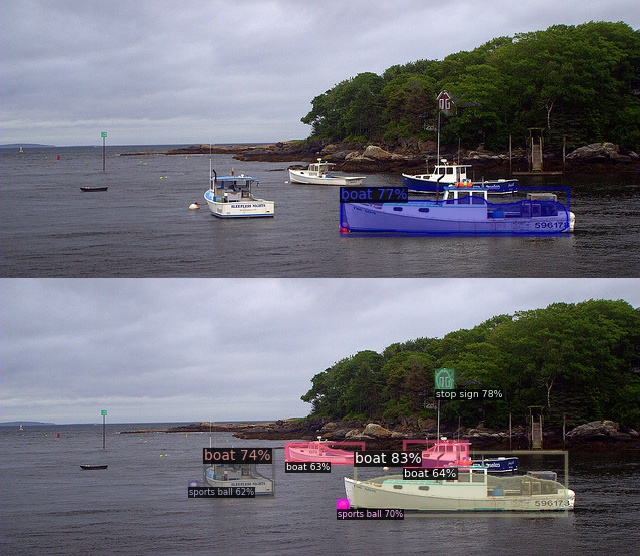}
  \hspace{-6pt}
  }
  \subfloat[]
  {
  \includegraphics[height= 0.2925\columnwidth]{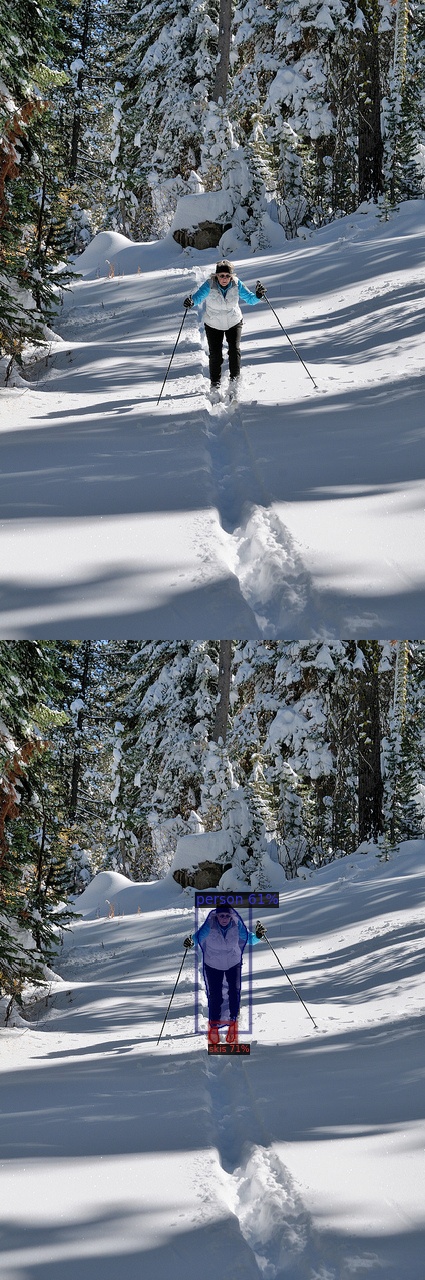}
  \hspace{-6pt}
  }
  \subfloat[]
  {
  \includegraphics[height= 0.2925\columnwidth]{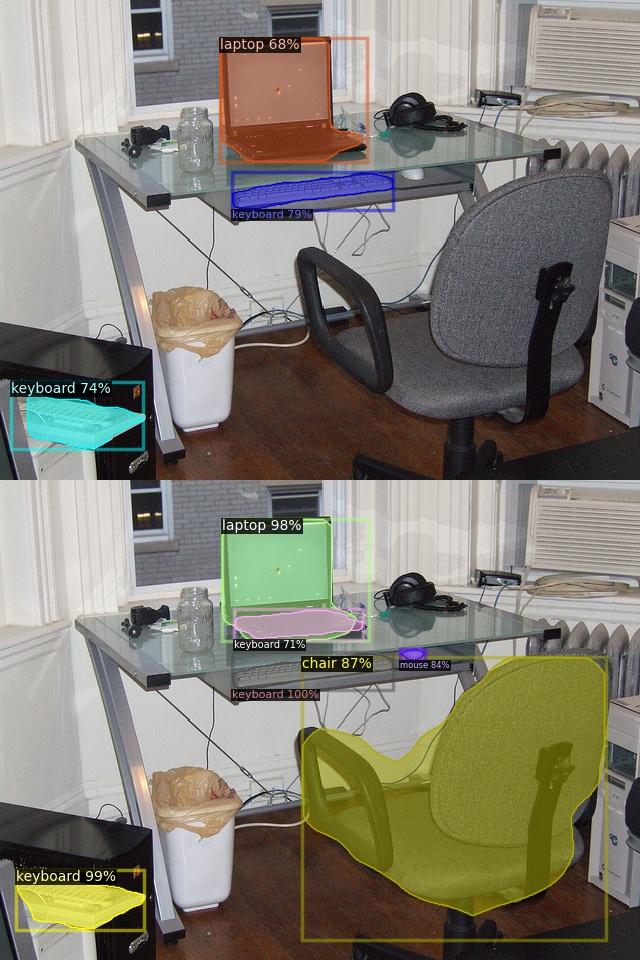}
  \hspace{-6pt}
  }
  \vspace{-12pt} \textcolor{red}{\hrule height 1.0pt width 1.005\textwidth }
  \vspace{-8pt}
  \subfloat[]   {\rotatebox{90}{\quad\textbf{\scriptsize{Ours}}   \quad\quad\textbf{\scriptsize{Mask-DeFRCN}}} \hspace{-3pt}}
  \subfloat[]
  {
  \includegraphics[height= 0.2485\columnwidth]{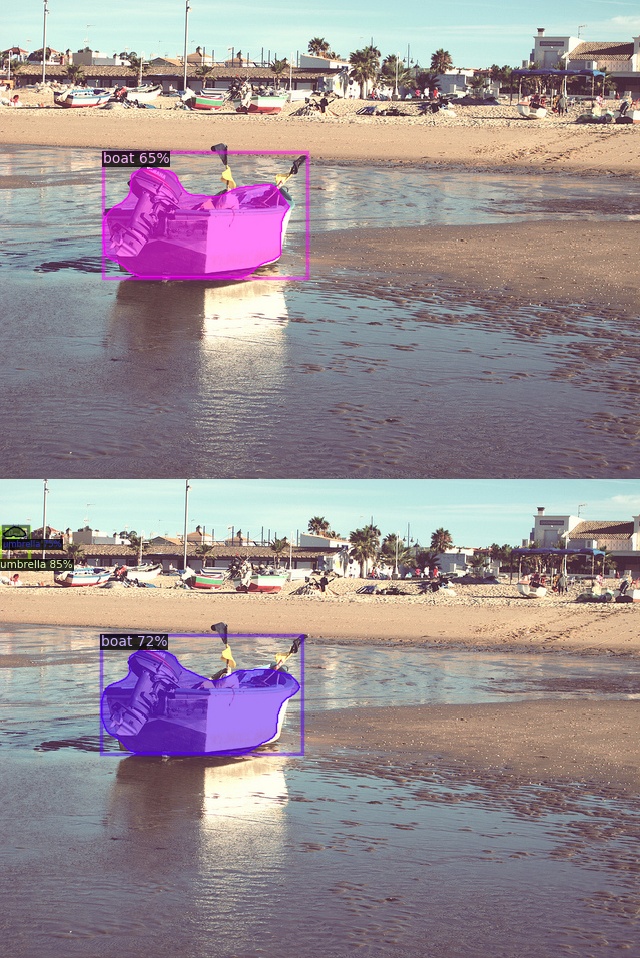}
   \hspace{-6pt}
  }
   \subfloat[]
  {
  \includegraphics[height=0.2485\columnwidth]{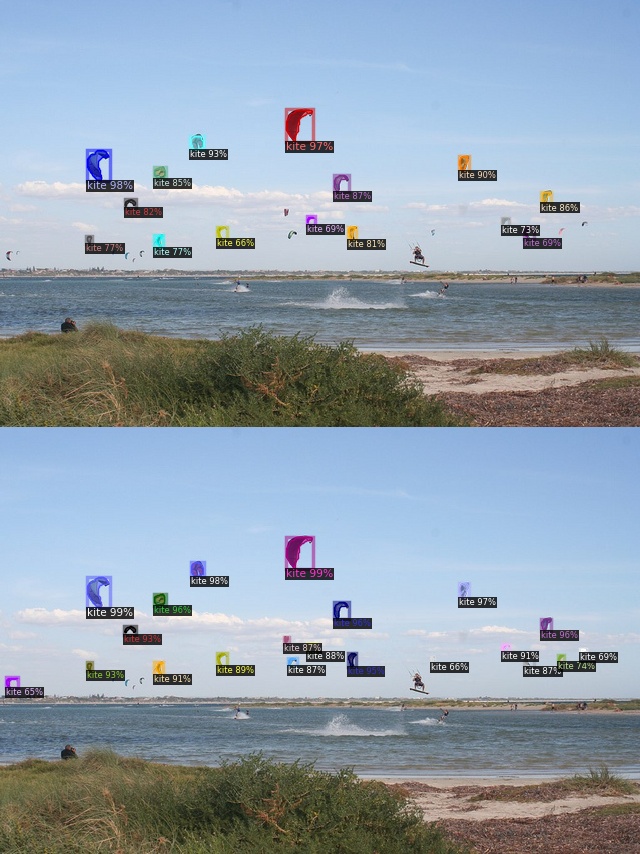}
   \hspace{-6pt}
  }
  \subfloat[]
  {
  \includegraphics[height=0.2485\columnwidth]{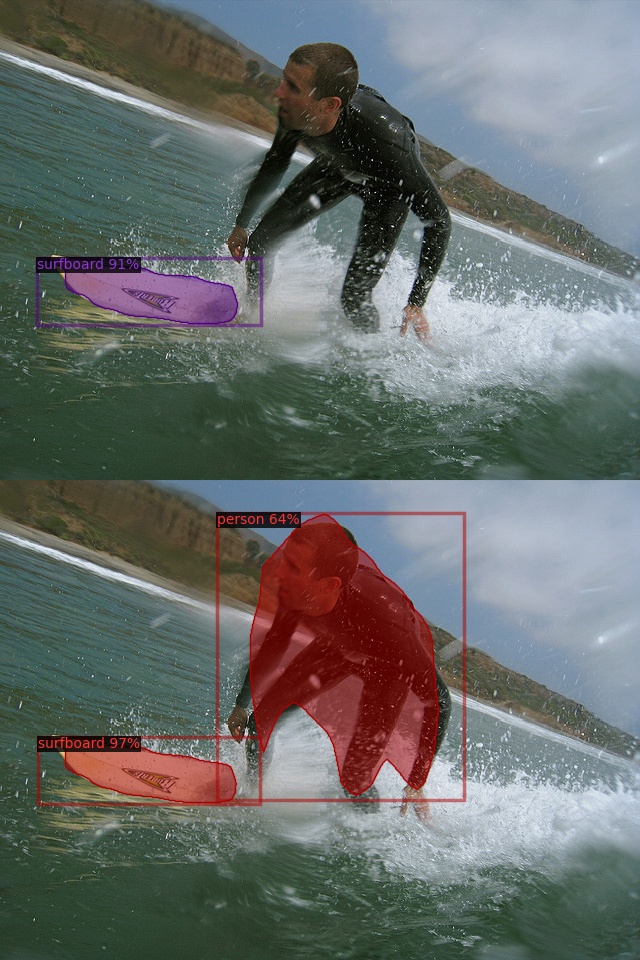}
   \hspace{-6pt}
  }
 \subfloat[]
  {
  \includegraphics[height=0.2485\columnwidth]{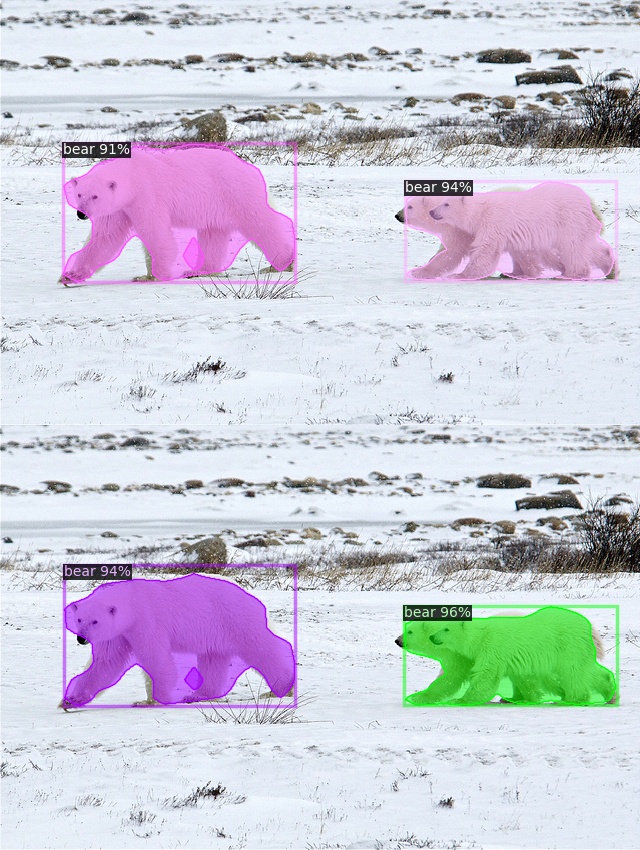}
   \hspace{-6pt}
  }
  \subfloat[]
  {
  \includegraphics[height= 0.2485\columnwidth]{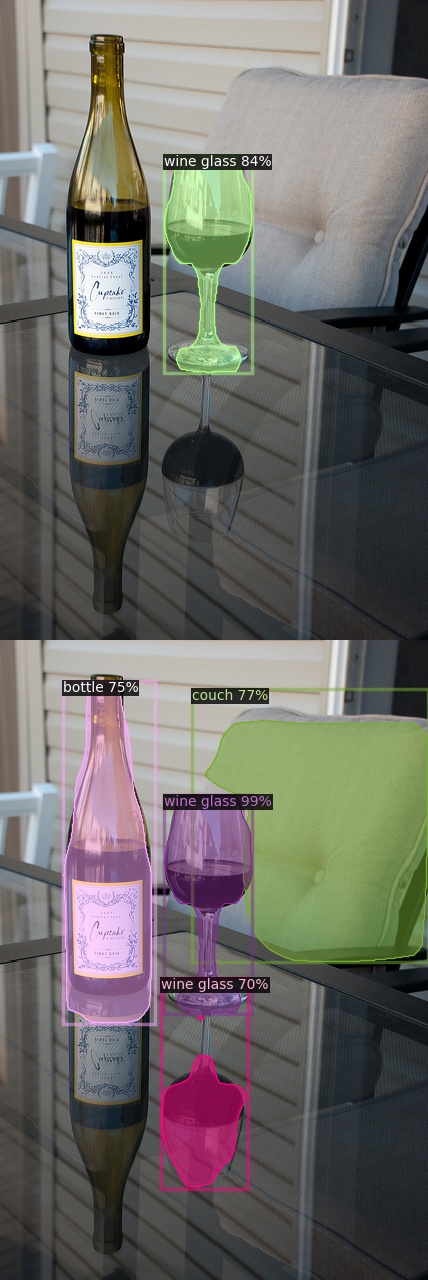}
   \hspace{-6pt}
  }
 \subfloat[]
  {
  \includegraphics[height=0.2485\columnwidth]{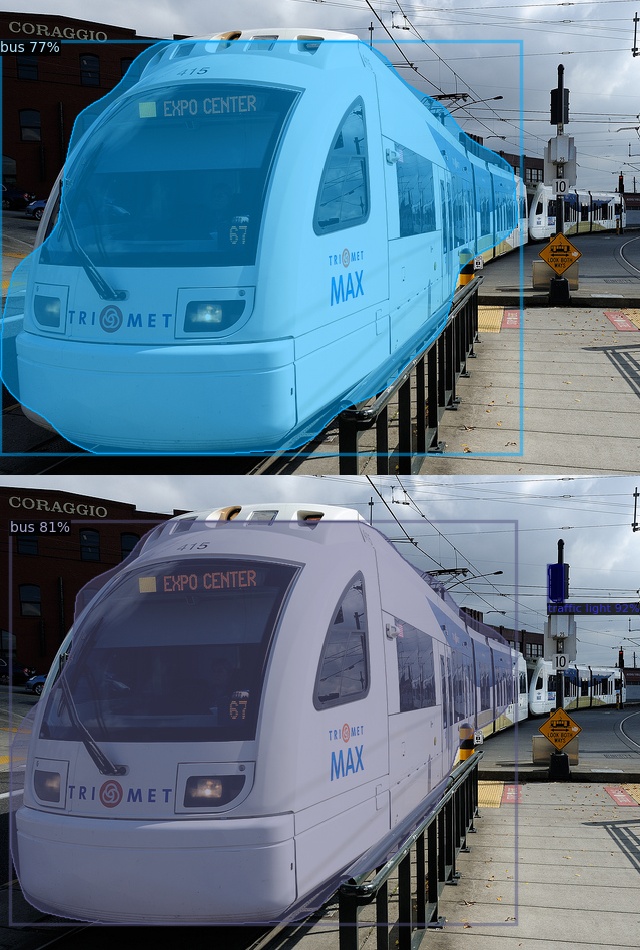}
   \hspace{-6pt}
  }  
  \vspace{-18pt}
  \caption{\small{Visualization results of our method and the strong baseline~(Mask-DeFRCN) on MS-COCO validation images. Best viewed in color and zoom in.}} \label{fig:vis-res}
  \vspace{-18pt}
 \end{figure}

\textbf{Qualitative Evaluation}
In Fig.~\ref{fig:vis-res}, we visualize the results of our method and the strong baseline (Mask-DeFRCN) on MS-COCO validation images with 10-shot setting for gFSIS task. In the top rows, we show success cases with our method but partly failures with the baseline. These failures are mainly caused by the missing detection because the baseline method may tend to incorrectly recognize positive objects as background (i.e., bias classiﬁcation). In addition, our method may also produce some failure predictions as shown in the bottom row from left to right, including the missing detection of small or occultation objects, coarse boundary segmentation, and the misclassiﬁcation of similar appearance objects.

\vspace{-5pt}
\section{Conclusion}\label{sec:con}
In this paper, we firstly find that the missing label widely exists in few-shot scenario. Furthermore, we analyze that the 
missing label issue may result in biased classification and thus limit the generalization ability on novel classes. Therefore, we 
propose a simple but effective method that decouples the standard classifier into two parallel heads to independently process positive 
and negative examples. Comprehensive experiments on the few-shot object detection and instance segmentation benchmark datasets show that our approach 
can effectively and efficiently boost FSOD/gFSOD and FSIS/gFSIS performance without any additional parameters and computation cost. 
We hope this study attract more interest in designing a simple method for FSOD or FSIS in the future.
A limitation of our method is that it may not be suitable when the missing label rate is small. However, our method is still comparable 
to its counterpart even if the missing label rate is zero, which indicates its robustness.

\clearpage
\bibliographystyle{plain}
\bibliography{bib}

\begin{thebibliography}{10}

\bibitem{cao2021fadi}
Yuhang Cao, Jiaqi Wang, Ying Jin, Tong Wu, Kai Chen, Ziwei Liu, and Dahua Lin.
\newblock Few-shot object detection via association and discrimination.
\newblock In {\em NeurIPS}, 2021.

\bibitem{chen2017deeplab}
Liang-Chieh Chen, George Papandreou, Iasonas Kokkinos, Kevin Murphy, and Alan~L
  Yuille.
\newblock Deeplab: Semantic image segmentation with deep convolutional nets,
  atrous convolution, and fully connected crfs.
\newblock {\em TPAMI}, 40(4):834--848, 2017.

\bibitem{dai2017deformable}
Jifeng Dai, Haozhi Qi, Yuwen Xiong, Yi~Li, Guodong Zhang, Han Hu, and Yichen
  Wei.
\newblock Deformable convolutional networks.
\newblock In {\em ICCV}, pages 764--773, 2017.

\bibitem{everingham2010pascal}
Mark Everingham, Luc Van~Gool, Christopher~KI Williams, John Winn, and Andrew
  Zisserman.
\newblock The pascal visual object classes (voc) challenge.
\newblock {\em IJCV}, 88(2):303--338, 2010.

\bibitem{fan2020few}
Qi~Fan, Wei Zhuo, Chi-Keung Tang, and Yu-Wing Tai.
\newblock Few-shot object detection with attention-rpn and multi-relation
  detector.
\newblock In {\em CVPR}, pages 4013--4022, 2020.

\bibitem{fan2021generalized}
Zhibo Fan, Yuchen Ma, Zeming Li, and Jian Sun.
\newblock Generalized few-shot object detection without forgetting.
\newblock In {\em CVPR}, pages 4527--4536, 2021.

\bibitem{fan2020fgn}
Zhibo Fan, Jin-Gang Yu, Zhihao Liang, Jiarong Ou, Changxin Gao, Gui-Song Xia,
  and Yuanqing Li.
\newblock Fgn: Fully guided network for few-shot instance segmentation.
\newblock In {\em CVPR}, pages 9172--9181, 2020.

\bibitem{fink2004object}
Michael Fink.
\newblock Object classification from a single example utilizing class relevance
  metrics.
\newblock In {\em NIPS}, 2004.

\bibitem{finn2017model}
Chelsea Finn, Pieter Abbeel, and Sergey Levine.
\newblock Model-agnostic meta-learning for fast adaptation of deep networks.
\newblock In {\em ICML}, pages 1126--1135, 2017.

\bibitem{ganea2021incremental}
Dan~Andrei Ganea, Bas Boom, and Ronald Poppe.
\newblock Incremental few-shot instance segmentation.
\newblock In {\em CVPR}, pages 1185--1194, 2021.

\bibitem{gidaris2018dynamic}
Spyros Gidaris and Nikos Komodakis.
\newblock Dynamic few-shot visual learning without forgetting.
\newblock In {\em CVPR}, pages 4367--4375, 2018.

\bibitem{girshick2015fast}
Ross Girshick.
\newblock {Fast R-CNN}.
\newblock In {\em ICCV}, pages 1440--1448, 2015.

\bibitem{han2022ct}
Guangxing Han, Jiawei Ma, Shiyuan Huang, Long Chen, and Shih-Fu Chang.
\newblock Few-shot object detection with fully cross-transformer.
\newblock In {\em CVPR}, pages 5321--5330, 2022.

\bibitem{he2017mask}
Kaiming He, Georgia Gkioxari, Piotr Doll{\'a}r, and Ross Girshick.
\newblock {Mask R-CNN}.
\newblock In {\em ICCV}, pages 2961--2969, 2017.

\bibitem{he2016deep}
Kaiming He, Xiangyu Zhang, Shaoqing Ren, and Jian Sun.
\newblock Deep residual learning for image recognition.
\newblock In {\em CVPR}, pages 770--778, 2016.

\bibitem{hu2021dcnet}
Hanzhe Hu, Shuai Bai, Aoxue Li, Jinshi Cui, and Liwei Wang.
\newblock Dense relation distillation with context-aware aggregation for
  few-shot object detection.
\newblock In {\em CVPR}, pages 10185--10194, 2021.

\bibitem{kang2019frsw}
Bingyi Kang, Zhuang Liu, Xin Wang, Fisher Yu, Jiashi Feng, and Trevor Darrell.
\newblock Few-shot object detection via feature reweighting.
\newblock In {\em ICCV}, pages 8420--8429, 2019.

\bibitem{kaul2022label}
Prannay Kaul, Weidi Xie, and Andrew Zisserman.
\newblock Label, verify, correct: A simple few shot object detection method.
\newblock In {\em CVPR}, pages 14237--14247, 2022.

\bibitem{li2021tip}
Aoxue Li and Zhenguo Li.
\newblock Transformation invariant few-shot object detection.
\newblock In {\em CVPR}, pages 3094--3102, 2021.

\bibitem{li2021cme}
Bohao Li, Boyu Yang, Chang Liu, Feng Liu, Rongrong Ji, and Qixiang Ye.
\newblock Beyond max-margin: Class margin equilibrium for few-shot object
  detection.
\newblock In {\em CVPR}, pages 7363--7372, 2021.

\bibitem{lin2014microsoft}
Tsung-Yi Lin, Michael Maire, Serge Belongie, James Hays, Pietro Perona, Deva
  Ramanan, Piotr Doll{\'a}r, and C~Lawrence Zitnick.
\newblock Microsoft coco: Common objects in context.
\newblock In {\em ECCV}, pages 740--755, 2014.

\bibitem{long2015fully}
Jonathan Long, Evan Shelhamer, and Trevor Darrell.
\newblock Fully convolutional networks for semantic segmentation.
\newblock In {\em CVPR}, pages 3431--3440, 2015.

\bibitem{michaelis2018one}
Claudio Michaelis, Ivan Ustyuzhaninov, Matthias Bethge, and Alexander~S Ecker.
\newblock One-shot instance segmentation.
\newblock {\em arXiv preprint arXiv:1811.11507}, 2018.

\bibitem{nguyen2021fapis}
Khoi Nguyen and Sinisa Todorovic.
\newblock Fapis: A few-shot anchor-free part-based instance segmenter.
\newblock In {\em CVPR}, pages 11099--11108, 2021.

\bibitem{nguyen2020incomplete}
Tam Nguyen and Raviv Raich.
\newblock Incomplete label multiple instance multiple label learning.
\newblock {\em TPAMI}, 2020.

\bibitem{niitani2019sampling}
Yusuke Niitani, Takuya Akiba, Tommi Kerola, Toru Ogawa, Shotaro Sano, and Shuji
  Suzuki.
\newblock Sampling techniques for large-scale object detection from sparsely
  annotated objects.
\newblock In {\em CVPR}, pages 6510--6518, 2019.

\bibitem{perez2020incremental}
Juan-Manuel Perez-Rua, Xiatian Zhu, Timothy~M Hospedales, and Tao Xiang.
\newblock Incremental few-shot object detection.
\newblock In {\em CVPR}, pages 13846--13855, 2020.

\bibitem{qiao2021defrcn}
Limeng Qiao, Yuxuan Zhao, Zhiyuan Li, Xi~Qiu, Jianan Wu, and Chi Zhang.
\newblock {DeFRCN}: Decoupled faster r-cnn for few-shot object detection.
\newblock In {\em ICCV}, pages 8681--8690, 2021.

\bibitem{redmon2017yolo9000}
Joseph Redmon and Ali Farhadi.
\newblock Yolo9000: better, faster, stronger.
\newblock In {\em CVPR}, pages 7263--7271, 2017.

\bibitem{ren2015faster}
Shaoqing Ren, Kaiming He, Ross Girshick, and Jian Sun.
\newblock {Faster R-CNN}: Towards real-time object detection with region
  proposal networks.
\newblock In {\em NIPS}, pages 91--99, 2015.

\bibitem{sun2021fsce}
Bo~Sun, Banghuai Li, Shengcai Cai, Ye~Yuan, and Chi Zhang.
\newblock {FSCE}: Few-shot object detection via contrastive proposal encoding.
\newblock In {\em CVPR}, pages 7352--7362, 2021.

\bibitem{vinyals2016matching}
Oriol Vinyals, Charles Blundell, Timothy Lillicrap, Daan Wierstra, et~al.
\newblock Matching networks for one shot learning.
\newblock In {\em NIPS}, 2016.

\bibitem{wang2020tfa}
Xin Wang, Thomas~E Huang, Trevor Darrell, Joseph~E Gonzalez, and Fisher Yu.
\newblock Frustratingly simple few-shot object detection.
\newblock In {\em ICML}, pages 9919--9928, 2020.

\bibitem{wang2019metalearn}
Yu-Xiong Wang, Deva Ramanan, and Martial Hebert.
\newblock Meta-learning to detect rare objects.
\newblock In {\em ICCV}, pages 9925--9934, 2019.

\bibitem{wu2020multi}
Jiaxi Wu, Songtao Liu, Di~Huang, and Yunhong Wang.
\newblock Multi-scale positive sample refinement for few-shot object detection.
\newblock In {\em ECCV}, pages 456--472, 2020.

\bibitem{wu2019detectron2}
Yuxin Wu, Alexander Kirillov, Francisco Massa, Wan-Yen Lo, and Ross Girshick.
\newblock Detectron2.
\newblock \url{https://github.com/facebookresearch/detectron2}, 2019.

\bibitem{wu2019soft}
Zhe Wu, Navaneeth Bodla, Bharat Singh, Mahyar Najibi, Rama Chellappa, and
  Larry~S Davis.
\newblock Soft sampling for robust object detection.
\newblock In {\em BMVC}, 2019.

\bibitem{Xiao2020FSDetView}
Yang Xiao and Renaud Marlet.
\newblock Few-shot object detection and viewpoint estimation for objects in the
  wild.
\newblock In {\em ECCV}, pages 192--210, 2020.

\bibitem{yan2019metarcnn}
Xiaopeng Yan, Ziliang Chen, Anni Xu, Xiaoxi Wang, Xiaodan Liang, and Liang Lin.
\newblock {Meta R-CNN}: Towards general solver for instance-level low-shot
  learning.
\newblock In {\em ICCV}, pages 9577--9586, 2019.

\bibitem{zareian2021open}
Alireza Zareian, Kevin~Dela Rosa, Derek~Hao Hu, and Shih-Fu Chang.
\newblock Open-vocabulary object detection using captions.
\newblock In {\em CVPR}, pages 14393--14402, 2021.

\bibitem{Zhang2020SolvingMO}
Han Zhang, Fangyi Chen, Zhiqiang Shen, Qiqi Hao, Chenchen Zhu, and Marios
  Savvides.
\newblock Solving missing-annotation object detection with background
  recalibration loss.
\newblock In {\em ICASSP}, pages 1888--1892, 2020.

\bibitem{zhang2020partial}
Min-Ling Zhang and Jun-Peng Fang.
\newblock Partial multi-label learning via credible label elicitation.
\newblock {\em TPAMI}, 43(10):3587--3599, 2020.

\bibitem{zhu2021semantic}
Chenchen Zhu, Fangyi Chen, Uzair Ahmed, Zhiqiang Shen, and Marios Savvides.
\newblock Semantic relation reasoning for shot-stable few-shot object
  detection.
\newblock In {\em CVPR}, pages 8782--8791, 2021.

\end{thebibliography}

\medskip

\clearpage

\clearpage
\appendix

\large{\textbf{Appendix: Supplementary Material}}

In this supplementary material, we first give the training details about the base pre-training and novel fine-tuning of our method in Sec.~\ref{sec:td}. Then, we provide the PyTorch-like style codes for our decoupling classifier in Sec.~\ref{sec:code}.
Next, we provide complete results including average and standard deviation of multiple runs on PASCAL VOC and MS-COCO for FSOD/FSIS and gFSOD/gFSIS in Sec.~\ref{sec:cr}. Furthermore, more visualized results of our method and the strong baseline~(Mask-DeFRCN) on MS-COCO validation images are showed in Sec.~\ref{sec:qe}. Finally, we include the missing rates on few-shot PASCAL VOC and MS-COCO in Sec.~\ref{sec:smli}.

\section{Training Details}\label{sec:td}
Following the two-stage training procedure of TFA~\cite{wang2020tfa} and DeFRCN~\cite{qiao2021defrcn}, we first pre-train model with abundant labeled images for base classes and then fine-tune the model with few-shot labeled images for novel classes or base-novel classes. For the first stage training, we employ the standard classifier~(i.e., cross entropy loss)  because all base class objects are completely labeled. In the second stage, we only simply replace the standard classifier with the proposed decoupling classifier for mitigating the bias classification under few-shot setting. For a fair comparison, we use the same hyper-parameters in the DeFRCN~\cite{qiao2021defrcn}, such as batch size, learning rate, and training iterations.

\section{The Core Code for Decoupling Classifier}\label{sec:code}
\begin{figure}[h]
  \centering
    \includegraphics[width= 0.95\columnwidth]{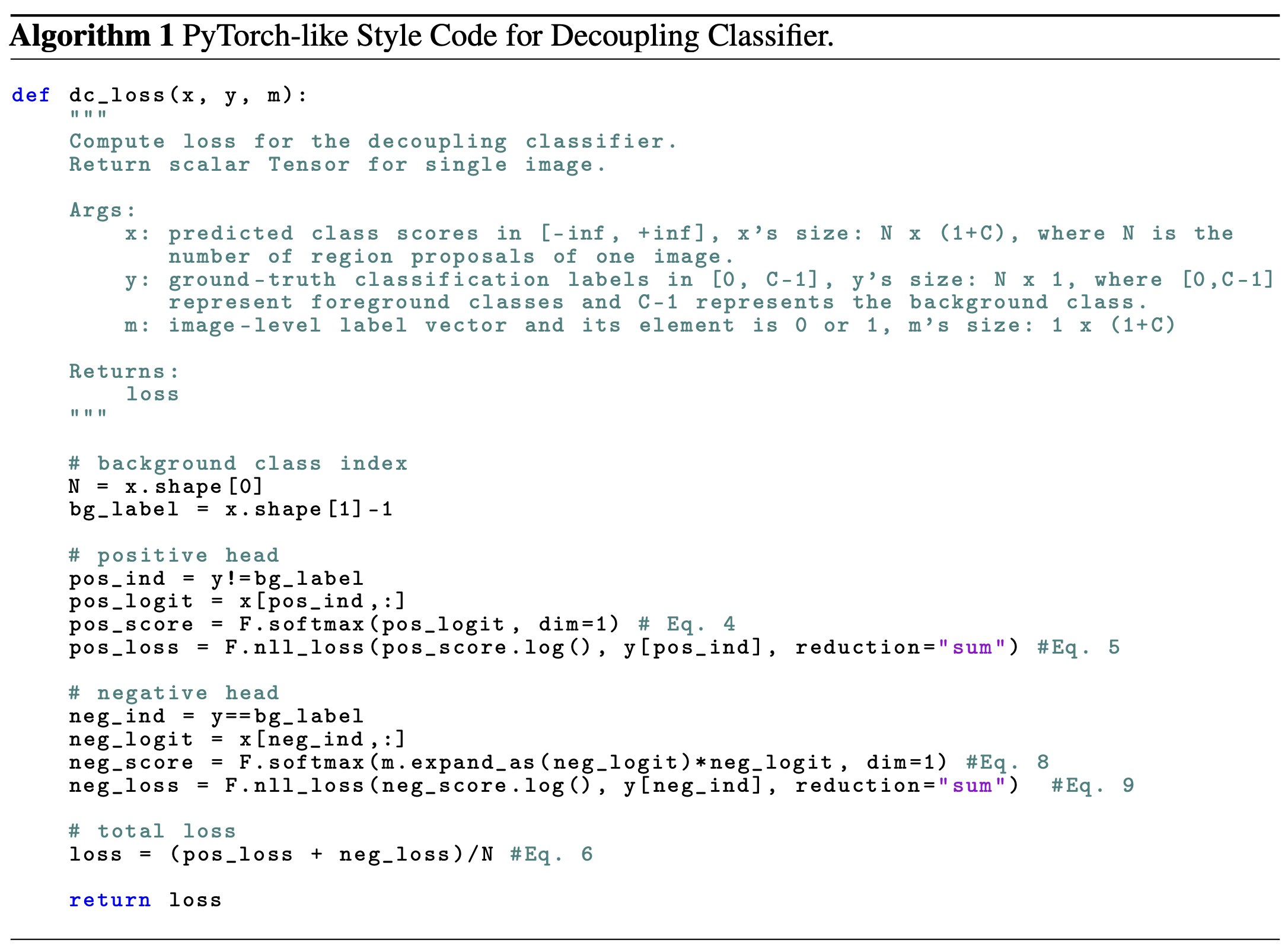}
   \caption*{\small{}}\label{alg:code}
 \end{figure}

Algorithm~1 provides the PyTorch-like style code for our decoupling classifier. It can be seen that it is very simple (core implementation only uses one line of code, the main change is to only introduce an image-level label vector, $\vec m$ in Eq.~\ref{eq:bgsm}, into the standard softmax function for the negative head and keep others unchanged like the positive head) but really effective (e.g., 5.6 AP50 improvements for detection and 4.5 AP50 improvements for segmentation on challenging MS-COCO with 5-shot setting in Table~\ref{tab:gfsis-coco-full}).

\section{Complete Results of FSOD/FSIS and gFSOD/gFSIS}\label{sec:cr}
In our main paper, we only report the average $\rm{AP}$/$\rm{AP50}$ metric for FSOD/FSIS and gFSOD/gFSIS on MS-COCO and PASCAL VOC  datasets. In this supplementary material, we report the average $\rm{AP}$/$\rm{AP50}$ metric with 95\% confidence interval over 10 seeds for FSOD/FSIS and gFSOD/gFSIS  in Tables 
\ref{tab:fsis-coco-full}, \ref{tab:gfsis-coco-full}, \ref{tab:fsod-coco-full}, \ref{tab:gfsod-coco-full} and \ref{tab:gfsod-voc-full}, respectively. $K$=$\{1, 2, 3, 5, 10, 30\}$ is the number of labeled instances of each class used in the fine-tuning stage.
\begin{table}[h]
  \caption{\small{FSIS performance (AP and AP50) for Novel classes on MS-COCO. Note that the superscript~$^\dagger$ indicates that the results are our re-implementation, the {\color{red}{red}} numers indicate the performance improvements of our method compared to the baseline, and the best results are in bold, the same below.}}\label{tab:fsis-coco-full}
  \centering
  \resizebox{1.0\textwidth}{!}{
  \begin{tabular}{c|c|rr|rr|rr|rr|rr|rr}
         \toprule
   \multirow{2}{*}{Methods}  &\multirow{2}{*}{Tasks}      
   &\multicolumn{2}{c|}{{1}} &\multicolumn{2}{c|}{{2}} &\multicolumn{2}{c|}{{3}} 
   &\multicolumn{2}{c|}{{5}} &\multicolumn{2}{c|}{{10}} &\multicolumn{2}{c}{{30}} \\
    \cmidrule(r){3-14}
     &     
     &\multicolumn{1}{c}{\textbf{AP}} &\multicolumn{1}{c|}{\textbf{AP50}} 
     &\multicolumn{1}{c}{\textbf{AP}} &\multicolumn{1}{c|}{\textbf{AP50}} 
     &\multicolumn{1}{c}{\textbf{AP}} &\multicolumn{1}{c|}{\textbf{AP50}} 
     &\multicolumn{1}{c}{\textbf{AP}} &\multicolumn{1}{c|}{\textbf{AP50}} 
     &\multicolumn{1}{c}{\textbf{AP}} &\multicolumn{1}{c|}{\textbf{AP50}} 
     &\multicolumn{1}{c}{\textbf{AP}} &\multicolumn{1}{c}{\textbf{AP50}} \\
     \toprule
     Mask-DeFRCN$^\dagger$ \cite{qiao2021defrcn}       & \multirow{3}{*}{Det}  
     &7.54$\pm$0.5    &14.46$\pm$0.9   
     &11.01$\pm$0.5  &20.20$\pm$0.7  
     &13.07$\pm$0.6  &23.28$\pm$1.0  
     &15.39$\pm$0.7  &27.29$\pm$1.0  
     &18.72$\pm$0.3  &32.80$\pm$0.6   
     &22.63$\pm$0.3  &38.95$\pm$0.5  \\                         
      \multirow{2}{*}{\textbf{Ours}}                         &      
     &\textbf{8.09}$\pm$0.4   &\textbf{15.85}$\pm$0.8 
     &\textbf{11.90}$\pm$0.4    &\textbf{22.39}$\pm$0.7   
     &\textbf{14.04}$\pm$0.6    &\textbf{25.74}$\pm$0.9 
     &\textbf{16.39}$\pm$0.6    &\textbf{29.96}$\pm$0.9 
     &\textbf{19.33}$\pm$0.4    &\textbf{34.78}$\pm$0.8  
     &\textbf{22.73}$\pm$0.4    &\textbf{40.24}$\pm$0.6      \\
     &   
     &\multicolumn{1}{c}{\textbf{\small{\color{red}+0.55}}}  &\multicolumn{1}{c|}{\textbf{\small{\color{red}+1.39}}}
     &\multicolumn{1}{c}{\textbf{\small{\color{red}+0.89}}} &\multicolumn{1}{c|}{\textbf{\small{\color{red}+2.19}}}    
     &\multicolumn{1}{c}{\textbf{\small{\color{red}+0.97}}} &\multicolumn{1}{c|}{\textbf{\small{\color{red}+2.46}}}    
     &\multicolumn{1}{c}{\textbf{\small{\color{red}+1.00}}} &\multicolumn{1}{c|}{\textbf{\small{\color{red}+2.67}}}    
     &\multicolumn{1}{c}{\textbf{\small{\color{red}+0.61}}} &\multicolumn{1}{c|}{\textbf{\small{\color{red}+1.98}}}    
     &\multicolumn{1}{c}{\textbf{\small{\color{red}+0.10}}} &\multicolumn{1}{c}{\textbf{\small{\color{red}+1.29}}}        \\
    \toprule
    Mask-DeFRCN$^\dagger$ \cite{qiao2021defrcn}                          & \multirow{3}{*}{Seg}                         
    &6.69$\pm$0.5   &13.24$\pm$0.8   
    &9.51$\pm$0.5 &18.58$\pm$0.7   
    &11.01$\pm$0.4 &21.27$\pm$0.9  
    &12.66$\pm$0.6 &24.58$\pm$1.0  
    &15.39$\pm$0.3 &29.71$\pm$0.6 
    &18.28$\pm$0.3 &35.20$\pm$0.5 \\
      \multirow{2}{*}{\textbf{Ours}}                        & 
     &\textbf{7.18}$\pm$0.5        &\textbf{14.33}$\pm$0.8 
     &\textbf{10.31}$\pm$0.4      &\textbf{20.43}$\pm$0.7   
     &\textbf{11.85}$\pm$0.4      &\textbf{23.24}$\pm$0.8  
     &\textbf{13.48}$\pm$0.5      &\textbf{26.67}$\pm$0.9   
     &\textbf{15.85}$\pm$0.4      &\textbf{31.33}$\pm$0.7  
     &\textbf{18.34}$\pm$0.3      &\textbf{35.99}$\pm$0.6   \\
     &    
     &\multicolumn{1}{c}{\textbf{\small{\color{red}+0.49}}} &\multicolumn{1}{c|}{\textbf{\small{\color{red}+1.09}}}  
     &\multicolumn{1}{c}{\textbf{\small{\color{red}+0.80}}} &\multicolumn{1}{c|}{\textbf{\small{\color{red}+1.85}}}    
     &\multicolumn{1}{c}{\textbf{\small{\color{red}+0.84}}} &\multicolumn{1}{c|}{\textbf{\small{\color{red}+1.97}}}   
     &\multicolumn{1}{c}{\textbf{\small{\color{red}+0.82}}} &\multicolumn{1}{c|}{\textbf{\small{\color{red}+2.09}}}    
     &\multicolumn{1}{c}{\textbf{\small{\color{red}+0.46}}} &\multicolumn{1}{c|}{\textbf{\small{\color{red}+1.62}}}     
     &\multicolumn{1}{c}{\textbf{\small{\color{red}+0.06}}} &\multicolumn{1}{c}{\textbf{\small{\color{red}+0.79}}}        \\
      \bottomrule
  \end{tabular}}
\end{table}

\begin{table}[h]
  \caption{\small{gFSIS performance (AP and AP50) for Overall, Base and Novel classes on MS-COCO.}}
  \label{tab:gfsis-coco-full}
  \centering
  \resizebox{1.0\textwidth}{!}{
  \begin{tabular}{l|c|rr|rr|rr|rr|rr|rr}
    \toprule
       \multirow{4}{*}{Shots}        &\multirow{4}{*}{Methods}        
       &\multicolumn{6}{c|}{\textbf{Object Detection}} &\multicolumn{6}{c}{\textbf{Instance Segmentation}} \\
    \cmidrule(r){3-14}
         &     
        &\multicolumn{2}{c|}{\textbf{Overall  \textit{\#80}}} &\multicolumn{2}{c|}{\textbf{Base  \textit{\#60}}} &\multicolumn{2}{c|}{\textbf{Novel  \textit{\#20}}} 
        &\multicolumn{2}{c|}{\textbf{Overall  \textit{\#80}}} &\multicolumn{2}{c|}{\textbf{Base  \textit{\#60}}} &\multicolumn{2}{c}{\textbf{Novel  \textit{\#20}}}\\
        \cmidrule(r){3-14}
        &     
        &\multicolumn{1}{c}{\textbf{AP}} &\multicolumn{1}{c|}{\textbf{AP50}} 
        &\multicolumn{1}{c}{\textbf{AP}} &\multicolumn{1}{c|}{\textbf{AP50}} 
        &\multicolumn{1}{c}{\textbf{AP}} &\multicolumn{1}{c|}{\textbf{AP50}} 
        &\multicolumn{1}{c}{\textbf{AP}} &\multicolumn{1}{c|}{\textbf{AP50}} 
        &\multicolumn{1}{c}{\textbf{AP}} &\multicolumn{1}{c|}{\textbf{AP50}} 
        &\multicolumn{1}{c}{\textbf{AP}} &\multicolumn{1}{c}{\textbf{AP50}} \\
    \midrule
    &Base-Only                && &\multicolumn{1}{c}{39.86} &\multicolumn{1}{c|}{59.25} &&    &&  &\multicolumn{1}{c}{32.58} &\multicolumn{1}{c|}{55.12}  && \\
    \midrule
    \multirow{2}{*}{1} 
    &Mask-DeFRCN$^\dagger$ \cite{qiao2021defrcn}
         &23.82$\pm$0.5       &35.70$\pm$0.7
         &30.11$\pm$0.6      &44.42$\pm$0.9      
         &4.95$\pm$0.4       &9.55$\pm$0.7      
         &19.58$\pm$0.4     &33.38$\pm$0.7      
         &24.63$\pm$0.5     &41.57$\pm$0.9      
         &4.45$\pm$0.5       &8.81$\pm$0.7\\
    & \multirow{2}{*}{{\textbf{Ours}}}      
         &{\textbf{27.35}$\pm$0.3}    
         &{\textbf{42.55}$\pm$0.3}   
         &{\textbf{34.35}$\pm$0.3}      
         &{\textbf{52.46}$\pm$0.3}     
         &{\textbf{6.34}$\pm$0.4}     
         &{\textbf{12.79}$\pm$0.9}      
         &{\textbf{22.45}$\pm$0.2}      
         &{\textbf{39.33}$\pm$0.3}     
         &{\textbf{28.03}$\pm$0.2}    
         &{\textbf{48.60}$\pm$0.3}      
         &{\textbf{5.72}$\pm$0.5}      
         &{\textbf{11.53}$\pm$0.9}\\
         &
         &\multicolumn{1}{c}{\textbf{\small{\color{red}+3.53}}}  &\multicolumn{1}{c|}{\textbf{\small{\color{red}+6.85}}} 
          &\multicolumn{1}{c}{\textbf{\small{\color{red}+4.24}}}  &\multicolumn{1}{c|}{\textbf{\small{\color{red}+8.04}}} 
         &\multicolumn{1}{c}{\textbf{\small{\color{red}+1.39}}}  &\multicolumn{1}{c|}{\textbf{\small{\color{red}+3.24}}} 
          &\multicolumn{1}{c}{\textbf{\small{\color{red}+2.87}}}  &\multicolumn{1}{c|}{\textbf{\small{\color{red}+6.45}}} 
          &\multicolumn{1}{c}{\textbf{\small{\color{red}+3.40}}}  &\multicolumn{1}{c|}{\textbf{\small{\color{red}+7.03}}} 
          &\multicolumn{1}{c}{\textbf{\small{\color{red}+1.27}}}  &\multicolumn{1}{c}{\textbf{\small{\color{red}+2.72}}} 
         \\

    \hline
     \multirow{2}{*}{2} 
    &Mask-DeFRCN$^\dagger$ \cite{qiao2021defrcn}
         &25.42$\pm$0.5     &38.31$\pm$0.8      &31.06$\pm$0.5      &45.82$\pm$0.7      &8.52$\pm$0.8       &15.79$\pm$1.1      &21.09$\pm$0.4      &35.92$\pm$0.8      &25.61$\pm$0.3      &43.03$\pm$0.7      &7.54$\pm$0.8       &14.59$\pm$1.1      \\
    &\multirow{2}{*}{\textbf{Ours}} 
         &{\textbf{28.63}}$\pm$0.3        &{\textbf{44.74}}$\pm$0.5      &{\textbf{34.67}}$\pm$0.3     &{\textbf{52.82}}$\pm$0.4     
         &{\textbf{10.52}}$\pm$0.7      &{\textbf{20.49}}$\pm$1.1      &{\textbf{23.73}}$\pm$0.3     &{\textbf{41.49}}$\pm$0.4      
         &{\textbf{28.52}}$\pm$0.2      &{\textbf{49.12}}$\pm$0.3      &{\textbf{9.38}}$\pm$0.7      &{\textbf{18.62}}$\pm$1.1  \\
          &
         &\multicolumn{1}{c}{\textbf{\small{\color{red}+3.21}}}  &\multicolumn{1}{c|}{\textbf{\small{\color{red}+6.43}}}
          &\multicolumn{1}{c}{\textbf{\small{\color{red}+3.61}}}  &\multicolumn{1}{c|}{\textbf{\small{\color{red}+7.00}}} 
         &\multicolumn{1}{c}{\textbf{\small{\color{red}+2.00}}}  &\multicolumn{1}{c|}{\textbf{\small{\color{red}+4.70}}} 
          &\multicolumn{1}{c}{\textbf{\small{\color{red}+2.64}}}  &\multicolumn{1}{c|}{\textbf{\small{\color{red}+5.57}}} 
          &\multicolumn{1}{c}{\textbf{\small{\color{red}+2.91}}}  &\multicolumn{1}{c|}{\textbf{\small{\color{red}+6.09}}} 
          &\multicolumn{1}{c}{\textbf{\small{\color{red}+1.84}}}  &\multicolumn{1}{c}{\textbf{\small{\color{red}+4.03}}} \\
         \hline
         \multirow{2}{*}{3} 
    &Mask-DeFRCN$^\dagger$ \cite{qiao2021defrcn}
         &26.54$\pm$0.5      & 40.01$\pm$0.7      & 31.77$\pm$0.4      & 46.83$\pm$0.6      & 10.87$\pm$0.8      & 19.55$\pm$1.2      & 22.04$\pm$0.4      & 37.48$\pm$0.7      & 26.22$\pm$0.3      & 43.95$\pm$0.5      & 9.48$\pm$0.7       &18.06$\pm$1.1      \\
         &\multirow{2}{*}{\textbf{Ours}} 
         &{\textbf{29.59}}$\pm$0.2     &{\textbf{46.21}}$\pm$0.4    &{\textbf{35.07}}$\pm$0.2      &{\textbf{53.30}}$\pm$0.4      
         &{\textbf{13.15}}$\pm$0.5      &{\textbf{24.95}}$\pm$0.8     &{\textbf{24.55}}$\pm$0.2      &{\textbf{42.81}}$\pm$0.3      
         &{\textbf{28.91}}$\pm$0.2      &{\textbf{49.61}}$\pm$0.4     &{\textbf{11.46}}$\pm$0.4     &{\textbf{22.43}}$\pm$0.8      \\
          &
         &\multicolumn{1}{c}{\textbf{\small{\color{red}+3.05}}}  &\multicolumn{1}{c|}{\textbf{\small{\color{red}+6.20}}} 
          &\multicolumn{1}{c}{\textbf{\small{\color{red}+3.30}}}  &\multicolumn{1}{c|}{\textbf{\small{\color{red}+6.47}}} 
         &\multicolumn{1}{c}{\textbf{\small{\color{red}+2.28}}}  &\multicolumn{1}{c|}{\textbf{\small{\color{red}+5.40}}} 
          &\multicolumn{1}{c}{\textbf{\small{\color{red}+2.51}}}  &\multicolumn{1}{c|}{\textbf{\small{\color{red}+5.33}}} 
          &\multicolumn{1}{c}{\textbf{\small{\color{red}+2.69}}}  &\multicolumn{1}{c|}{\textbf{\small{\color{red}+5.66}}} 
          &\multicolumn{1}{c}{\textbf{\small{\color{red}+1.98}}}  &\multicolumn{1}{c}{\textbf{\small{\color{red}+4.37}}} \\
\hline
    \multirow{2}{*}{5} 
       &Mask-DeFRCN$^\dagger$ \cite{qiao2021defrcn}
        &27.82$\pm$0.4      & 42.12$\pm$0.6      & 32.54$\pm$0.4      & 48.03$\pm$0.5      & 13.69$\pm$0.7      & 24.41$\pm$1.3      & 23.03$\pm$0.3      & 39.37$\pm$0.6      & 26.84$\pm$0.3      & 45.04$\pm$0.5      & 11.60$\pm$0.7      &22.36$\pm$1.2  \\
     &\multirow{2}{*}{\textbf{Ours}}      
        &{\textbf{30.48}}$\pm$0.2     &{\textbf{47.75}}$\pm$0.3      &{\textbf{35.30}}$\pm$0.2     &{\textbf{53.65}}$\pm$0.3      &{\textbf{16.02}}$\pm$0.5      &{\textbf{30.05}}$\pm$0.8      &{\textbf{25.20}}$\pm$0.2      &{\textbf{44.12}}$\pm$0.3      &{\textbf{29.10}}$\pm$0.2     &{\textbf{49.87}}$\pm$0.3      &{\textbf{13.50}}$\pm$0.5     &\textbf{26.86}$\pm$0.9\\
         &
         &\multicolumn{1}{c}{\textbf{\small{\color{red}+2.66}}}  &\multicolumn{1}{c|}{\textbf{\small{\color{red}+5.63}}} 
          &\multicolumn{1}{c}{\textbf{\small{\color{red}+2.76}}}  &\multicolumn{1}{c|}{\textbf{\small{\color{red}+5.62}}} 
         &\multicolumn{1}{c}{\textbf{\small{\color{red}+2.33}}}  &\multicolumn{1}{c|}{\textbf{\small{\color{red}+5.64}}} 
          &\multicolumn{1}{c}{\textbf{\small{\color{red}+2.17}}}  &\multicolumn{1}{c|}{\textbf{\small{\color{red}+4.75}}} 
          &\multicolumn{1}{c}{\textbf{\small{\color{red}+2.26}}}  &\multicolumn{1}{c|}{\textbf{\small{\color{red}+4.83}}} 
          &\multicolumn{1}{c}{\textbf{\small{\color{red}+1.90}}}  &\multicolumn{1}{c}{\textbf{\small{\color{red}+4.50}}} \\
    \hline
    \multirow{2}{*}{10} 
    &Mask-DeFRCN$^\dagger$ \cite{qiao2021defrcn}
         &29.88$\pm$0.3    &45.25$\pm$0.7    &34.17$\pm$0.3    &50.48$\pm$0.5    &17.02$\pm$0.6    &29.58$\pm$1.2    &24.75$\pm$0.3    &42.32$\pm$0.6    &28.23$\pm$0.2    &47.33$\pm$0.5    &14.32$\pm$0.6    &27.29$\pm$1.1    \\
         &\multirow{2}{*}{\textbf{Ours}}       
         &{\textbf{31.77}}$\pm$0.2      &{\textbf{49.77}}$\pm$0.3    &{\textbf{36.14}}$\pm$0.2      &{\textbf{54.85}}$\pm$0.2      
         &{\textbf{18.67}}$\pm$0.4      &{\textbf{34.55}}$\pm$0.7     &{\textbf{26.36}}$\pm$0.2     &{\textbf{46.13}}$\pm$0.3     
         &{\textbf{29.91}}$\pm$0.2      &{\textbf{51.11}}$\pm$0.2    &{\textbf{15.71}}$\pm$0.4      &{\textbf{31.19}}$\pm$0.7  \\
          &
         &\multicolumn{1}{c}{\textbf{\small{\color{red}+1.89}}}  &\multicolumn{1}{c|}{\textbf{\small{\color{red}+4.52}}} 
          &\multicolumn{1}{c}{\textbf{\small{\color{red}+1.97}}}  &\multicolumn{1}{c|}{\textbf{\small{\color{red}+4.37}}} 
         &\multicolumn{1}{c}{\textbf{\small{\color{red}+1.65}}}  &\multicolumn{1}{c|}{\textbf{\small{\color{red}+4.97}}} 
          &\multicolumn{1}{c}{\textbf{\small{\color{red}+1.61}}}  &\multicolumn{1}{c|}{\textbf{\small{\color{red}+3.81}}} 
          &\multicolumn{1}{c}{\textbf{\small{\color{red}+1.68}}}  &\multicolumn{1}{c|}{\textbf{\small{\color{red}+3.78}}} 
          &\multicolumn{1}{c}{\textbf{\small{\color{red}+1.39}}}  &\multicolumn{1}{c}{\textbf{\small{\color{red}+3.90}}} \\
         \hline
         \multirow{2}{*}{30}
    &Mask-DeFRCN$^\dagger$ \cite{qiao2021defrcn}
         &31.66$\pm$0.1      &48.11$\pm$0.2      &35.10$\pm$0.1     &52.01$\pm$0.2      &21.33$\pm$0.4      &36.44$\pm$0.7      &26.23$\pm$0.1      &44.97$\pm$0.2      &29.12$\pm$0.1     &48.82$\pm$0.1     &17.57$\pm$0.4      &33.42$\pm$0.7      \\
    &\multirow{2}{*}{\textbf{Ours}}
         &{\textbf{32.92}}$\pm$0.2     &{\textbf{51.37}}$\pm$0.4      &{\textbf{36.45}}$\pm$0.3      &{\textbf{55.05}}$\pm$0.4      
         &{\textbf{22.30}}$\pm$0.4     &{\textbf{40.31}}$\pm$0.6      &{\textbf{27.31}}$\pm$0.2      &{\textbf{47.61}}$\pm$0.4    
         &{\textbf{30.32}}$\pm$0.2     &{\textbf{51.41}}$\pm$0.4      &{\textbf{18.29}}$\pm$0.3      &{\textbf{36.22}}$\pm$0.6     \\  
          &
         &\multicolumn{1}{c}{\textbf{\small{\color{red}+1.26}}}  &\multicolumn{1}{c|}{\textbf{\small{\color{red}+3.26}}} 
          &\multicolumn{1}{c}{\textbf{\small{\color{red}+1.35}}}  &\multicolumn{1}{c|}{\textbf{\small{\color{red}+3.04}}} 
         &\multicolumn{1}{c}{\textbf{\small{\color{red}+0.97}}}  &\multicolumn{1}{c|}{\textbf{\small{\color{red}+3.87}}} 
          &\multicolumn{1}{c}{\textbf{\small{\color{red}+1.08}}}  &\multicolumn{1}{c|}{\textbf{\small{\color{red}+2.64}}} 
          &\multicolumn{1}{c}{\textbf{\small{\color{red}+1.20}}}  &\multicolumn{1}{c|}{\textbf{\small{\color{red}+2.59}}} 
          &\multicolumn{1}{c}{\textbf{\small{\color{red}+0.72}}}  &\multicolumn{1}{c}{\textbf{\small{\color{red}+2.80}}} \\
    \bottomrule
  \end{tabular}}
\end{table}

\begin{table}[htb]
    \centering
    \caption{FSOD performance ($\rm{AP}$ and $\rm{AP50}$) for Novel classes on MS-COCO.}\label{tab:fsod-coco-full}
    \resizebox{1.0\textwidth}{!}{
    \begin{tabular}{cc|rr|rr|rr|rr|rr|rr}
        \toprule[1.1pt]
        \multirow{2}{*}{Methods / Shots}  & 
        &\multicolumn{2}{c|}{{1}} 
        &\multicolumn{2}{c|}{{2}} 
        &\multicolumn{2}{c|}{{3}} 
        &\multicolumn{2}{c|}{{5}} 
        &\multicolumn{2}{c|}{{10}} 
        &\multicolumn{2}{c}{{30}} \\     
                \cmidrule{3-14}
        &
        &\multicolumn{1}{c}{\textbf{AP}} &\multicolumn{1}{c|}{\textbf{AP50}} 
        &\multicolumn{1}{c}{\textbf{AP}} &\multicolumn{1}{c|}{\textbf{AP50}} 
        &\multicolumn{1}{c}{\textbf{AP}} &\multicolumn{1}{c|}{\textbf{AP50}} 
        &\multicolumn{1}{c}{\textbf{AP}} &\multicolumn{1}{c|}{\textbf{AP50}} 
        &\multicolumn{1}{c}{\textbf{AP}} &\multicolumn{1}{c|}{\textbf{AP50}} 
        &\multicolumn{1}{c}{\textbf{AP}} &\multicolumn{1}{c}{\textbf{AP50}} \\
   \midrule[0.9pt]
        {DeFRCN$^\dagger$ \cite{qiao2021defrcn}} &\emph{ICCV 21}   
        &7.7$\pm$0.6   &15.1$\pm$0.9
        &11.4$\pm$0.5 &21.4$\pm$0.8
        &13.3$\pm$0.4 &24.5$\pm$0.9
        &15.5$\pm$0.5 &28.3$\pm$0.9
        &18.5$\pm$0.4 &33.4$\pm$0.6
        &22.5$\pm$0.3 &39.5$\pm$0.4\\ 
       
      \multirow{2}{*} {\textbf{Ours}}  &  
        &\textbf{8.1}$\pm$0.6  &\textbf{16.3}$\pm$0.9
        &\textbf{12.1}$\pm$0.5 &\textbf{23.4}$\pm$0.7
        &\textbf{14.4}$\pm$0.4 &\textbf{27.1}$\pm$1.0
        &\textbf{16.6}$\pm$0.5 &\textbf{31.1}$\pm$0.9
        &\textbf{19.5}$\pm$0.5 &\textbf{35.8}$\pm$0.8
        &\textbf{22.7}$\pm$0.4 &\textbf{41.0}$\pm$0.6  \\ 
        &
        &\multicolumn{1}{c}{\textbf{\small{\color{red}+0.4}}}  &\multicolumn{1}{c|}{\textbf{\small{\color{red}+1.2}}} 
          &\multicolumn{1}{c}{\textbf{\small{\color{red}+0.7}}}  &\multicolumn{1}{c|}{\textbf{\small{\color{red}+2.0}}} 
  &\multicolumn{1}{c}{\textbf{\small{\color{red}+1.1}}}  &\multicolumn{1}{c|}{\textbf{\small{\color{red}+2.6}}} 

  &\multicolumn{1}{c}{\textbf{\small{\color{red}+1.1}}}  &\multicolumn{1}{c|}{\textbf{\small{\color{red}+2.8}}} 

  &\multicolumn{1}{c}{\textbf{\small{\color{red}+1.0}}}  &\multicolumn{1}{c|}{\textbf{\small{\color{red}+2.4}}} 

  &\multicolumn{1}{c}{\textbf{\small{\color{red}+0.2}}}  &\multicolumn{1}{c}{\textbf{\small{\color{red}+0.5}}} 
\\
        \bottomrule[1.1pt]
    \end{tabular}
    }
  \end{table}

\begin{table}[htb] 
\centering
	\caption{gFSOD performance ($\rm{AP}$ and $\rm{AP50}$) for Overall, Base and Novel classes on MS-COCO.}
	\label{tab:gfsod-coco-full}
	\scalebox{0.7}{
		\begin{tabular}{c|cc|ccc|c|cc}
			\toprule			
			\multirow{2}{*}{\# shots} &\multirow{2}{*}{Methods}  &&  \multicolumn{3}{c|}{Overall \textit{\#80}}  &\multicolumn{1}{c|}{Base \textit{\#60}} & \multicolumn{2}{c}{Novel \textit{\#20}} \\ \cmidrule{4-9} 
			&&  & AP & AP50 & AP75 & AP & AP & AP50 \\ \midrule
			\multirow{3}{*}{1} & FRCN+ft \cite{yan2019metarcnn} &\emph{ICCV 19} & 16.2$\pm$0.9 & 25.8$\pm$1.2 & 17.6$\pm$1.0 & 21.0$\pm$1.2 & 1.7$\pm$0.2  &3.3 \\
			& TFA \cite{wang2020tfa} &\emph{ICML 20} & {24.4$\pm$0.6} & {39.8$\pm$0.8} & 26.1$\pm$0.8 &{31.9$\pm$0.7} & 1.9$\pm$0.4  &3.8\\
			& {DeFRCN}  \cite{qiao2021defrcn} &\emph{ICCV 21} 
			&24.0$\pm$0.4 
			&36.9$\pm$0.6 
			&26.2$\pm$0.4
		        &30.4$\pm$0.4
			&4.8$\pm$0.6  &9.5$\pm$0.9\\ 
			&\cellcolor{Gray} {\textbf{Ours}} &\cellcolor{Gray} {}
			& \cellcolor{Gray} \textbf{27.4$\pm$}0.2 \textbf{\scriptsize{\color{red}+3.4}} 
			& \cellcolor{Gray} \textbf{43.4$\pm$}0.4 \textbf{\scriptsize{\color{red}+6.5}}
			& \cellcolor{Gray} \textbf{29.4$\pm$}0.3 \textbf{\scriptsize{\color{red}+3.2}} 
			& \cellcolor{Gray} \textbf{34.4$\pm$}0.3 \textbf{\scriptsize{\color{red}+4.0}} 
			& \cellcolor{Gray} \textbf{ 6.2$\pm$}0.6 \textbf{\scriptsize{\color{red}+1.4}}
			& \cellcolor{Gray} \textbf{12.7$\pm$}0.9 \textbf{\scriptsize{\color{red}+3.2}}
\\ 			
			\midrule
			\multirow{3}{*}{2} & FRCN+ft \cite{yan2019metarcnn} &\emph{ICCV 19}& 15.8$\pm$0.7 & 25.0$\pm$1.1 & 17.3$\pm$0.7 & 20.0$\pm$0.9 & 3.1$\pm$0.3  &6.1\\
			& TFA \cite{wang2020tfa} &\emph{ICML 20}& 24.9$\pm$0.6 &{40.1$\pm$0.9} & 27.0$\pm$0.7 & {31.9$\pm$0.7}  &3.9$\pm$0.4  &7.8\\
				& {DeFRCN}  \cite{qiao2021defrcn} &\emph{ICCV 21} 
			&25.7$\pm$0.5 
			&39.6$\pm$0.8 
			&28.0$\pm$0.5
		        &31.4$\pm$0.4
			&8.5$\pm$0.8  &16.3$\pm$1.4\\
			 &\cellcolor{Gray} {\textbf{Ours}} &\cellcolor{Gray} {}
			& \cellcolor{Gray} \textbf{28.6$\pm$}0.3 \textbf{\scriptsize{\color{red}+2.9}} 
			& \cellcolor{Gray} \textbf{45.6$\pm$}0.5 \textbf{\scriptsize{\color{red}+6.0}}
			& \cellcolor{Gray} \textbf{30.7$\pm$}0.4 \textbf{\scriptsize{\color{red}+2.7}} 
			& \cellcolor{Gray} \textbf{34.7$\pm$}0.3 \textbf{\scriptsize{\color{red}+3.3}} 
			& \cellcolor{Gray} \textbf{10.4$\pm$}0.8 \textbf{\scriptsize{\color{red}+1.9}} 
			&\cellcolor{Gray} \textbf{20.9$\pm$}1.3 \textbf{\scriptsize{\color{red}+4.6}} \\ 

			\midrule
			\multirow{3}{*}{3} & FRCN+ft \cite{yan2019metarcnn} &\emph{ICCV 19}& 15.0$\pm$0.7 & 23.9$\pm$1.2 & 16.4$\pm$0.7 & 18.8$\pm$0.9 & 3.7$\pm$0.4  &7.1\\
			& TFA \cite{wang2020tfa} &\emph{ICML 20}&25.3$\pm$0.6 & 40.4$\pm$1.0 & 27.6$\pm$0.7 & 32.0$\pm$0.7 & 5.1$\pm$0.6 &9.9 \\			
				& {DeFRCN}  \cite{qiao2021defrcn} &\emph{ICCV 21} 
			&26.6$\pm$0.4
			&41.1$\pm$0.7 
			&28.9$\pm$0.4
		        &32.1$\pm$0.3
			&10.7$\pm$0.8 
			&20.0$\pm$1.2 \\
			 &\cellcolor{Gray} {\textbf{Ours}} &\cellcolor{Gray} {}
			& \cellcolor{Gray} \textbf{29.4$\pm$}0.2 \textbf{\scriptsize{\color{red}+2.8}} 
			& \cellcolor{Gray} \textbf{46.8$\pm$}0.3 \textbf{\scriptsize{\color{red}+5.7}}
			& \cellcolor{Gray} \textbf{31.4$\pm$}0.3 \textbf{\scriptsize{\color{red}+2.5}} 
			& \cellcolor{Gray} \textbf{34.9$\pm$}0.2 \textbf{\scriptsize{\color{red}+2.8}} 
			& \cellcolor{Gray} \textbf{12.9$\pm$}0.6 \textbf{\scriptsize{\color{red}+2.2}}  
			& \cellcolor{Gray} \textbf{25.1$\pm$}1.0 \textbf{\scriptsize{\color{red}+5.1}}  \\ 
			
			\midrule
			\multirow{3}{*}{5} & FRCN+ft  \cite{yan2019metarcnn}&\emph{ICCV 19} &14.4$\pm$0.8 & 23.0$\pm$1.3 & 15.6$\pm$0.8 & 17.6$\pm$0.9 & 4.6$\pm$0.5  &8.7 \\
			& TFA \cite{wang2020tfa} &\emph{ICML 20}& 25.9$\pm$0.6 & 41.2$\pm$0.9 &28.4$\pm$0.6 & 32.3$\pm$0.6 & 7.0$\pm$0.7 &13.3 \\			
			& {DeFRCN}  \cite{qiao2021defrcn} &\emph{ICCV 21} 
			&27.8$\pm$0.3
			&43.0$\pm$0.6
			&30.2$\pm$0.3
		        &32.6$\pm$0.3
			&13.6$\pm$0.7 
			&24.7$\pm$1.1 \\
			&\cellcolor{Gray} {\textbf{Ours}} &\cellcolor{Gray} {}
			& \cellcolor{Gray} \textbf{30.2$\pm$}0.2 \textbf{\scriptsize{\color{red}+2.4}} 
			& \cellcolor{Gray} \textbf{48.2$\pm$}0.3 \textbf{\scriptsize{\color{red}+5.2}}
			& \cellcolor{Gray} \textbf{32.2$\pm$}0.2 \textbf{\scriptsize{\color{red}+2.0}} 
			& \cellcolor{Gray} \textbf{35.0$\pm$}0.2 \textbf{\scriptsize{\color{red}+3.6}} 
			& \cellcolor{Gray} \textbf{15.7$\pm$}0.5 \textbf{\scriptsize{\color{red}+2.1}} 
			&\cellcolor{Gray} \textbf{30.3$\pm$}0.9 \textbf{\scriptsize{\color{red}+5.6}} \\ 

			\midrule
			\multirow{3}{*}{10} & FRCN+ft \cite{yan2019metarcnn} &\emph{ICCV 19}& 13.4$\pm$1.0 & 21.8$\pm$1.7 & 14.5$\pm$0.9 & 16.1$\pm$1.0 & 5.5$\pm$0.9 &10.0\\
			& TFA \cite{wang2020tfa}  &\emph{ICML 20}& 26.6$\pm$0.5 &42.2$\pm$0.8 & 29.0$\pm$0.6 & 32.4$\pm$0.6 & 9.1$\pm$0.5 &17.1 \\			
			& {DeFRCN}  \cite{qiao2021defrcn} &\emph{ICCV 21} 
			&29.7$\pm$0.2
			&46.0$\pm$0.5
			&32.1$\pm$0.2
		        &34.0$\pm$0.2
			&16.8$\pm$0.6 
			&29.6$\pm$1.3 \\
			&\cellcolor{Gray} {\textbf{Ours}} &\cellcolor{Gray} {}
			& \cellcolor{Gray} \textbf{31.4$\pm$}0.2 \textbf{\scriptsize{\color{red}+1.7}} 
			& \cellcolor{Gray} \textbf{49.9$\pm$}0.3 \textbf{\scriptsize{\color{red}+3.9}}
			& \cellcolor{Gray} \textbf{33.4$\pm$}0.2 \textbf{\scriptsize{\color{red}+1.3}} 
			& \cellcolor{Gray} \textbf{35.7$\pm$}0.2 \textbf{\scriptsize{\color{red}+1.7}} 
			& \cellcolor{Gray} \textbf{18.3$\pm$}0.4 \textbf{\scriptsize{\color{red}+1.5}} 
			& \cellcolor{Gray} \textbf{34.5$\pm$}0.6 \textbf{\scriptsize{\color{red}+4.9}} \\ 

			\midrule
			\multirow{3}{*}{30} & FRCN+ft \cite{yan2019metarcnn} &\emph{ICCV 19}& 13.5$\pm$1.0 & 21.8$\pm$1.9 & 14.5$\pm$1.0 & 15.6$\pm$1.0 & 7.4$\pm$1.1 &13.1\\
			& TFA \cite{wang2020tfa} &\emph{ICML 20}& 28.7$\pm$0.4 &44.7$\pm$0.7 & 31.5$\pm$0.4 & 34.2$\pm$0.4 &12.1$\pm$0.4  &22.0\\
			& {DeFRCN}  \cite{qiao2021defrcn} &\emph{ICCV 21} 
			&31.4$\pm$0.1
			&48.8$\pm$0.2
			&33.9$\pm$0.1
		        &34.8$\pm$0.1
			&21.2$\pm$0.4 
			&36.7$\pm$0.8 \\
			&\cellcolor{Gray} {\textbf{Ours}} &\cellcolor{Gray} {}
			& \cellcolor{Gray} \textbf{32.3$\pm$}0.2 \textbf{\scriptsize{\color{red}+0.9}} 
			& \cellcolor{Gray} \textbf{51.3$\pm$}0.3 \textbf{\scriptsize{\color{red}+2.5}}
			& \cellcolor{Gray} \textbf{34.5$\pm$}0.2 \textbf{\scriptsize{\color{red}+0.6}} 
			& \cellcolor{Gray} \textbf{35.8$\pm$}0.2 \textbf{\scriptsize{\color{red}+1.0}} 
			& \cellcolor{Gray} \textbf{21.9$\pm$}0.3 \textbf{\scriptsize{\color{red}+0.7}} 
			& \cellcolor{Gray} \textbf{40.2$\pm$}0.5 \textbf{\scriptsize{\color{red}+3.5}} \\ 
			\bottomrule
	\end{tabular}}
	\end{table}

\begin{table}[htb] 
         \centering
	\caption{gFSOD performance ($\rm{AP}$ and $\rm{AP50}$) on PASCAL VOC dataset.}	\label{tab:gfsod-voc-full}
	\scalebox{0.7}{
		\begin{tabular}{c|c|c|ccc|c|cc}
			\toprule
			\multirow{2}{*}{Set} & \multirow{2}{*}{\# shots} &\multirow{2}{*}{Method} &  \multicolumn{3}{c|}{Overall \textit{\#20}}  &\multicolumn{1}{c|}{Base \textit{\#15}} & \multicolumn{2}{c}{Novel \textit{\#5}} \\ \cmidrule{4-9}
			& & & AP & AP50 & AP75 & AP & AP & AP50 \\ \midrule
			\multirow{21}{*}{Set 1} & 
			\multirow{4}{*}{1}
			& FSRW~\cite{kang2019frsw} & 27.6$ \pm $0.5 & 50.8$\pm $0.9 & 26.5$ \pm $0.6 & 34.1$ \pm $0.5 & 8.0$ \pm $1.0 &14.2\\
			& & FRCN+ft \cite{yan2019metarcnn}& 30.2$\pm$0.6 & 49.4$\pm$0.7 & 32.2$\pm$0.9 & 38.2$\pm$0.8 & 6.0$\pm$0.7 &9.9\\
			& & TFA \cite{wang2020tfa} & 40.6$\pm$0.5 & 64.5$\pm$0.6 & 44.7$\pm$0.6 &49.4$\pm$0.4 & 14.2$\pm$1.4  &25.3\\ 
			& & {DeFRCN} \cite{qiao2021defrcn}
			&42.0$\pm$0.6
			&66.7$\pm$0.8
			&45.5$\pm$0.7  
			&48.4$\pm$0.4 
			&22.5$\pm$1.7  
			&40.2\\
			& &\cellcolor{Gray} \textbf{Ours}
			& \cellcolor{Gray}\textbf{43.5$\pm$}0.9 \textbf{\scriptsize{\color{red}+1.5}} 
			& \cellcolor{Gray} \textbf{69.7$\pm$}1.4 \textbf{\scriptsize{\color{red}+3.0}} 
			& \cellcolor{Gray} \textbf{47.2$\pm$}1.1 \textbf{\scriptsize{\color{red}+1.7}} 
			& \cellcolor{Gray} \textbf{49.5$\pm$}0.7  \textbf{\scriptsize{\color{red}+1.1}} 
			& \cellcolor{Gray} \textbf{25.6$\pm$}2.6 \textbf{\scriptsize{\color{red}+3.1}} 
			& \cellcolor{Gray} \textbf{45.8$\pm$}4.5 \textbf{\scriptsize{\color{red}+5.6}} \\
			\cmidrule{2-9}
			& \multirow{4}{*}{2} & FSRW~\cite{kang2019frsw} & 
			28.7$\pm$0.4&52.2$\pm$0.6&27.7$\pm$0.5&33.9$\pm$0.4&13.2$\pm$1.0 &23.6\\
			& & FRCN+ft \cite{yan2019metarcnn} & 30.5$\pm$0.6 & 49.4$\pm$0.8 & 32.6$\pm$0.7 & 37.3$\pm$0.7 & 9.9$\pm$0.9  &15.6\\
			& &TFA \cite{wang2020tfa} & 42.6$\pm$0.3 & 67.1$\pm$0.4 &47.0$\pm$0.4 &{49.6$\pm$0.3} &21.7$\pm$1.0 &36.4 \\ 
			& & {DeFRCN} \cite{qiao2021defrcn}
			& {44.3$\pm$0.4} 
			& {70.2$\pm$0.5}
			& {48.0$\pm$0.6}
			& 49.1$\pm$0.3
			& {30.6$\pm$1.2} 
			&53.6\\
			& &\cellcolor{Gray} \textbf{Ours}
			& \cellcolor{Gray}\textbf{45.6$\pm$}0.5 \textbf{\scriptsize{\color{red}+1.3}} 
			& \cellcolor{Gray} \textbf{73.2$\pm$}0.8 \textbf{\scriptsize{\color{red}+3.0}}
			& \cellcolor{Gray} \textbf{49.2$\pm$}0.8 \textbf{\scriptsize{\color{red}+1.2}}
			& \cellcolor{Gray} \textbf{49.7$\pm$}0.4 \textbf{\scriptsize{\color{red}+0.6}}
			& \cellcolor{Gray} \textbf{33.4$\pm$}1.6 \textbf{\scriptsize{\color{red}+2.8}} 
			& \cellcolor{Gray} \textbf{59.1$\pm$}2.7 \textbf{\scriptsize{\color{red}+5.5}}\\
			\cmidrule{2-9}
			& \multirow{4}{*}{3} & FSRW~\cite{kang2019frsw} & 29.5$\pm$0.3&53.3$\pm$0.6&28.6$\pm$0.4&33.8$\pm$0.3&16.8$\pm$0.9 &29.8\\
			& & FRCN+ft \cite{yan2019metarcnn} & 31.8$\pm$0.5 & 51.4$\pm$0.8 & 34.2$\pm$0.6 & 37.9$\pm$0.5  & 13.7$\pm$1.0  &21.6 \\
			& &TFA  \cite{wang2020tfa} & 43.7$\pm$0.3 & 68.5$\pm$0.4 & 48.3$\pm$0.4 &{49.8$\pm$0.3} &25.4$\pm$0.9  &42.1\\			
			& &{DeFRCN} \cite{qiao2021defrcn}
			& {45.3$\pm$0.3} 
			&  {71.5$\pm$0.4}
			&  {49.0$\pm$0.5}
			& 49.3$\pm$0.3 
			& 33.7$\pm$0.8 
			& 58.2\\ 
			& &\cellcolor{Gray} \textbf{Ours} 
			& \cellcolor{Gray}\textbf{46.4$\pm$}0.6 \textbf{\scriptsize{\color{red}+1.1}} 
			& \cellcolor{Gray} \textbf{74.1$\pm$}0.6 \textbf{\scriptsize{\color{red}+2.6}} 
			& \cellcolor{Gray} \textbf{50.1$\pm$}0.8 \textbf{\scriptsize{\color{red}+1.1}} 
			& \cellcolor{Gray} 50.0$\pm$0.5 \textbf{\scriptsize{\color{red}+0.7}} 
			& \cellcolor{Gray} \textbf{35.5$\pm$}1.6  \textbf{\scriptsize{\color{red}+1.8}} 
			& \cellcolor{Gray} \textbf{62.1$\pm$}2.1  \textbf{\scriptsize{\color{red}+3.9}} 

			\\ 
			\cmidrule{2-9}
			& \multirow{4}{*}{5} & FSRW~\cite{kang2019frsw} & 30.4$\pm$0.3&54.6$\pm$0.5&29.6$\pm$0.4&33.7$\pm$0.3&20.6$\pm$0.8  &36.5\\
			& & FRCN+ft \cite{yan2019metarcnn} & 32.7$\pm$0.5 & 52.5$\pm$0.8 & 35.0$\pm$0.6 & 37.6$\pm$0.4  & 17.9$\pm$1.1 &28.0 \\
			& &TFA  \cite{wang2020tfa} & 44.8$\pm$0.3 &70.1$\pm$0.4 & 49.4$\pm$0.4 &{50.1$\pm$0.2 }& 28.9$\pm$0.8  &47.9\\
			& & {DeFRCN} \cite{qiao2021defrcn}
			& {46.4$\pm$0.3}
			&  {73.1$\pm$0.3}
			&  {50.4$\pm$0.4}
			&  49.6$\pm$0.3
			&  {37.3$\pm$0.8 } 
			& 63.6\\ 
			& &\cellcolor{Gray} \textbf{Ours} 
			& \cellcolor{Gray}\textbf{47.5$\pm$}0.5 \textbf{\scriptsize{\color{red}+1.1}} 
			& \cellcolor{Gray} \textbf{75.3$\pm$}0.4 \textbf{\scriptsize{\color{red}+2.2}} 
			& \cellcolor{Gray} \textbf{51.4$\pm$}0.6 \textbf{\scriptsize{\color{red}+1.0}} 
			& \cellcolor{Gray} \textbf{50.4$\pm$}0.4 \textbf{\scriptsize{\color{red}+0.8}} 
			& \cellcolor{Gray} \textbf{38.6$\pm$}0.8 \textbf{\scriptsize{\color{red}+1.3}} 
			& \cellcolor{Gray} \textbf{66.8$\pm$}0.8 \textbf{\scriptsize{\color{red}+3.2}} 

			\\ 

			\cmidrule{2-9}
			& \multirow{3}{*}{10} 
			& FRCN+ft \cite{wang2020tfa} & 33.3$\pm$0.4 & 53.8$\pm$0.6 & 35.5$\pm$0.4 & 36.8$\pm$0.4 & 22.7$\pm$0.9 &52.0\\
			& &TFA  \cite{wang2020tfa} & 45.8$\pm$0.2 & 71.3$\pm$0.3 & 50.4$\pm$0.3 &{50.4$\pm$0.2} & 32.0$\pm$0.6  &52.8 \\ 
			& & {DeFRCN} \cite{qiao2021defrcn}
			&{47.2$\pm$0.2} 
			&{74.0$\pm$0.3}
			&{51.3$\pm$0.3} 
			&49.9$\pm$0.2
			&\textbf{39.8$\pm$}0.7 &66.5\\ 
			& &\cellcolor{Gray} \textbf{Ours} 
			& \cellcolor{Gray}\textbf{47.7$\pm$}0.3 \textbf{\scriptsize{\color{red}+0.5}} 
			& \cellcolor{Gray} \textbf{75.5$\pm$}0.4 \textbf{\scriptsize{\color{red}+1.5}} 
			& \cellcolor{Gray} \textbf{51.8$\pm$}0.6 \textbf{\scriptsize{\color{red}+0.5}} 
			& \cellcolor{Gray} \textbf{50.4$\pm$}0.3\textbf{\scriptsize{\color{red}+0.5}} 
			& \cellcolor{Gray} {39.7$\pm$}0.9 \textbf{\scriptsize{\color{blue}-0.1}}   
			& \cellcolor{Gray} \textbf{68.0$\pm$}1.3 \textbf{\scriptsize{\color{red}+1.5}}   
			\\ 
						\midrule
			\multirow{21}{*}{Set 2} & \multirow{4}{*}{1} & FSRW~\cite{kang2019frsw} &
			28.4$\pm$0.5&51.7$\pm$0.9&27.3$\pm$0.6&35.7$\pm$0.5&6.3$\pm$0.9 &12.3\\
			& & FRCN+ft \cite{yan2019metarcnn} & 30.3$\pm$0.5 & 49.7$\pm$0.5 & 32.3$\pm$0.7 & 38.8$\pm$0.6 & 5.0$\pm$0.6  &9.4 \\
			& &TFA  \cite{wang2020tfa} &36.7$\pm$0.6 & 59.9$\pm$0.8 & 39.3$\pm$0.8 & 45.9$\pm$0.7 & 9.0$\pm$1.2 &18.3 \\
			& &{DeFRCN} \cite{qiao2021defrcn}
			&{40.7$\pm$}0.5
			&{64.8$\pm$}0.7
			&{43.8$\pm$}0.6
			&{49.6$\pm$}0.4
			&{14.6$\pm$}1.5 &29.5\\
			& &\cellcolor{Gray} \textbf{Ours} 
			& \cellcolor{Gray}\textbf{41.7$\pm$}0.9 \textbf{\scriptsize{\color{red}+1.0}} 
			& \cellcolor{Gray} \textbf{66.8$\pm$}1.1 \textbf{\scriptsize{\color{red}+2.0}} 
			& \cellcolor{Gray} \textbf{44.5$\pm$}1.1 \textbf{\scriptsize{\color{red}+0.7}} 
			& \cellcolor{Gray} \textbf{50.5$\pm$}0.9 \textbf{\scriptsize{\color{red}+0.9}} 
			& \cellcolor{Gray} \textbf{15.1$\pm$}2.3 \textbf{\scriptsize{\color{red}+0.5}}
			& \cellcolor{Gray} \textbf{31.8$\pm$}3.7 \textbf{\scriptsize{\color{red}+2.3}}

			 \\ 
			 \cmidrule{2-9}
			& \multirow{4}{*}{2} & FSRW~\cite{kang2019frsw} & 
			29.4$\pm$0.3&53.1$\pm$0.6&28.5$\pm$0.4&35.8$\pm$0.4&9.9$\pm$0.7 &19.6\\
			& & FRCN+ft \cite{yan2019metarcnn} & 30.7$\pm$0.5 & 49.7$\pm$0.7 & 32.9$\pm$0.6 & 38.4$\pm$0.5 & 7.7$\pm$0.8 &13.8\\
			& & TFA  \cite{wang2020tfa} & 39.0$\pm$0.4 & 63.0$\pm$0.5 & 42.1$\pm$0.6 &47.3$\pm$0.4 &14.1$\pm$0.9 &27.5 \\
			& &{DeFRCN} \cite{qiao2021defrcn}
			&{42.7$\pm$}0.3
			&{67.7$\pm$}0.5 
			&{45.7$\pm$}0.5
			&{50.3$\pm$}0.2 
			&{20.5$\pm$}1.0 &39.7\\ 
			& &\cellcolor{Gray} \textbf{Ours} 
			& \cellcolor{Gray}\textbf{43.6$\pm$}0.7 \textbf{\scriptsize{\color{red}+0.9}} 
			& \cellcolor{Gray} \textbf{69.6$\pm$}0.9 \textbf{\scriptsize{\color{red}+1.9}} 
			& \cellcolor{Gray} \textbf{46.6$\pm$}1.0 \textbf{\scriptsize{\color{red}+0.9}} 
			& \cellcolor{Gray} \textbf{51.1$\pm$}0.4 \textbf{\scriptsize{\color{red}+0.8}} 
			& \cellcolor{Gray} \textbf{21.2$\pm$}1.9 \textbf{\scriptsize{\color{red}+0.7}} 
			& \cellcolor{Gray} \textbf{41.7$\pm$}2.5 \textbf{\scriptsize{\color{red}+2.0}} 
			\\ 
			\cmidrule{2-9}
			& \multirow{4}{*}{3} & FSRW~\cite{kang2019frsw} & 29.9$\pm$0.3&53.9$\pm$0.4&29.0$\pm$0.4&35.7$\pm$0.3&12.5$\pm$0.7 &25.1 \\
			& & FRCN+ft \cite{yan2019metarcnn} & 31.1$\pm$0.3 & 50.1$\pm$0.5 & 33.2$\pm$0.5 & 38.1$\pm$0.4 & 9.8$\pm$0.9  &17.4\\
			& & TFA  \cite{wang2020tfa} &40.1$\pm$0.3 & 64.5$\pm$0.5 & 43.3$\pm$0.4 & 48.1$\pm$0.3 & 16.0$\pm$0.8  &30.9\\
			& &\ {DeFRCN} \cite{qiao2021defrcn}
			&{43.5$\pm$0.3} 
			&{68.9$\pm$0.4} 
			 &{46.6$\pm$0.4}
			 &{50.6$\pm$0.3}
			 &{22.9$\pm$1.0} &43.4\\
			& &\cellcolor{Gray} \textbf{Ours} 
			& \cellcolor{Gray}\textbf{44.6$\pm$}0.6 \textbf{\scriptsize{\color{red}+1.1}} 
			& \cellcolor{Gray} \textbf{70.9$\pm$}0.6 \textbf{\scriptsize{\color{red}+2.0}} 
			& \cellcolor{Gray} \textbf{47.7$\pm$}0.7 \textbf{\scriptsize{\color{red}+1.1}} 
			& \cellcolor{Gray} \textbf{51.4$\pm$}0.5 \textbf{\scriptsize{\color{red}+0.8}} 
			& \cellcolor{Gray} \textbf{24.4$\pm$}1.2 \textbf{\scriptsize{\color{red}+1.5}} 
			& \cellcolor{Gray} \textbf{46.6$\pm$}1.8 \textbf{\scriptsize{\color{red}+3.2}} 
			\\ 
			\cmidrule{2-9}
			& \multirow{4}{*}{5} & FSRW~\cite{kang2019frsw} & 30.4$\pm$0.4&54.6$\pm$0.5&29.5$\pm$0.5&35.3$\pm$0.3&15.7$\pm$0.8 &31.4\\
			& & FRCN+ft \cite{yan2019metarcnn} & 31.5$\pm$0.3 & 50.8$\pm$0.7 & 33.6$\pm$0.4 & 37.9$\pm$0.4 & 12.4$\pm$0.9 &21.9\\
			& & TFA  \cite{wang2020tfa} & 40.9$\pm$0.4 &65.7$\pm$0.5 & 44.1$\pm$0.5 & 48.6$\pm$0.4  & 17.8$\pm$0.8  &34.1\\
			& & {DeFRCN} \cite{qiao2021defrcn}
			& {44.6$\pm$0.3} 
			& {70.2$\pm$0.5} 
			& {47.8$\pm$0.4}
			& {51.0$\pm$0.2} 
			& {25.8$\pm$0.9} &48.1 \\ 
			& &\cellcolor{Gray} \textbf{Ours} 
			& \cellcolor{Gray}\textbf{45.2$\pm$}0.4 \textbf{\scriptsize{\color{red}+0.6}} 
			& \cellcolor{Gray} \textbf{71.6$\pm$}0.5 \textbf{\scriptsize{\color{red}+1.4}} 
			& \cellcolor{Gray} \textbf{48.3$\pm$}0.6 \textbf{\scriptsize{\color{red}+0.5}} 
			& \cellcolor{Gray} \textbf{51.5$\pm$}0.4 \textbf{\scriptsize{\color{red}+0.5}} 
			& \cellcolor{Gray} \textbf{26.4$\pm$}0.8 \textbf{\scriptsize{\color{red}+0.6}}
			& \cellcolor{Gray} \textbf{50.3$\pm$}1.3 \textbf{\scriptsize{\color{red}+2.2}}
			 \\ 
    
			\cmidrule{2-9}
			& \multirow{3}{*}{10} & FRCN+ft \cite{wang2020tfa} & 32.2$\pm$0.3 & 52.3$\pm$0.4 & 34.1$\pm$0.4 & 37.2$\pm$0.3 & 17.0$\pm$0.8 &39.7\\
			& & TFA  \cite{wang2020tfa} & 42.3$\pm$0.3 & 67.6$\pm$0.4 & 45.7$\pm$0.3 & 49.4$\pm$0.2 & 20.8$\pm$0.6  &39.5\\ 
			& & {DeFRCN} \cite{qiao2021defrcn}
			&{45.6$\pm$0.2}
			&{71.5$\pm$0.3}
			&{49.0$\pm$0.3}
			&{51.3$\pm$0.2}  
			&\textbf{29.3}$\pm$0.7 &52.8\\
			& &\cellcolor{Gray} \textbf{Ours} 
			& \cellcolor{Gray} \textbf{45.9$\pm$}0.3 \textbf{\scriptsize{\color{red}+0.3}} 
			& \cellcolor{Gray} \textbf{72.5$\pm$}0.3 \textbf{\scriptsize{\color{red}+1.0}} 
			& \cellcolor{Gray} \textbf{49.1$\pm$}0.5 \textbf{\scriptsize{\color{red}+0.1}} 
			& \cellcolor{Gray} \textbf{51.5$\pm$}0.2 \textbf{\scriptsize{\color{red}+0.2}} 
			& \cellcolor{Gray} 29.1$\pm$0.8 \textbf{\scriptsize{\color{blue}-0.2}} 
			& \cellcolor{Gray} \textbf{53.7$\pm$}1.1 \textbf{\scriptsize{\color{red}+0.9}} 
			\\ 

			 \midrule
			\multirow{21}{*}{Set 3} & \multirow{4}{*}{1} & FSRW~\cite{kang2019frsw} &
			27.5$\pm$0.6&50.0$\pm$1.0&26.8$\pm$0.7&34.5$\pm$0.7&6.7$\pm$1.0 &12.5\\
			& & FRCN+ft \cite{yan2019metarcnn} & 30.8$\pm$0.6 & 49.8$\pm$0.8 & 32.9$\pm$0.8 & 39.6$\pm$0.8 & 4.5$\pm$0.7 &8.1\\
			& & TFA  \cite{wang2020tfa} & 40.1$\pm$0.3 & 63.5$\pm$0.6 & 43.6$\pm$0.5 &{50.2$\pm$0.4} &9.6$\pm$1.1  &17.9 \\ 			
			& & {DeFRCN} \cite{qiao2021defrcn}
			&41.6$\pm$0.5
			&66.0$\pm$0.9
			&44.9$\pm$0.6 
			&49.4$\pm$0.4
			&17.9$\pm$1.6  &35.0  \\
			& &\cellcolor{Gray} \textbf{Ours}
			& \cellcolor{Gray}\textbf{43.3$\pm$}1.0 \textbf{\scriptsize{\color{red}+1.7}} 
			& \cellcolor{Gray} \textbf{69.1$\pm$}1.7 \textbf{\scriptsize{\color{red}+3.1}} 
			& \cellcolor{Gray} \textbf{46.8$\pm$}1.1 \textbf{\scriptsize{\color{red}+1.9}} 
			& \cellcolor{Gray} \textbf{50.9$\pm$}0.6 \textbf{\scriptsize{\color{red}+1.5}} 
			& \cellcolor{Gray} \textbf{20.5$\pm$}3.7 \textbf{\scriptsize{\color{red}+2.6}} 
			& \cellcolor{Gray} \textbf{39.6$\pm$}6.2 \textbf{\scriptsize{\color{red}+4.6}} \\

			 \cmidrule{2-9}
			& \multirow{4}{*}{2} & FSRW~\cite{kang2019frsw} & 28.7$\pm$0.4&51.8$\pm$0.7&28.1$\pm$0.5&34.5$\pm$0.4&11.3$\pm$0.7 &21.3\\
			& & FRCN+ft \cite{yan2019metarcnn} & 31.3$\pm$0.5 & 50.2$\pm$0.9 & 33.5$\pm$0.6 & 39.1$\pm$0.5 & 8.0$\pm$0.8 &13.9 \\
			& & TFA  \cite{wang2020tfa} &41.8$\pm$0.4 & 65.6$\pm$0.6 &45.3$\pm$0.4 &{50.7$\pm$0.3}& 15.1$\pm$1.3 &27.2 \\
			& &{DeFRCN} \cite{qiao2021defrcn}
			&{44.0$\pm$0.4} 
			&{69.5$\pm$0.7} 
			&{47.7$\pm$0.5} 
			&50.2$\pm$0.2 
			&{26.0$\pm$1.3 } &38.3\\ 
			& &\cellcolor{Gray} \textbf{Ours} 
			& \cellcolor{Gray}\textbf{45.3$\pm$}0.5 \textbf{\scriptsize{\color{red}+1.3}} 
			& \cellcolor{Gray} \textbf{72.3$\pm$}0.6 \textbf{\scriptsize{\color{red}+2.5}} 
			& \cellcolor{Gray} \textbf{48.6$\pm$}0.9 \textbf{\scriptsize{\color{red}+0.9}} 
			& \cellcolor{Gray} \textbf{51.3$\pm$}0.4 \textbf{\scriptsize{\color{red}+1.1}} 
			& \cellcolor{Gray} \textbf{27.6$\pm$}1.7 \textbf{\scriptsize{\color{red}+1.6}} 
			& \cellcolor{Gray} \textbf{52.1$\pm$}2.4 \textbf{\scriptsize{\color{red}+13.8}} \\ 				        

			\cmidrule{2-9}
			& \multirow{4}{*}{3} & FSRW~\cite{kang2019frsw} & 
			29.2$\pm$0.4&52.7$\pm$0.6&28.5$\pm$0.4&34.2$\pm$0.3&14.2$\pm$0.7&26.8 \\
			& & FRCN+ft \cite{yan2019metarcnn} & 32.1$\pm$0.5 & 51.3$\pm$0.8 & 34.3$\pm$0.6 & 39.1$\pm$0.5 & 11.1$\pm$0.9 &19.0\\
			& &TFA  \cite{wang2020tfa} & 43.1$\pm$0.4 & 67.5$\pm$0.5 & 46.7$\pm$0.5 &\textbf{51.1$\pm$0.3 }& 18.9$\pm$1.1 &34.3 \\
			& &{DeFRCN} \cite{qiao2021defrcn}
			&{45.1$\pm$0.3}
			&{70.9$\pm$0.5}
			&{48.8$\pm$0.4} 
			&50.5$\pm$0.2 
			&{29.2$\pm$1.0} &52.9 \\ 
			& &\cellcolor{Gray} \textbf{Ours} 
			& \cellcolor{Gray}\textbf{46.2$\pm$}0.4 \textbf{\scriptsize{\color{red}+1.1}} 
			& \cellcolor{Gray} \textbf{73.4$\pm$}0.5 \textbf{\scriptsize{\color{red}+2.5}} 
			& \cellcolor{Gray} \textbf{49.4$\pm$}0.6 \textbf{\scriptsize{\color{red}+0.6}} 
			& \cellcolor{Gray} \textbf{51.5$\pm$}0.3 \textbf{\scriptsize{\color{red}+1.0}} 
			& \cellcolor{Gray} \textbf{30.5$\pm$}1.0 \textbf{\scriptsize{\color{red}+1.3}} 
			& \cellcolor{Gray} \textbf{56.3$\pm$}1.9 \textbf{\scriptsize{\color{red}+3.4}} 
			\\ 

			\cmidrule{2-9}
			& \multirow{4}{*}{5} & FSRW~\cite{kang2019frsw} & 
			30.1$\pm$0.3&53.8$\pm$0.5&29.3$\pm$0.4&34.1$\pm$0.3&18.0$\pm$0.7 &33.8\\
			& & FRCN+ft \cite{yan2019metarcnn} & 32.4$\pm$0.5 & 51.7$\pm$0.8 & 34.4$\pm$0.6 & 38.5$\pm$0.5 & 14.0$\pm$0.9 &23.9 \\
			& & TFA  \cite{wang2020tfa} & 44.1$\pm$0.3 & 69.1$\pm$0.4 & 47.8$\pm$0.4 &{51.3$\pm$0.2} & 22.8$\pm$0.9 &40.8 \\
			& & {DeFRCN} \cite{qiao2021defrcn}
			&{46.2$\pm$0.3} 
			&{72.4$\pm$0.4}  
			&{50.0$\pm$0.5} 
			& 51.0$\pm$0.2 
			&{32.3$\pm$0.9 } &57.7 \\ 
			& &\cellcolor{Gray} \textbf{Ours} 
			& \cellcolor{Gray} \textbf{47.2$\pm$}0.4 \textbf{\scriptsize{\color{red}+1.0}} 
			& \cellcolor{Gray} \textbf{74.5$\pm$}0.5 \textbf{\scriptsize{\color{red}+2.1}} 
			& \cellcolor{Gray} \textbf{50.8$\pm$}0.6 \textbf{\scriptsize{\color{red}+0.8}} 
			& \cellcolor{Gray} \textbf{51.8$\pm$}0.3 \textbf{\scriptsize{\color{red}+0.8}} 
			& \cellcolor{Gray} \textbf{33.5$\pm$}0.9 \textbf{\scriptsize{\color{red}+1.2}} 
			& \cellcolor{Gray} \textbf{60.3$\pm$}1.2 \textbf{\scriptsize{\color{red}+2.6}} 
			\\ 

			\cmidrule{2-9}
			& \multirow{3}{*}{10} & FRCN+ft \cite{yan2019metarcnn} & 33.1$\pm$0.5 & 53.1$\pm$0.7 & 35.2$\pm$0.5 & 38.0$\pm$0.5 & 18.4$\pm$0.8  &44.6 \\
			& & TFA  \cite{wang2020tfa} & 45.0$\pm$0.3 & 70.3$\pm$0.4 & 48.9$\pm$0.4 &{51.6$\pm$0.2} & 25.4$\pm$0.7 &45.6 \\			
			& &{DeFRCN} \cite{qiao2021defrcn}
			&{47.0$\pm$0.3} 
			&{ 73.3$\pm$0.3}
			&{51.0$\pm$0.4}
			&51.3$\pm$0.2 
			&{34.7$\pm$0.7}  &60.8 \\
			& &\cellcolor{Gray} \textbf{Ours} 
			& \cellcolor{Gray}\textbf{47.8$\pm$}0.3 \textbf{\scriptsize{\color{red}+0.8}} 
			& \cellcolor{Gray} \textbf{75.1$\pm$}0.3 \textbf{\scriptsize{\color{red}+1.8}} 
			& \cellcolor{Gray} \textbf{51.6$\pm$}0.5 \textbf{\scriptsize{\color{red}+0.6}} 
			& \cellcolor{Gray} \textbf{51.9$\pm$}0.2 \textbf{\scriptsize{\color{red}+0.6}} 
			& \cellcolor{Gray} \textbf{35.6$\pm$}1.2 \textbf{\scriptsize{\color{red}+0.9}} 
			& \cellcolor{Gray} \textbf{63.3$\pm$}1.2 \textbf{\scriptsize{\color{red}+2.5}} 
			\\ 

			\bottomrule
	\end{tabular}}
	\end{table}

\section{Qualitative Evaluation}\label{sec:qe}
\begin{figure}[h]
 \captionsetup[subfigure]{labelformat=empty}
  \centering
\subfloat[]  {\rotatebox{90} {\quad \quad \textbf{\scriptsize{Ours}}  \quad  \quad ~~  \textbf{\scriptsize{Mask-DeFRCN}}} \hspace{-3pt}}
  \subfloat[]
  {
 \includegraphics[height=0.30\columnwidth]{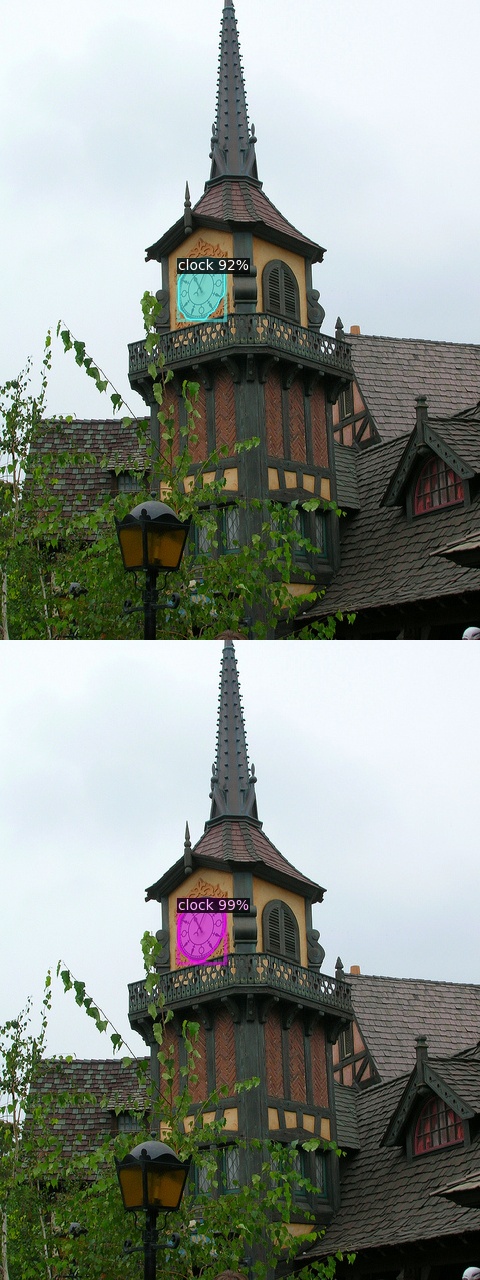}
 \hspace{-6pt}
  }
  \subfloat[]
  {
  \includegraphics[height=0.30\columnwidth]{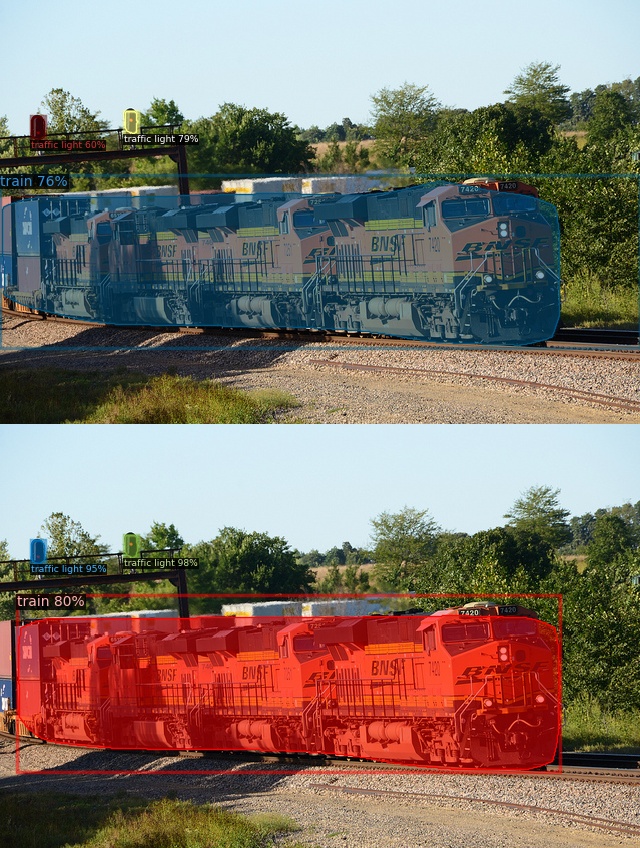}
   \hspace{-6pt}
  }
  \subfloat[]
  {
  \includegraphics[height=0.30\columnwidth]{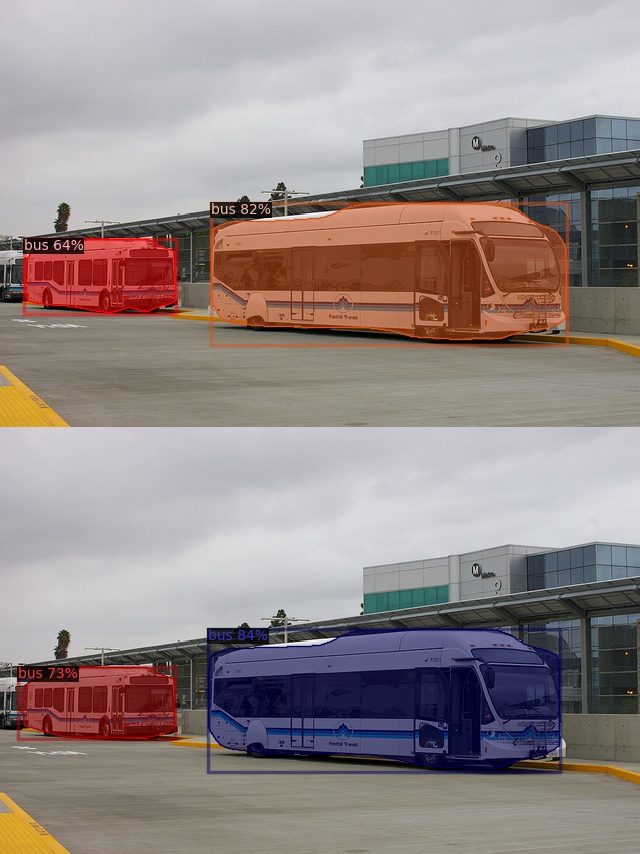}
   \hspace{-6pt}
  }
  \subfloat[]
  {
  \includegraphics[height= 0.30\columnwidth]{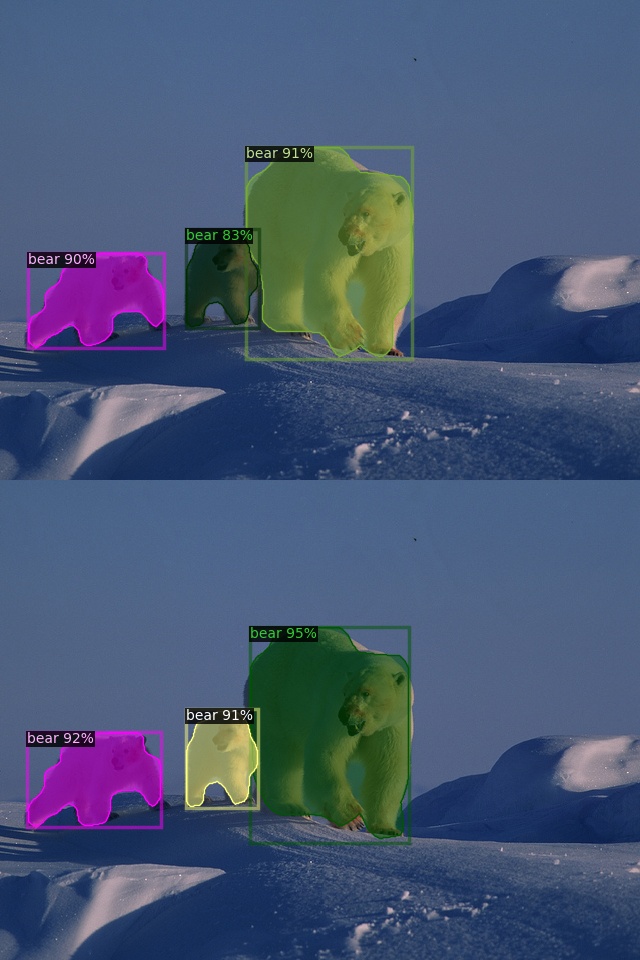}
   \hspace{-6pt}
  }
  \subfloat[]
  {
  \includegraphics[height= 0.30\columnwidth]{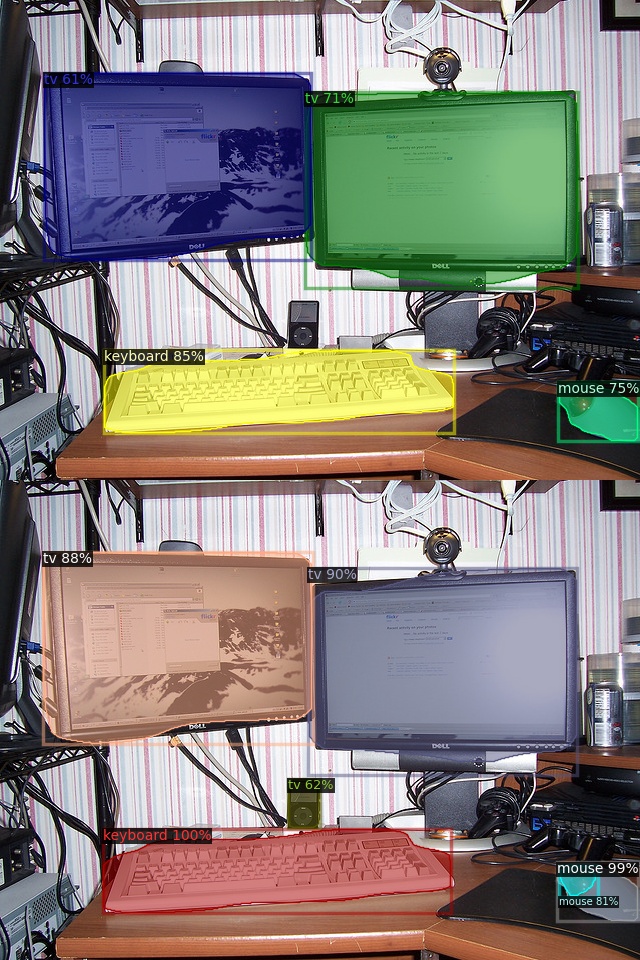}
   \hspace{-5pt}
  }\\ \vspace{-22pt}
   \subfloat[]   {\rotatebox{90}{\quad~~~\textbf{\scriptsize{Ours}}  \quad  \quad  ~  \textbf{\scriptsize{Mask-DeFRCN}}} \hspace{-3pt}}
  \subfloat[]
  {
  \includegraphics[height=0.27\columnwidth]{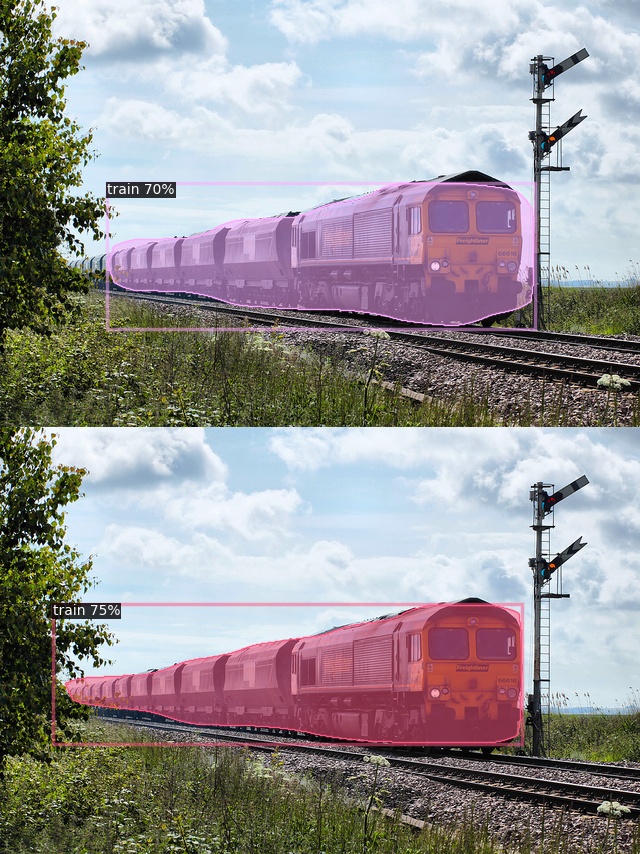}
  \hspace{-6pt}
  }
  \subfloat[]
  {
  \includegraphics[height=0.27\columnwidth]{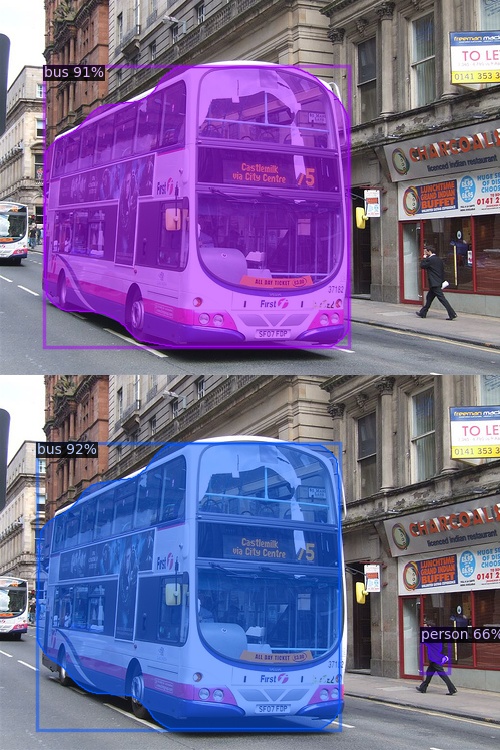}
  \hspace{-6pt}
  }
  \subfloat[]
  {
  \includegraphics[height=0.27\columnwidth]{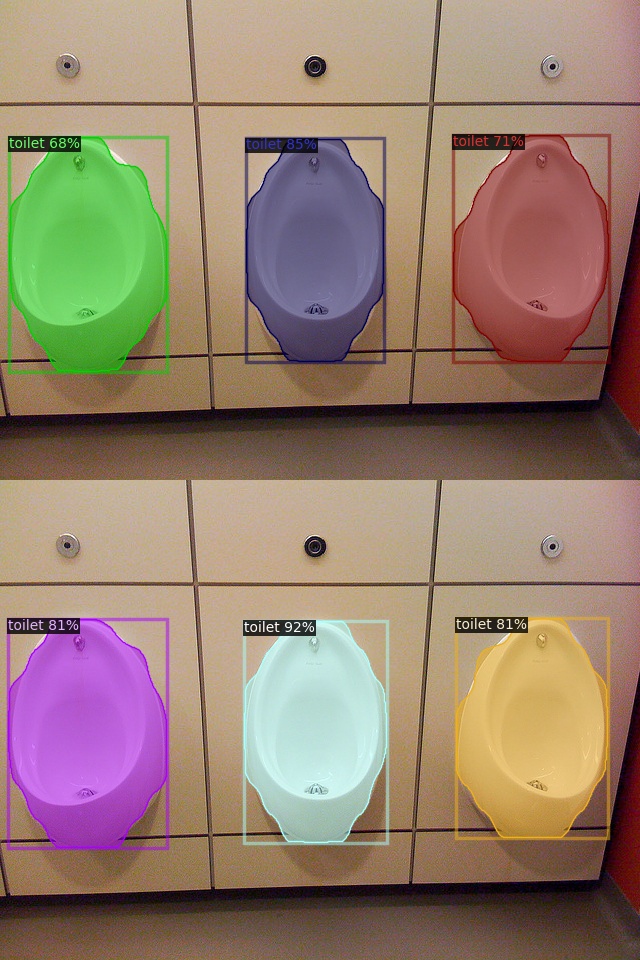}
  \hspace{-6pt}
  }
  \subfloat[]
  {
  \includegraphics[height=0.27\columnwidth]{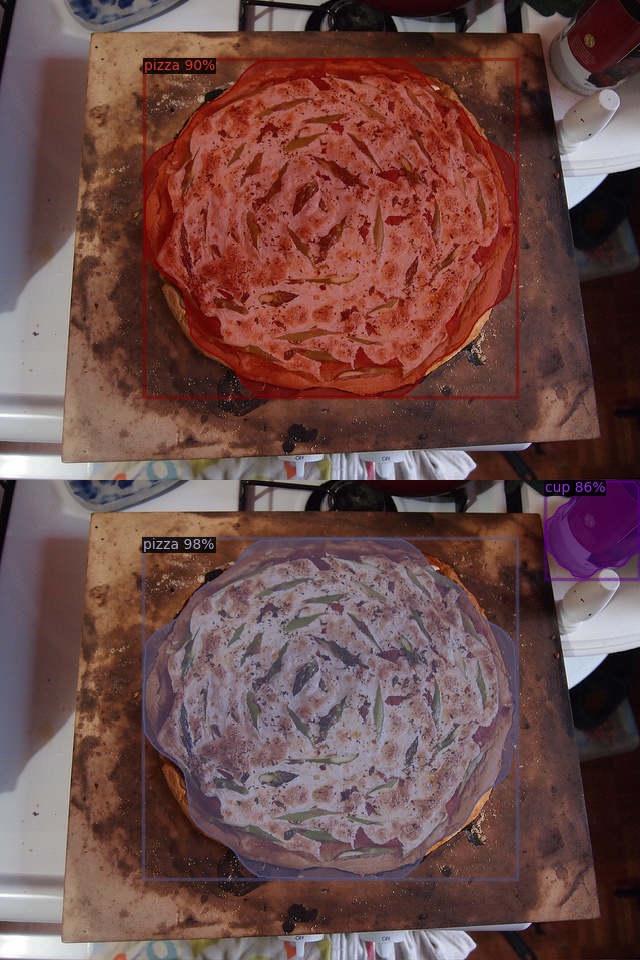}
  \hspace{-6pt}
  }
  \subfloat[]
  {
  \includegraphics[height=0.27\columnwidth]{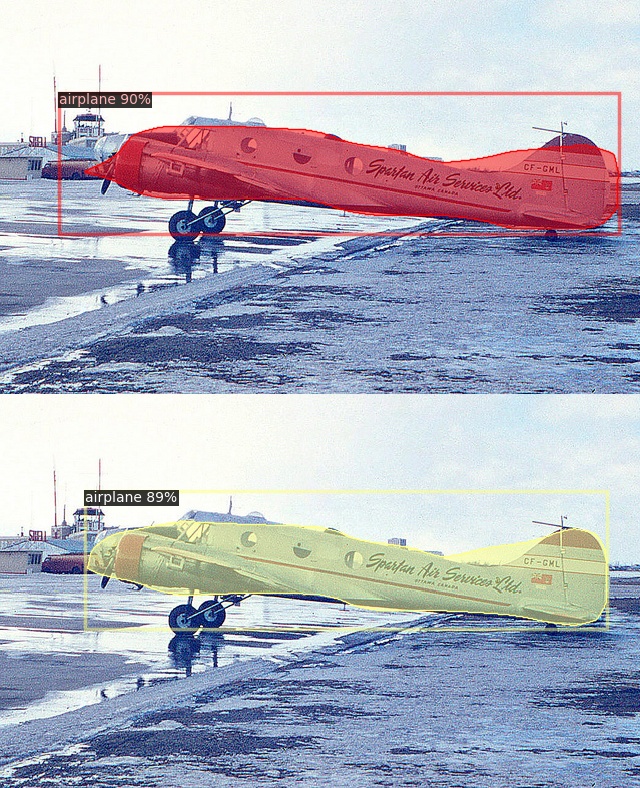}
  \hspace{-6pt}
  }
  \\
    \vspace{-10pt}\textcolor{red}{\hrule height 1.0pt width 1.005\textwidth }
    \vspace{-6pt}
   \subfloat[]   {\rotatebox{90}{\quad \quad \textbf{\scriptsize{Ours}}  \quad  \quad  \quad \textbf{\scriptsize{Mask-DeFRCN}}} \hspace{-3pt}}
  \subfloat[]
  {
  \includegraphics[height= 0.2925\columnwidth]{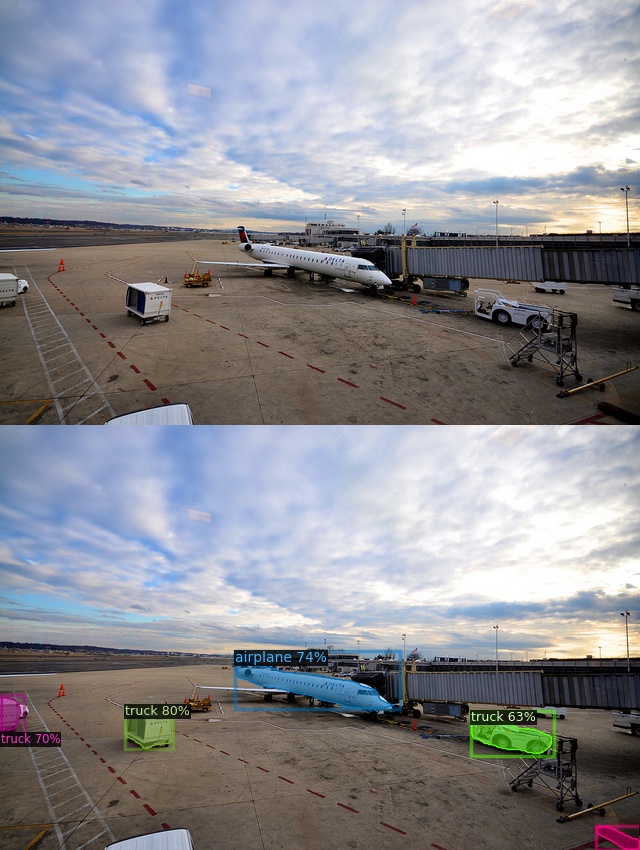}
  \hspace{-6pt}
  }
  \subfloat[]
  {
  \includegraphics[height=0.2925\columnwidth]{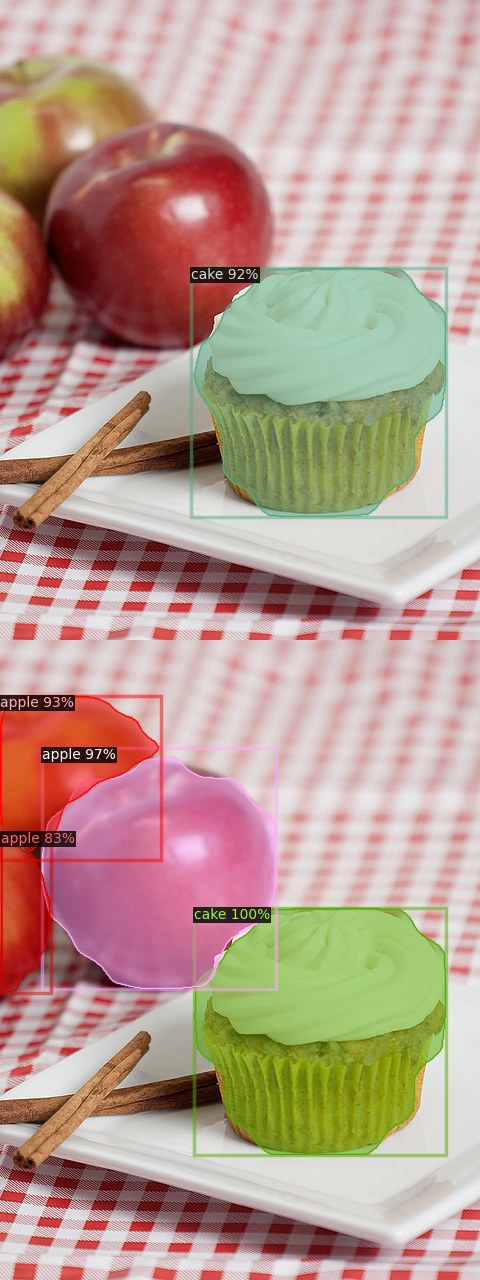}
  \hspace{-6pt}
  }  
  \subfloat[]
  {
  \includegraphics[height=0.2925\columnwidth]{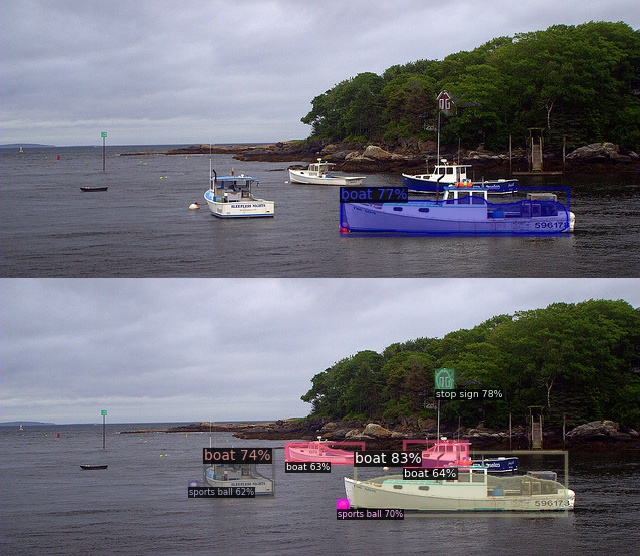}
  \hspace{-6pt}
  }
  \subfloat[]
  {
  \includegraphics[height= 0.2925\columnwidth]{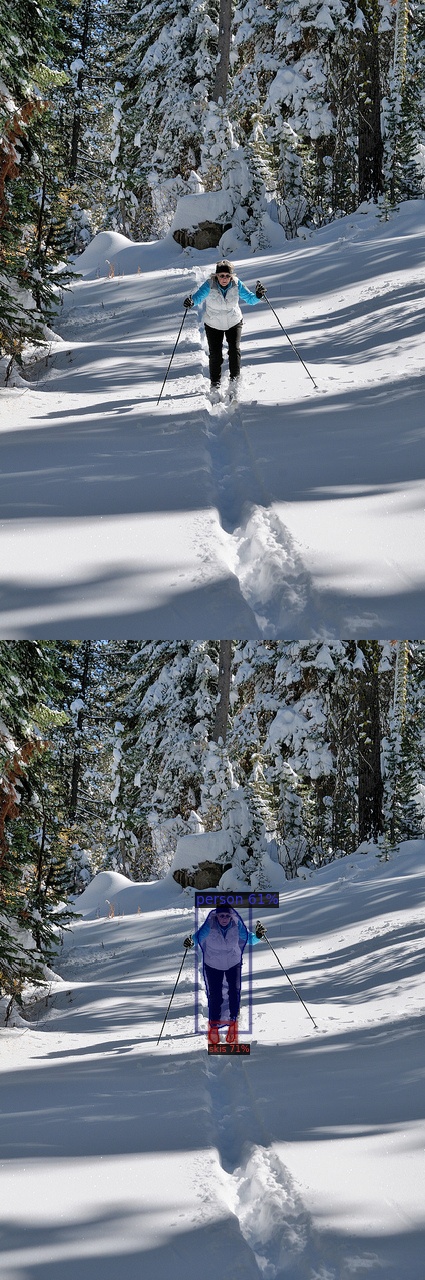}
  \hspace{-6pt}
  }
  \subfloat[]
  {
  \includegraphics[height= 0.2925\columnwidth]{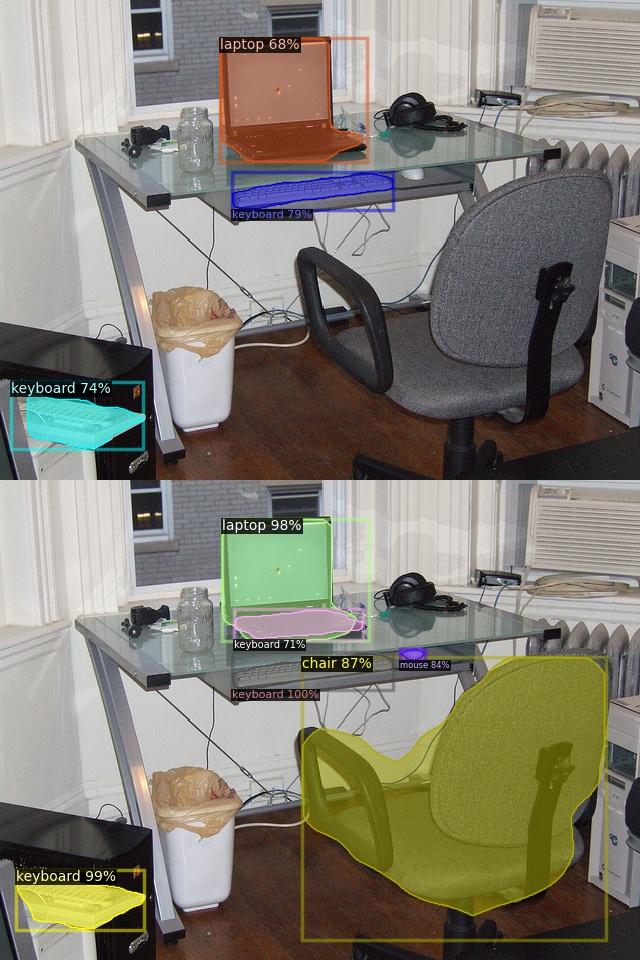}
  \hspace{-6pt}
  }
  \\
   \vspace{-22pt}
  \subfloat[]   {\rotatebox{90}{\quad~~\textbf{\scriptsize{Ours}}  \quad  \quad~~\textbf{\scriptsize{Mask-DeFRCN}}} \hspace{-3pt}}
  \subfloat[]
  {
  \includegraphics[height= 0.272\columnwidth]{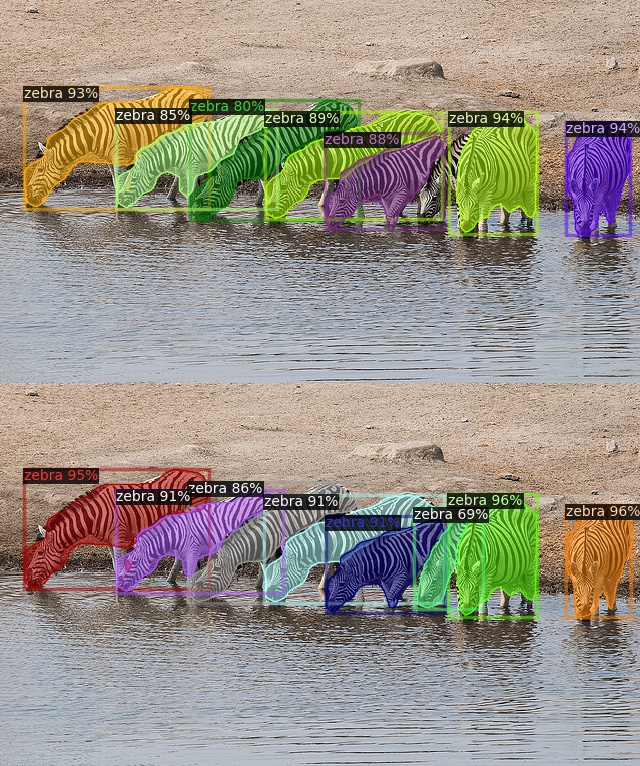}
  \hspace{-6pt}
  }
  \subfloat[]
  {
  \includegraphics[height= 0.272\columnwidth]{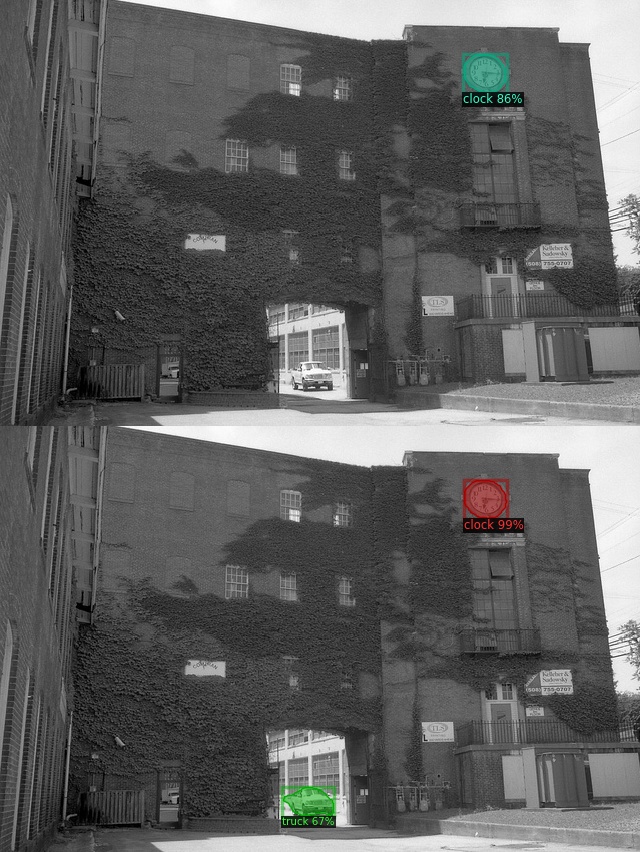}
  \hspace{-6pt}
  }
  \subfloat[]
  {
  \includegraphics[height=0.272\columnwidth]{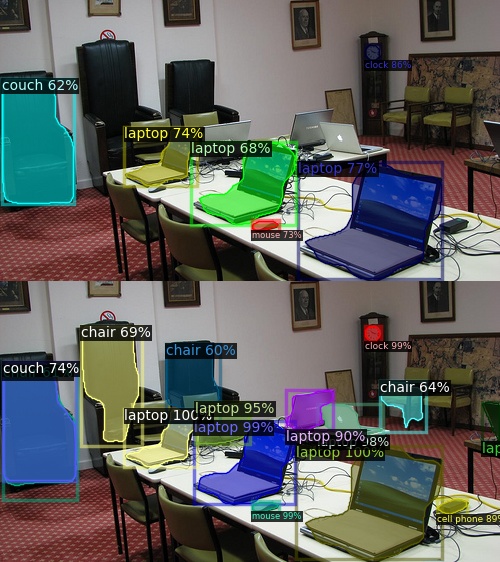}
  \hspace{-6pt}
  }
 \subfloat[]
  {
  \includegraphics[height=0.272\columnwidth]{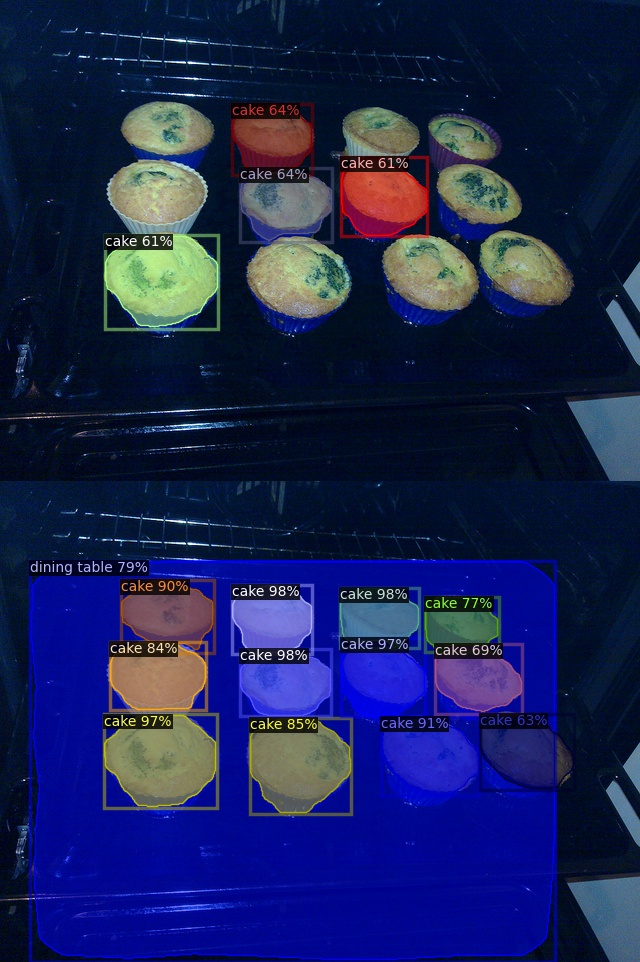}
  \hspace{-6pt}
  }
 \subfloat[]
  {
  \includegraphics[height=0.272\columnwidth]{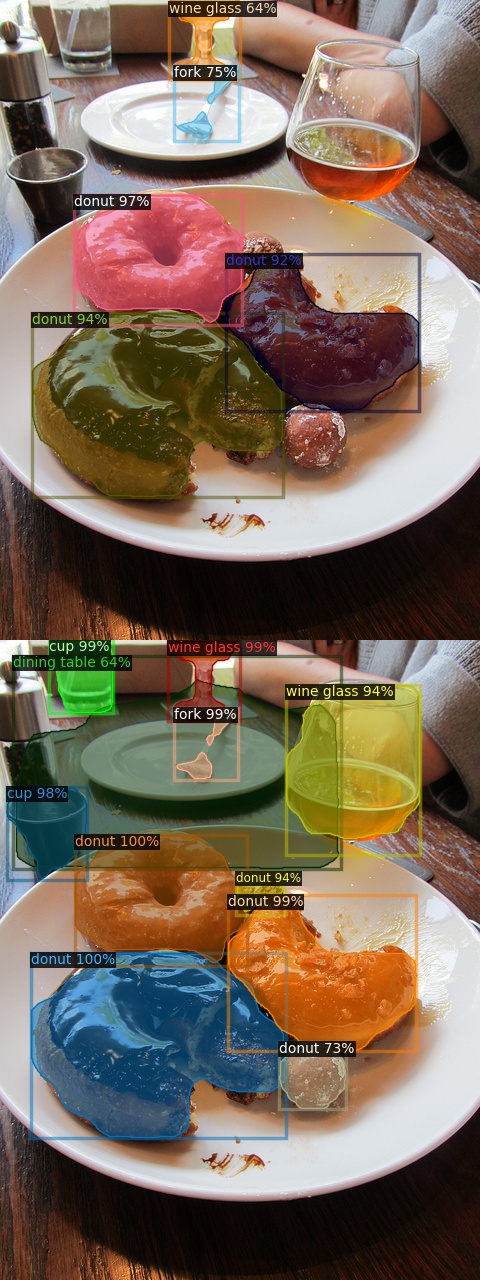}
   \hspace{-6pt}
  }
  \vspace{-10pt} \textcolor{red}{\hrule height 1.0pt width 1.005\textwidth }
  \vspace{-6pt}
  \subfloat[]   {\rotatebox{90}{\quad\textbf{\scriptsize{Ours}}  \quad  \quad\textbf{\scriptsize{Mask-DeFRCN}}} \hspace{-3pt}}
  \subfloat[]
  {
  \includegraphics[height= 0.2485\columnwidth]{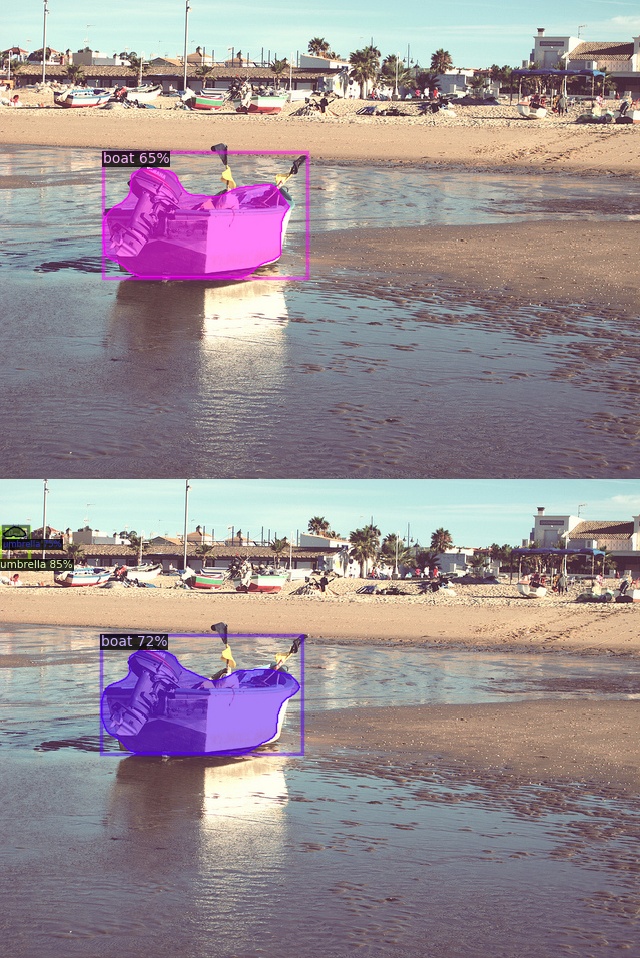}
   \hspace{-6pt}
  }
   \subfloat[]
  {
  \includegraphics[height=0.2485\columnwidth]{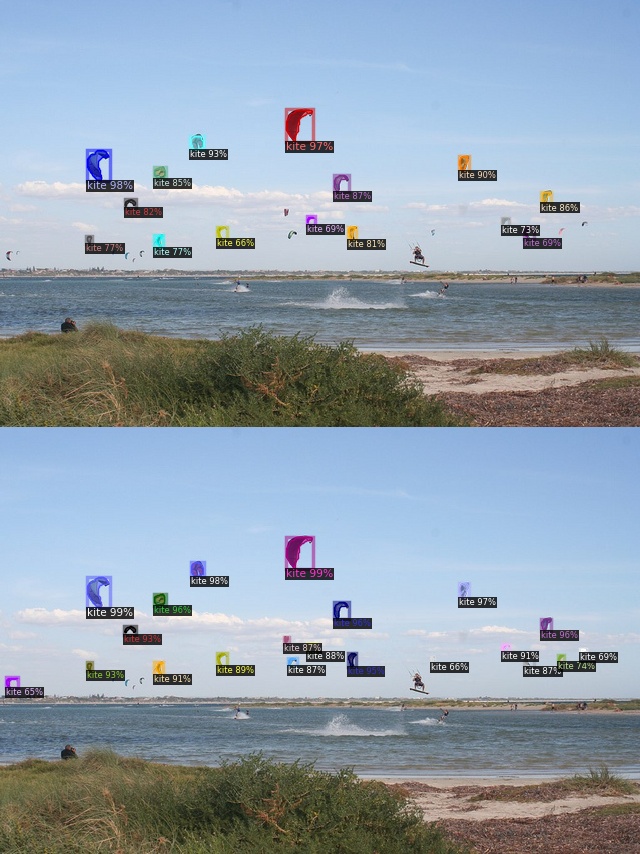}
   \hspace{-6pt}
  }
  \subfloat[]
  {
  \includegraphics[height=0.2485\columnwidth]{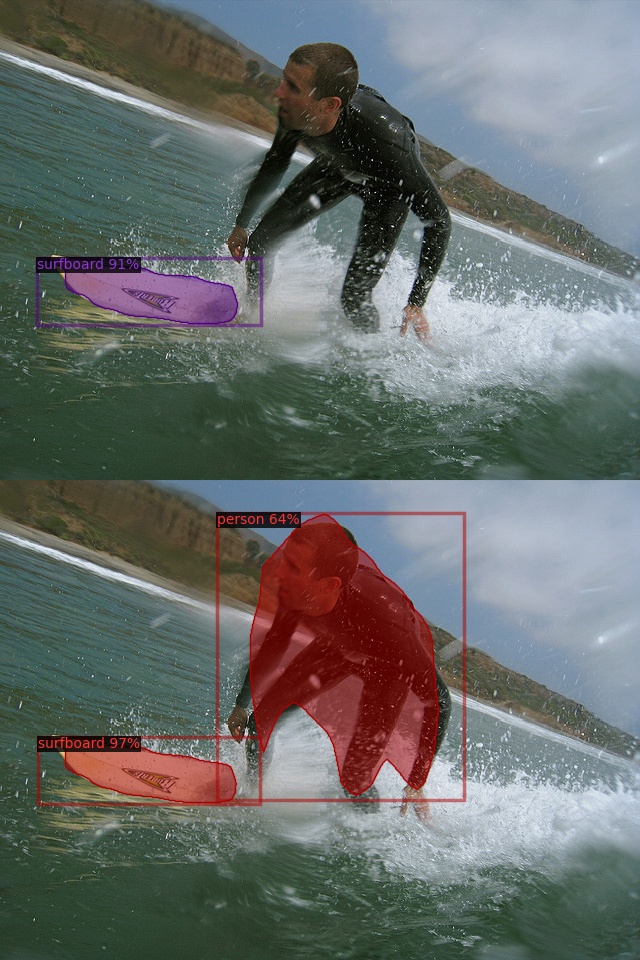}
   \hspace{-6pt}
  }
 \subfloat[]
  {
  \includegraphics[height=0.2485\columnwidth]{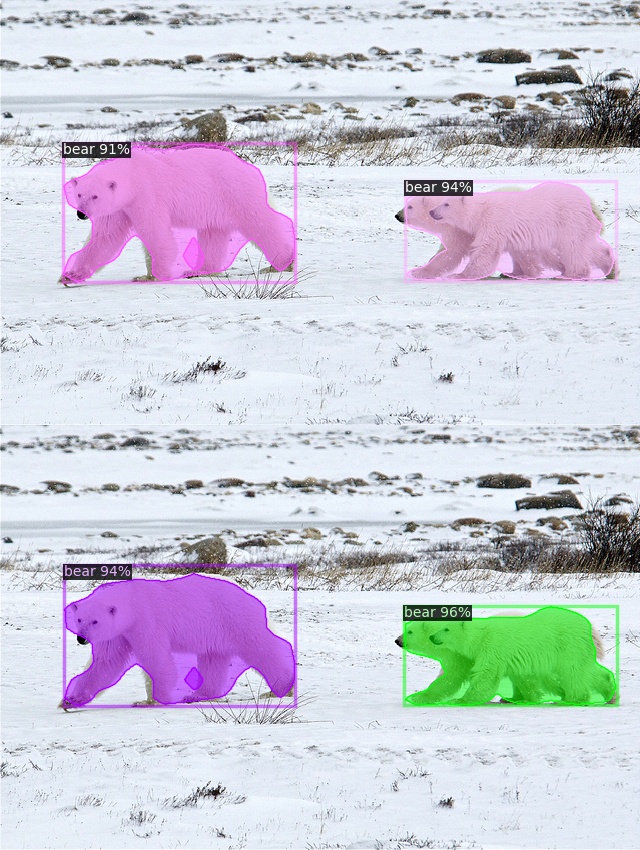}
   \hspace{-6pt}
  }
  \subfloat[]
  {
  \includegraphics[height= 0.2485\columnwidth]{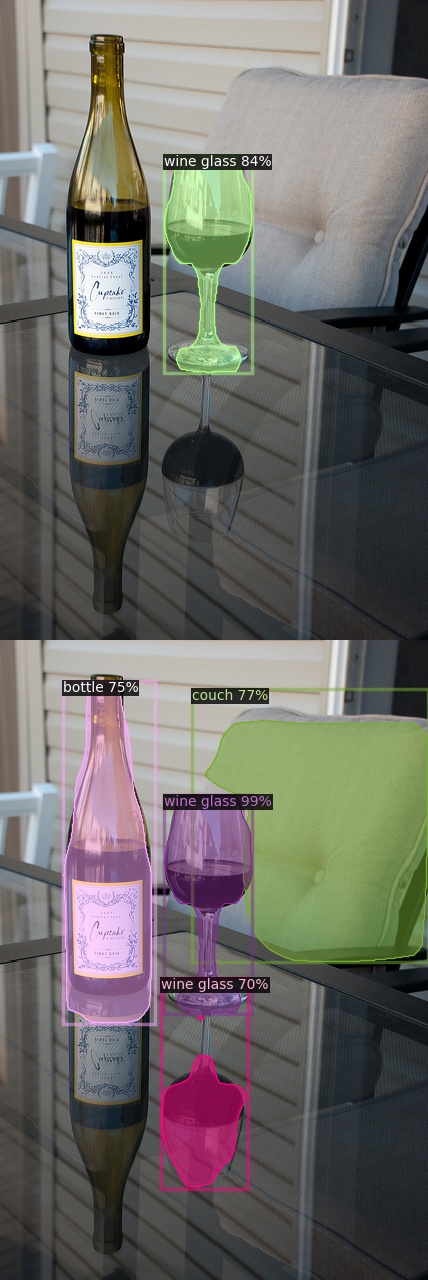}
   \hspace{-6pt}
  }
 \subfloat[]
  {
  \includegraphics[height=0.2485\columnwidth]{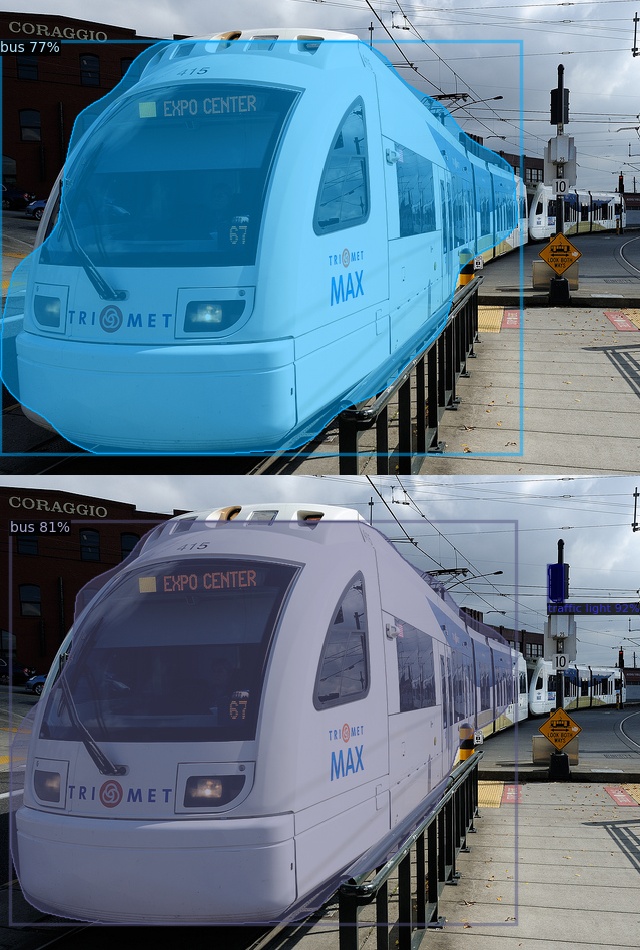}
   \hspace{-6pt}
  }  
  \vspace{-18pt}
  \caption{\small{Visualization results of our method and the strong baseline~(Mask-DeFRCN) on MS-COCO validation images under the gFSIS setting with $K$=10. These bounding boxes and segmentation masks are visualized using classification scores larger than 0.6. The top two rows show success cases with our method and the baseline while the middle two rows show success cases with our method but partly failure ones with the baseline. The baseline may tend to incorrectly recognize positive object regions as background due to the biased classification. The bottom row shows some failure cases from left to right, small objects (e.g., the small boats and the person), coarse boundary segmentation (e.g., the surfer), occlusion (e.g., two bears are detected to one), and misclassification of similar appearance objects (e.g., the shadow of wine glass is recognized to wine glass and the train is detected to bus). Best viewed in color and zoom in.}} \label{fig:vis-res-full}
 \end{figure}

In Fig.~\ref{fig:vis-res-full}, we visualize the results of our method and the strong baseline~(Mask-DeFRCN) on MS-COCO validation images using the gFSIS setting with $K$=10. The top rows show success cases while the bottom row shows failure cases. In the middle rows, we show success cases with our method but partly failures with the baseline. These failures are mainly caused by the missing detection because the baseline method may tend to incorrectly recognize positive objects as background~(i.e., bias classification). In addition, our method may also produce failure predictions as shown in the bottom row from left to right, including the missing detection of small or occultation objects, coarse boundary segmentation, and the misclassification of similar appearance objects.

\section{The Proportion of Missing Labeled Instances}\label{sec:smli}
\begin{figure}[h]
\captionsetup[subfigure]{labelformat=empty}
  \centering
  \vspace{-10pt}
  \subfloat[]
  {
  \includegraphics[width= 0.3\columnwidth]{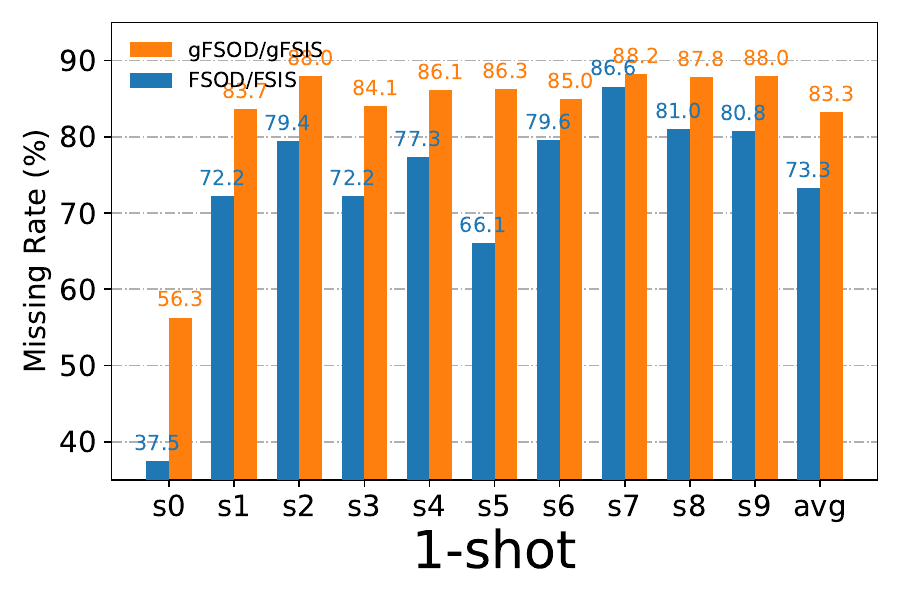}
  }
  \subfloat[]
  {
  \includegraphics[width= 0.3\columnwidth]{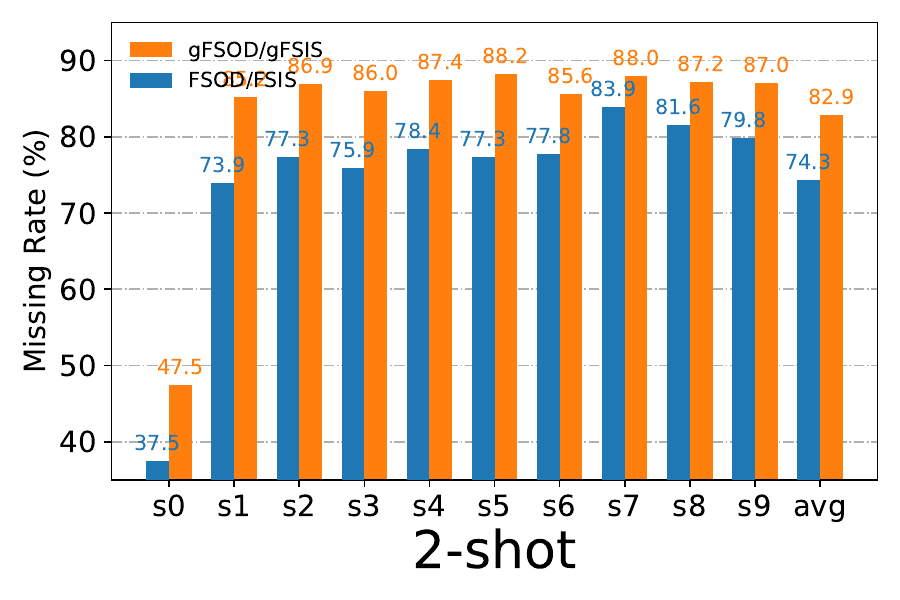}
  }
  \subfloat[]
  {
  \includegraphics[width=0.3\columnwidth]{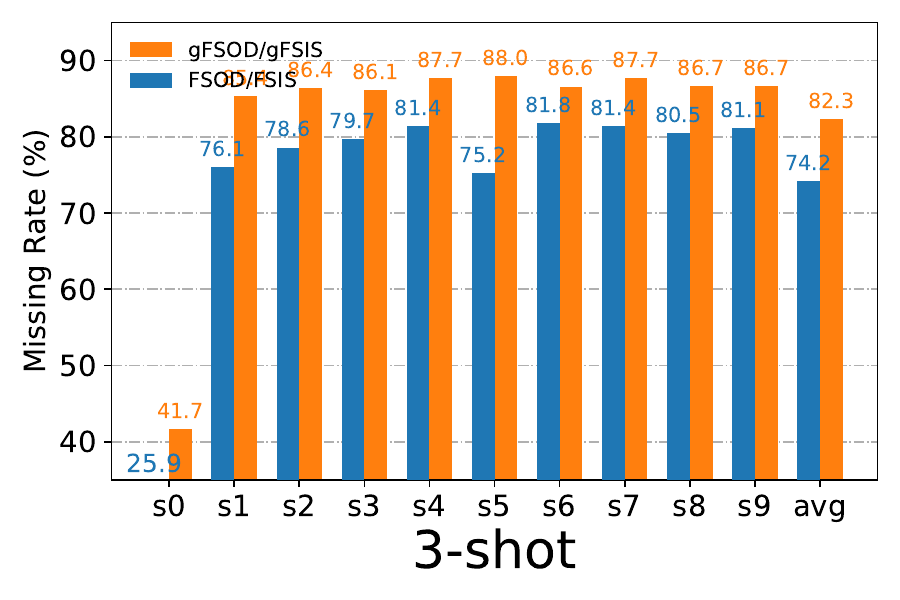}
  }\\
  \vspace{-30pt}
  \subfloat[]
  {
  \includegraphics[width= 0.3\columnwidth]{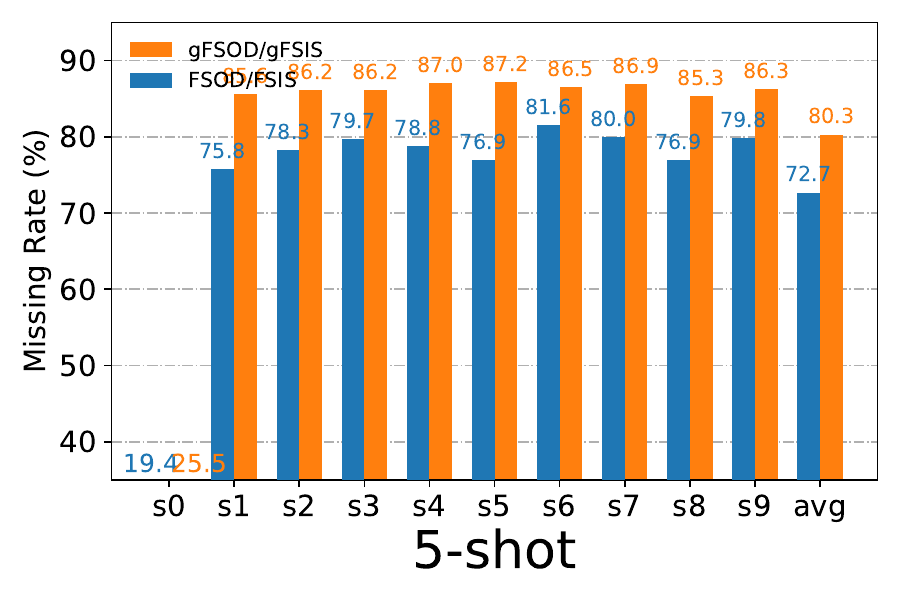}
  }
  \subfloat[]
  {
  \includegraphics[width= 0.3\columnwidth]{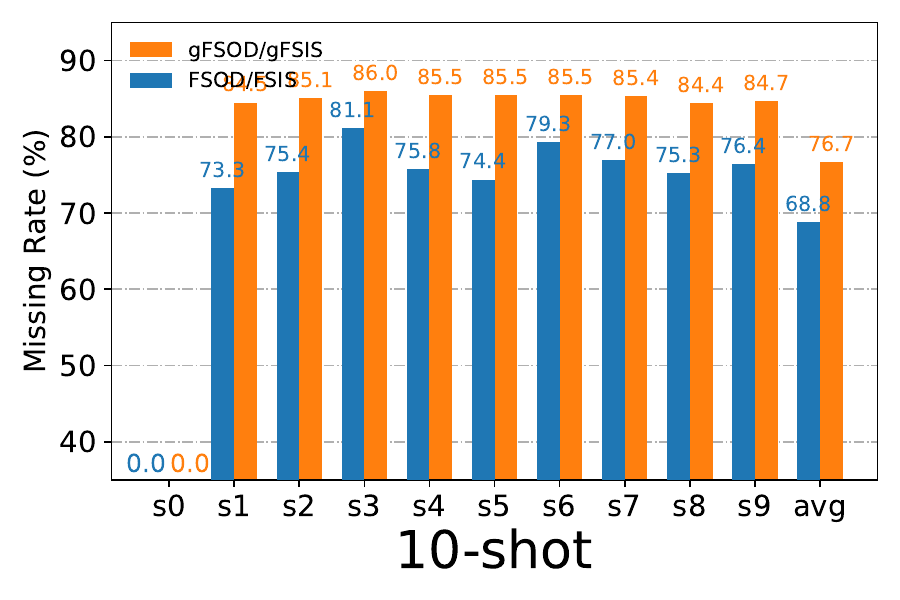}
  }
  \subfloat[]
  {
  \includegraphics[width=0.3\columnwidth]{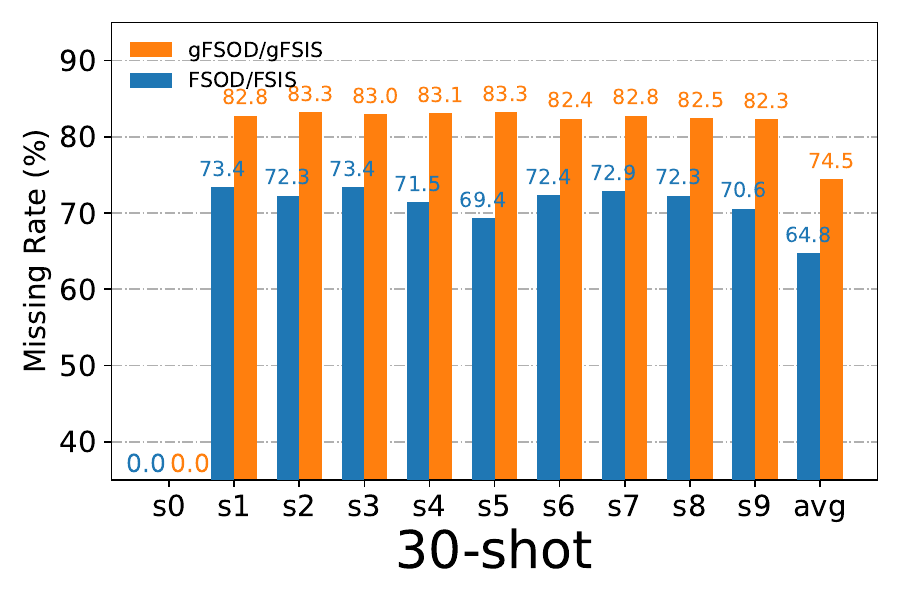}
  }
  \vspace{-20pt}
  \caption{\small{Comparisons of the proportion of missing labeled instances of FSOD/FSIS and gFSOD/gFSIS on the MS-COCO dataset. We can see that there are high proportions almost on all shots using different seeds except the seed0; and the gFSOD/gFSIS setting generally has higher missing rates than that of FSOD/FSIS.}} \label{fig:miss-coco}
 \end{figure}

\begin{figure}[t]
 \captionsetup[subfigure]{labelformat=empty}
  \centering
  \subfloat[]   {{Set1}}
  \subfloat[]
  {
  \includegraphics[width= 0.3\columnwidth]{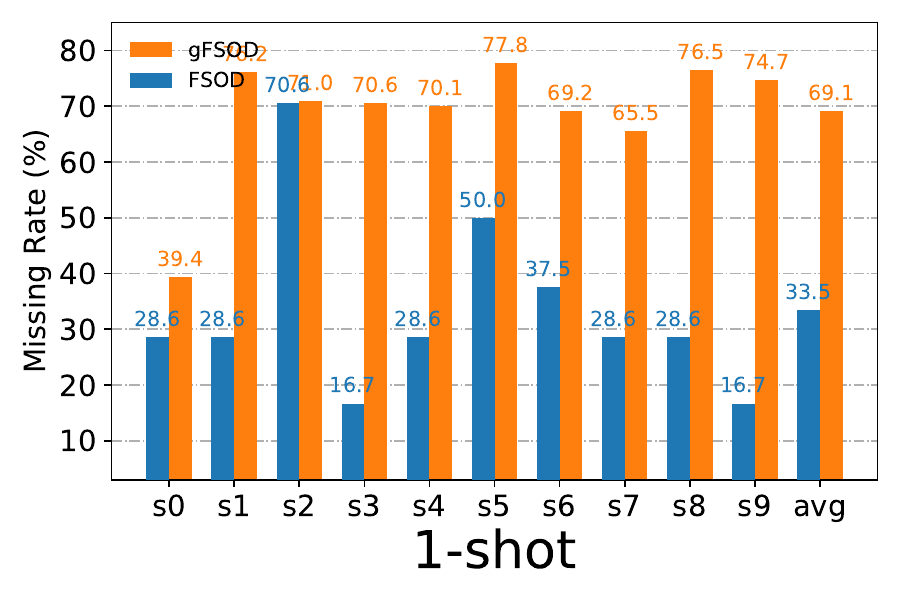}
  }
  \subfloat[]
  {
  \includegraphics[width= 0.3\columnwidth]{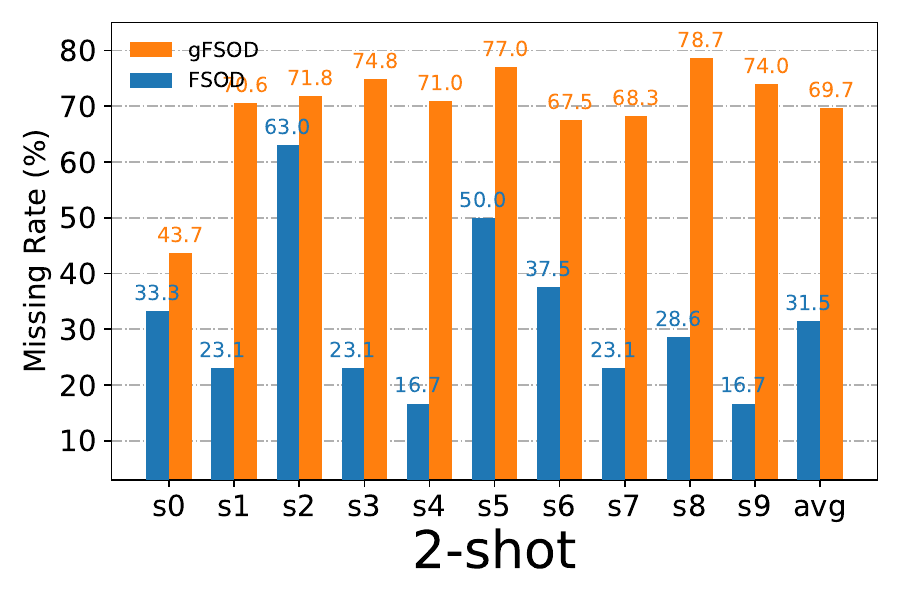}
  }
  \subfloat[]
  {
  \includegraphics[width=0.3\columnwidth]{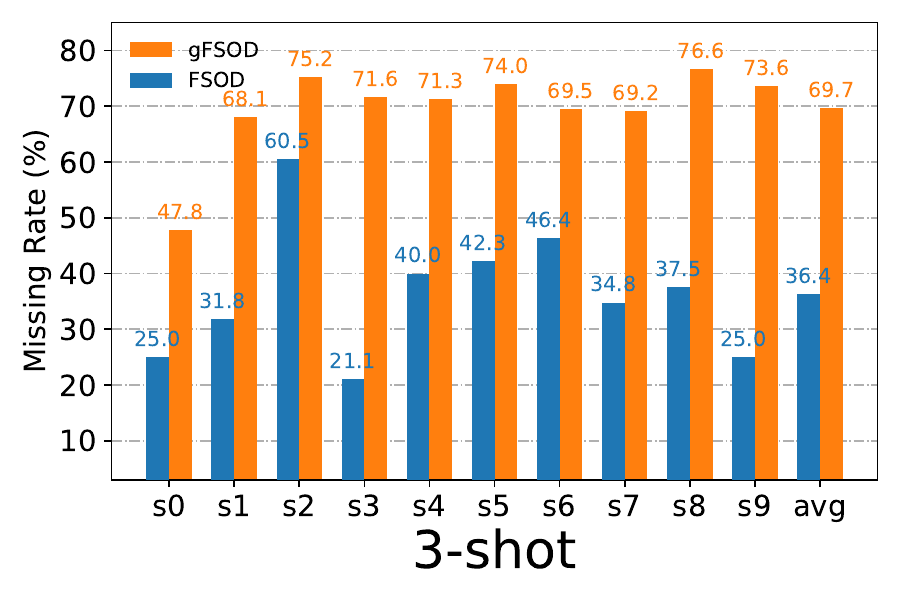}
  }\\
  \vspace{-10pt}
  \quad  \quad
  \subfloat[]
  {
  \includegraphics[width= 0.3\columnwidth]{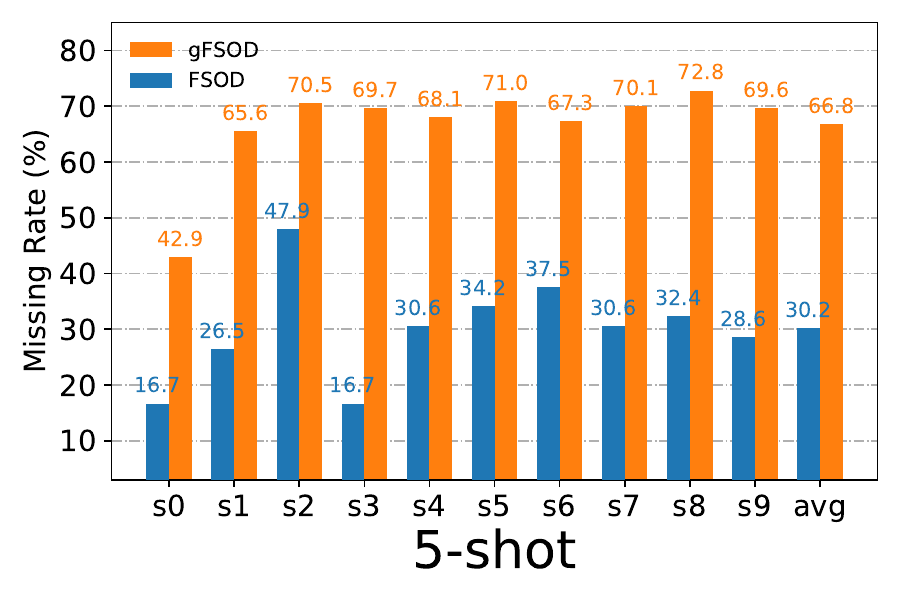}
  }
  \subfloat[]
  {
  \includegraphics[width= 0.3\columnwidth]{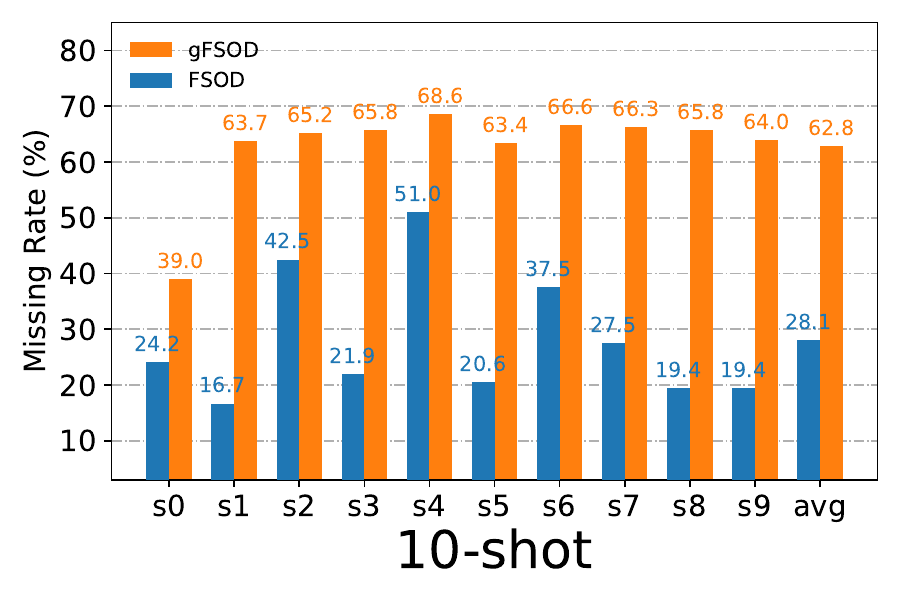}
  }
  \subfloat[]
  {
  \includegraphics[width=0.3\columnwidth]{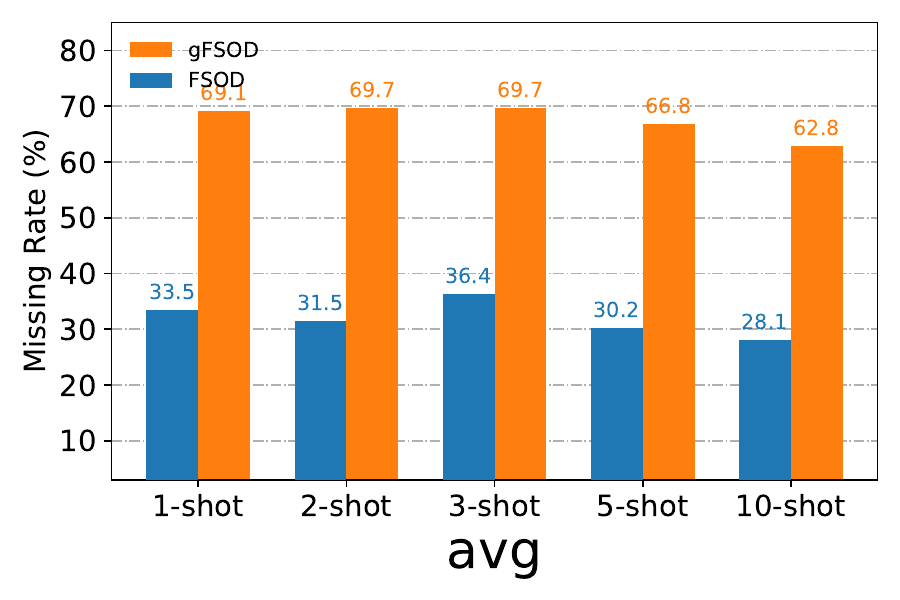}
  }\\
  \subfloat[]   {{Set2}}
  \subfloat[]
  {
  \includegraphics[width= 0.3\columnwidth]{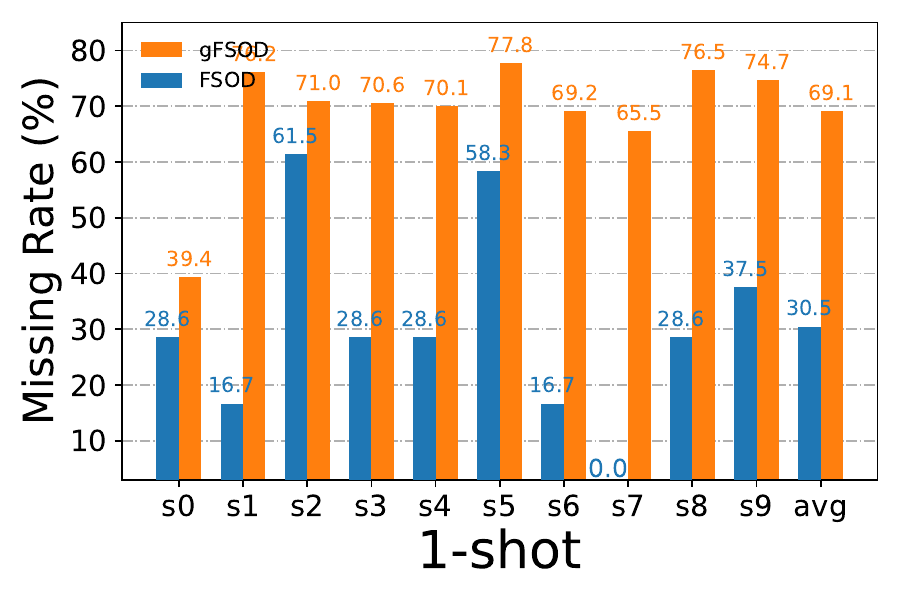}
  }
  \subfloat[]
  {
  \includegraphics[width= 0.3\columnwidth]{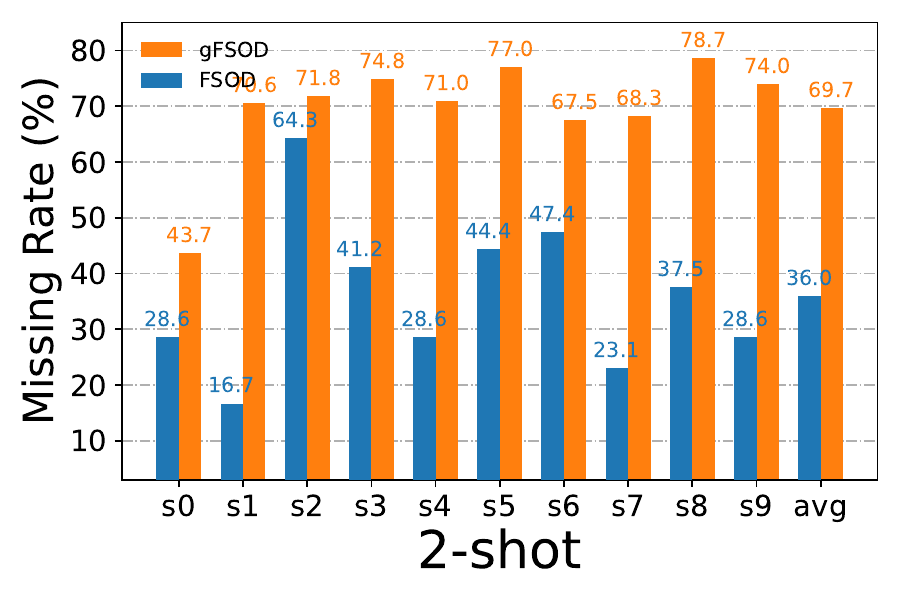}
  }
  \subfloat[]
  {
  \includegraphics[width=0.3\columnwidth]{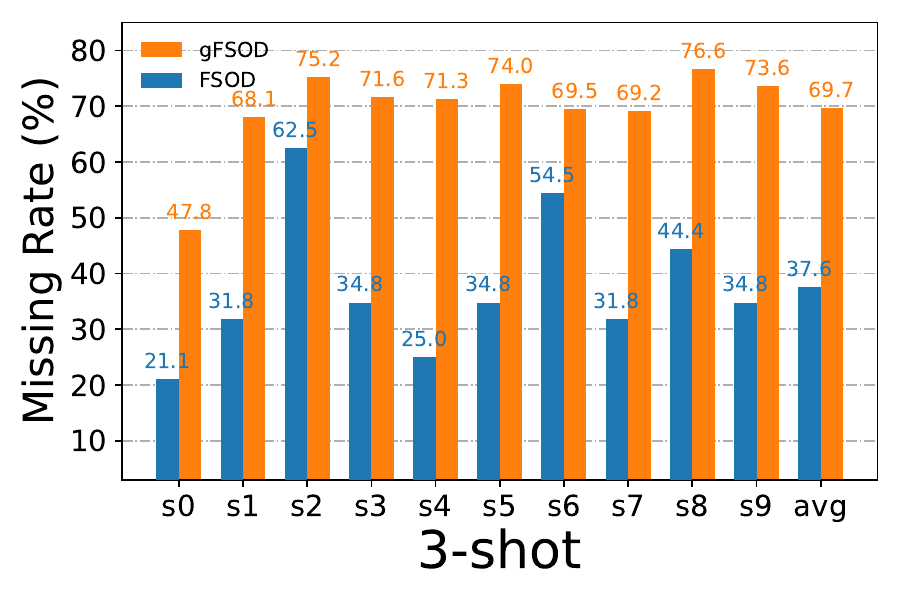}
  }\\
   \vspace{-10pt}
  \quad  \quad
  \subfloat[]
  {
  \includegraphics[width= 0.3\columnwidth]{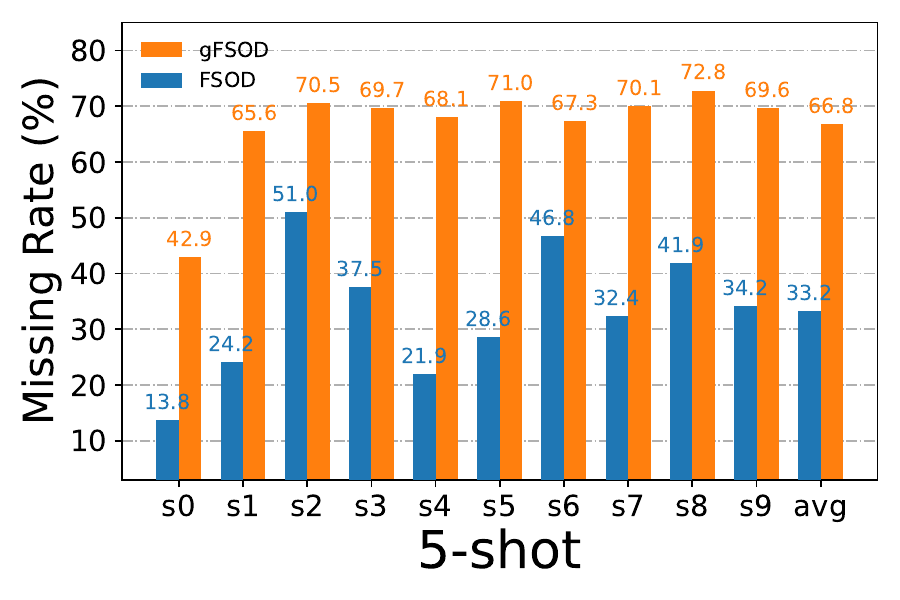}
  }
  \subfloat[]
  {
  \includegraphics[width= 0.3\columnwidth]{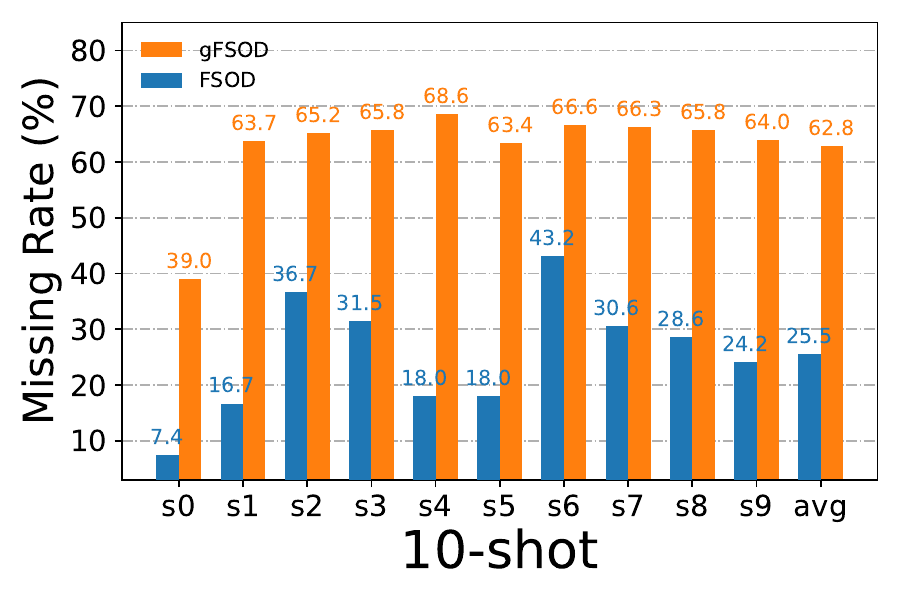}
  }
  \subfloat[]
  {
  \includegraphics[width=0.3\columnwidth]{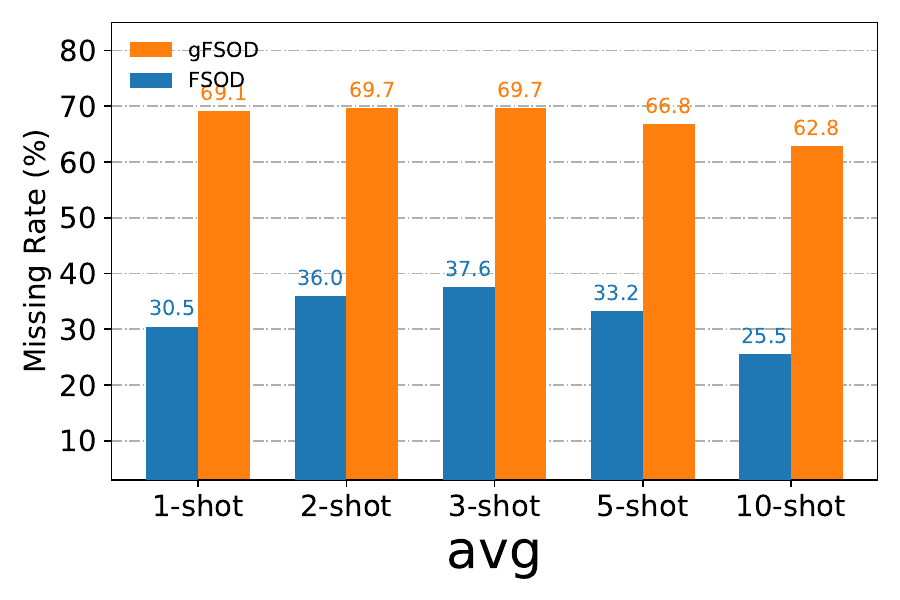}
  }  \\
  \subfloat[]   {{Set3}}
  \subfloat[]
  {
  \includegraphics[width= 0.3\columnwidth]{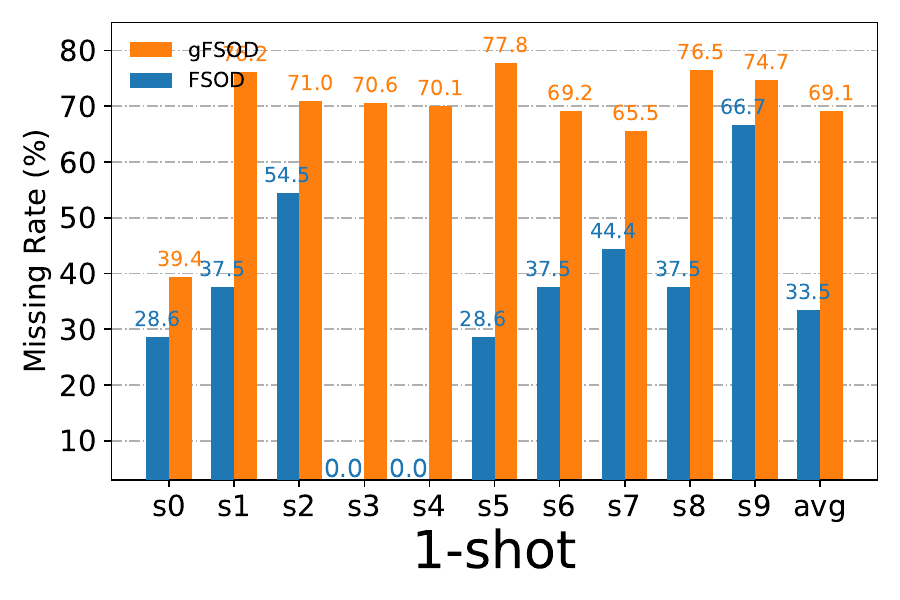}
  }
  \subfloat[]
  {
  \includegraphics[width= 0.3\columnwidth]{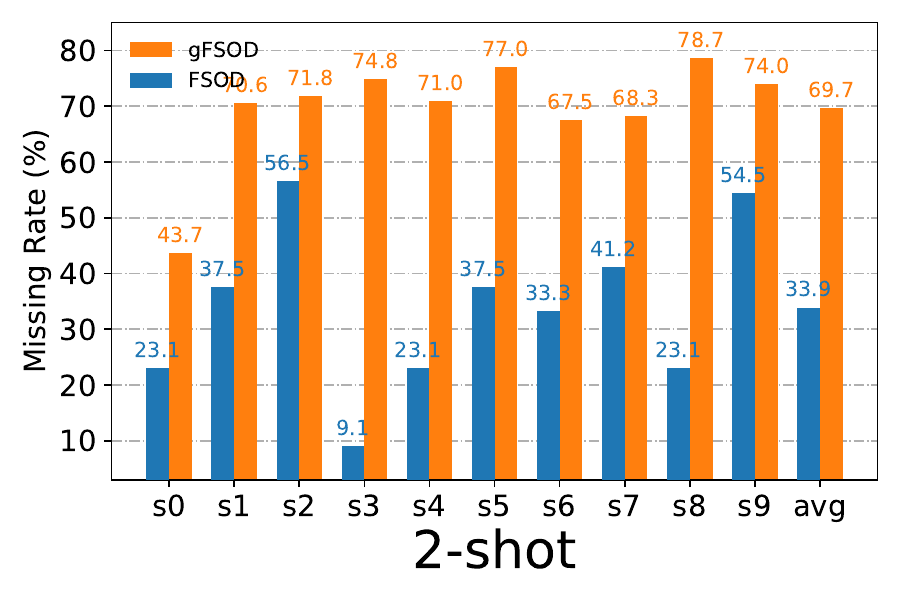}
  }
  \subfloat[]
  {
  \includegraphics[width=0.3\columnwidth]{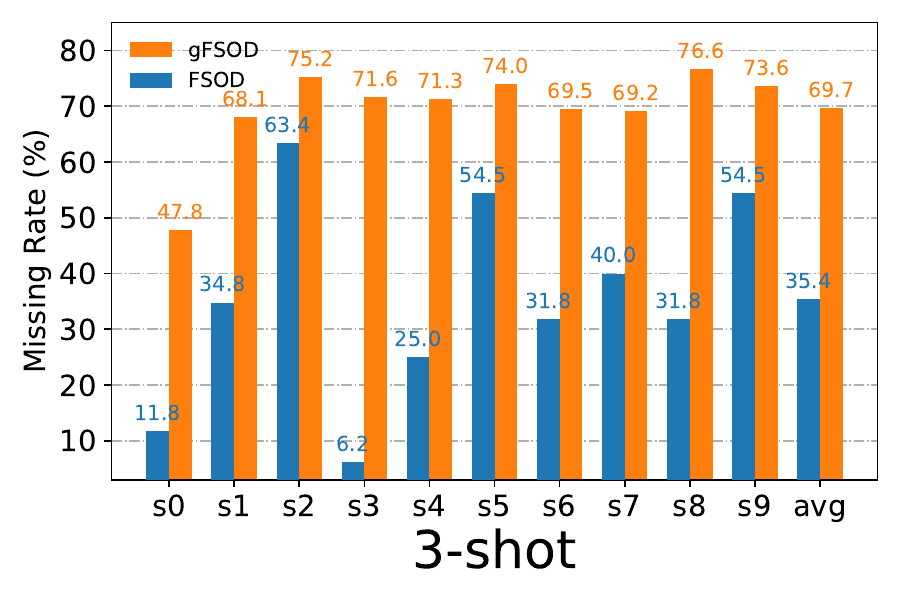}
  }\\
   \vspace{-10pt}
  \quad  \quad
  \subfloat[]
  {
  \includegraphics[width= 0.3\columnwidth]{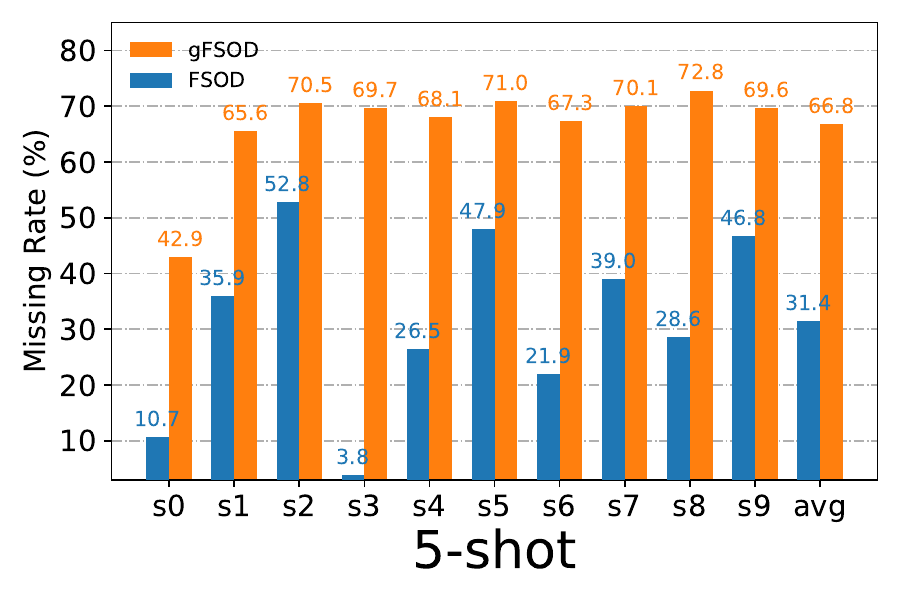}
  }
  \subfloat[]
  {
  \includegraphics[width= 0.3\columnwidth]{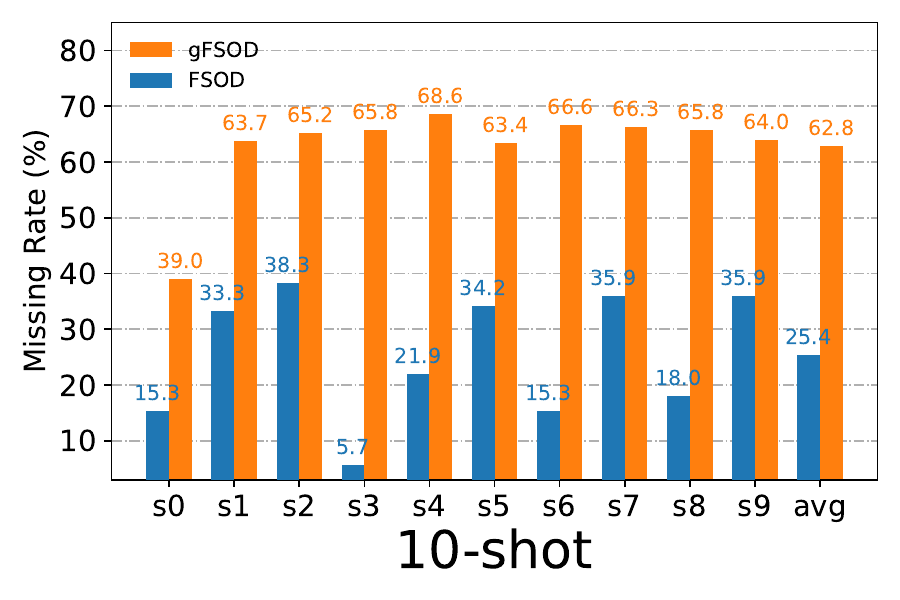}
  }
  \subfloat[]
  {
  \includegraphics[width=0.3\columnwidth]{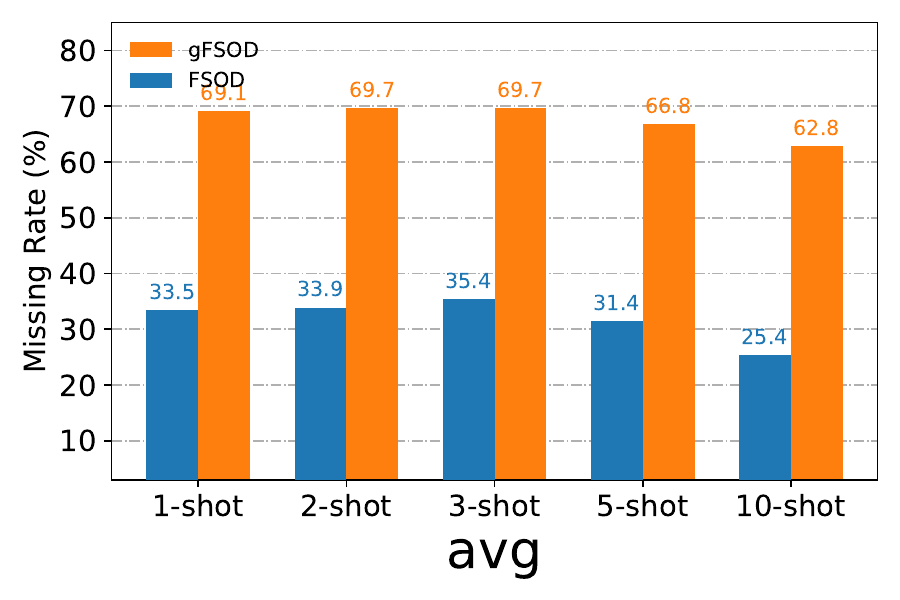}
  }
  \caption{\small{Comparisons of the proportion of missing labeled instances of FSOD and gFSOD on the PASCAL VOC dataset.  Although PASCAL VOC is simpler than MS-COCO, there are still similar observations (high missing rates) on the PASCAL VOC dataset. Different from the MS-COCO, the missing rate is the same among three sets on each shot for the gFSOD setting.}} \label{fig:miss-voc}
 \end{figure}

Here, we provide the detailed missing rates on each seed for MS-COCO in Fig.~\ref{fig:miss-coco} and PASCAL VOC in~Fig.~\ref{fig:miss-voc}.

\end{document}